\documentclass[twoside,11pt]{article}

\usepackage{jmlr2e_arxiv}

\usepackage{amsmath}
\usepackage{subfigure}
\usepackage{graphicx}
\usepackage{bm}
\usepackage[ruled]{algorithm2e}
\usepackage{url}

\usepackage[breaklinks,hidelinks,bookmarks=false]{hyperref}

\usepackage{pslatex}

\newcommand{\w}{\mathbf{w}}
\newcommand{\vv}{\mathbf{v}}
\newcommand{\z}{\mathbf{z}}
\newcommand{\x}{\mathbf{x}}
\newcommand{\X}{\mathbf{X}}
\newcommand{\y}{\mathbf{y}}
\newcommand{\h}{\mathbf{h}}
\newcommand{\g}{\mathbf{g}}
\newcommand{\A}{\mathbf{A}}
\newcommand{\B}{\mathbf{B}}
\newcommand{\I}{\mathbf{I}}
\newcommand{\D}{\mathcal{D}}

\newcommand{\m}{\mathbf{m}}
\newcommand{\V}{\mathbf{V}}
\newcommand{\uu}{\mathbf{u}}
\newcommand{\Sq}{\bm{\Sigma}}
\newcommand{\mq}{\bm{\mu}}
\newcommand{\mut}{\tilde{\mu}}
\newcommand{\sigmat}{\tilde{\sigma}}

\newcommand{\Tt}{\tilde{\mathbf{T}}}
\newcommand{\bt}{\tilde{\mathbf{b}}}

\newcommand{\Nut}{\tilde{\bm{\nu}}}
\newcommand{\Taut}{\tilde{\bm{\tau}}}

\newcommand{\Tauvl}{ \tilde{\bm{\alpha}} }
\newcommand{\Nuvl}{ \tilde{\bm{\beta}} }

\newcommand{\tauvl}{ \tilde{\alpha} }

\newcommand{\Tauwl}{ \tilde{\bm{\tau}} }
\newcommand{\Nuwl}{ \tilde{\bm{\nu}} }

\newcommand{\tauwl}{ \tilde{\tau} }
\newcommand{\nuwl}{ \tilde{\nu} }

\DeclareMathOperator{\argmin}{arg\,min}
\DeclareMathOperator{\KL}{KL}
\DeclareMathOperator{\cov}{Cov}
\DeclareMathOperator{\E}{E}
\DeclareMathOperator{\var}{Var}
\DeclareMathOperator{\erf}{erf}

\newcommand{\mb}[1]{\mathbf{#1}}

\newcommand{\Tr}{\text{T}}
\newcommand{\N}{\mathcal{N}}
\newcommand{\diag}{\textrm{diag}}

\ShortHeadings{Expectation Propagation for Neural Networks with Sparse Priors}{Jyl\"{a}nki, Nummenmaa and Vehtari}
\firstpageno{1}

\begin{document}

\title{Expectation Propagation for Neural Networks with Sparsity-promoting Priors}

\author{\name Pasi Jyl\"{a}nki \email pasi.jylanki@aalto.fi \\
       \addr Department of Biomedical Engineering and Computational Science\\
       Aalto University\\
       P.O. Box 12200 \\ FI-00076 AALTO, Espoo \\ Finland
       \AND
       \name Aapo Nummenmaa \email nummenma@nmr.mgh.harvard.edu\\
       \addr Athinoula A. Martinos Center for Biomedical Imaging\\
       Massachusetts General Hospital\\
       Harvard Medical School\\ Boston\\ MA 02129
       \AND
       \name Aki Vehtari \email aki.vehtari@aalto.fi \\
       \addr Department of Biomedical Engineering and Computational Science\\
       Aalto University\\
       P.O. Box 12200 \\ FI-00076 AALTO, Espoo \\ Finland}

\editor{}

\maketitle \thispagestyle{empty}

\begin{abstract}%   <- trailing '%' for backward compatibility of .sty file
We propose a novel approach for nonlinear regression using a two-layer neural
network (NN) model structure with sparsity-favoring hierarchical priors on
the network weights. We present an expectation propagation (EP) approach for
approximate integration over the posterior distribution of the weights, the
hierarchical scale parameters of the priors, and the residual scale. Using a
factorized posterior approximation we derive a computationally efficient
algorithm, whose complexity scales similarly to an ensemble of independent
sparse linear models. The approach enables flexible definition of weight
priors with different sparseness properties such as independent Laplace
priors with a common scale parameter or Gaussian automatic relevance
determination (ARD) priors with different relevance parameters for all
inputs. The approach can be extended beyond standard activation functions and
NN model structures to form flexible nonlinear predictors from multiple
sparse linear models. The effects of the hierarchical priors and the
predictive performance of the algorithm are assessed using both simulated and
real-world data. Comparisons are made to two alternative models with ARD
priors: a Gaussian process with a NN covariance function and marginal maximum
a posteriori estimates of the relevance parameters, and a NN with Markov
chain Monte Carlo integration over all the unknown model parameters.
\end{abstract}


\section{Introduction}

Gaussian priors may not be the best possible choice for the input layer weights %$\w_k$
of a feedforward neural network (NN)
%
%(or analogously for the weights of a linear model)
%
because allowing, {\it a priori}, a large weight $w_{j}$ for a potentially
irrelevant feature $x_j$ may deteriorate the predictive performance. This
behavior is analogous to a linear model because the input layer weights
associated with each hidden unit of a multilayer perceptron (MLP) network can
be interpreted as separate linear models whose outputs are combined
nonlinearly in the next layer.
Integrating over the posterior uncertainty of the unknown input weights
mitigates the potentially harmful effects of irrelevant features but it may
not be sufficient if the number of input features, or the total number of
unknown variables, grows large compared with the number of observations. In
such cases, an alternative strategy is required to suppress the effect of the
irrelevant features.
In this article we focus on a two-layer MLP model structure but aim to form a
more general framework that can be used to combine linear models with
sparsity-promoting priors using general activation functions and interaction
terms between the hidden units.

A popular approach has been to apply hierarchical automatic relevance
determination (ARD) priors \citep{MacKay:1995, Neal:1996a}, where individual
Gaussian priors are assigned for each weight, $w_{j} \sim
\N(0,\alpha_{l_j})$, with separate variance hyperparameters $\alpha_{l_j}$
controlling the relevance of the group of weights related to each feature.
\citet{MacKay:1995} described an ARD approach for NNs, where point estimates
for the relevance parameters $\alpha_{l_j}$ along with other model
hyperparameters, such as the noise level, are determined using a marginal
likelihood estimate obtained by approximate integration over the weights with
Laplace's method.
\citet{Neal:1996a} proposed an alternative Markov chain Monte Carlo (MCMC)
approach, where approximate integration is performed over the posterior
uncertainty of all the model parameters including $w_{j}$ and $\alpha_{l_j}$.
In connection with linear models, various computationally more efficient
algorithms have been proposed for determining marginal likelihood based point
estimates for the relevance parameters \citep{Tipping:2001, Qi:2004,
Wipf:2008}.
%
% by utilizing the structure resulting from the linear observation model
%
The point-estimate based methods, however, may suffer from overfitting
because the maximum a posteriori (MAP) estimate of $\alpha_{l_j}$ may be
close to zero also for relevant features as demonstrated by \citet{Qi:2004}.
The same applies also for infinite neural networks implemented using Gaussian
process (GP) priors when separate hyperparameters controlling the
nonlinearity of each input are optimized \citep[][]{Williams:1998,
Rasmussen+Williams:2006}.

Recently, appealing surrogates for ARD priors have been presented for linear
models. These approaches are based on sparsity favoring priors, such as the
Laplace prior \citep{Seeger:2008} and the spike and slab prior
\citep{hernandez:2008,hernandez:2010a}. The methods utilize the expectation
propagation (EP) \citep{Minka:2001b} algorithm to efficiently integrate over
the analytically intractable posterior distributions. Importantly, these
sparse priors do not suffer from similar overfitting as many ARD approaches
because point estimates of feature specific parameters such as $\alpha_{l_j}$
are not used, but instead, approximate integration is done over the posterior
uncertainty resulting from a sparse prior on the weights. Expectation
propagation provides a useful alternative to MCMC for carrying out the
approximate integration because it has been found computationally efficient
and very accurate in many practical applications
\citep{Nickisch+Rasmussen:2008,hernandez:2010a}.

In nonlinear regression, sparsity favoring Laplace priors have been
considered for NNs, for instance, by \citet{Williams:1995}, where the
inference is performed using the Laplace approximation. However, a problem
with the Laplace approximation is that the curvature of the log-posterior
density at the posterior mode may not be well defined for all types of prior
distributions, such as, the Laplace distribution whose derivatives are not
continuous at the origin \citep{Williams:1995,Seeger:2008}. Implementing a
successful algorithm requires some additional approximations as described by
\citet{Williams:1995}, whereas with EP the implementation is straightforward
since it relies only on expectations of the prior terms with respect to a
Gaussian measure.

Another possibly undesired characteristic of the Laplace approximation is
that it approximates the posterior mean of the unknowns with their MAP
estimate and their posterior covariance with the negative Hessian of the
posterior distribution at the mode. This local estimate may not represent
well the overall uncertainty on the unknown variables and it may lead to
worse predictive performance for example when the posterior distribution is
skewed \citep{Nickisch+Rasmussen:2008} or multimodal \citep{Jylanki:2011}.
Furthermore, when there are many unknowns compared to the effective number of
observations, which is typical in practical NN applications, the MAP solution
may differ significantly from the posterior mean. For example, with linear
models the Laplace prior leads to strictly sparse estimates with many zero
weight values only when the MAP estimator of the weights is used. The
posterior mean estimate, on the other hand, can result in many clearly
nonzero values for the same weights whose MAP estimates are zero
\citep{Seeger:2008}.
In such case the Laplace approximation underestimates the uncertainty of the
feature relevances, that is, the joint mode is sharply peaked at zero but the
bulk of the probability mass is distributed widely at nonzero weight values.
%
%produce overconfident marginal probability estimates on the significance of
%the effects a particular feature has on the predictions.
%
Recently, it has also been shown that in connection with linear models the
ARD solution is exactly equivalent to a MAP estimate of the coefficients
obtained using a particular class of non-factorized coefficient prior
distributions which includes models that have desirable advantages over MAP
weight estimates \citep{Wipf:2008,Wipf:2011}. This connection suggests that
the Laplace approximation of the input weights with a sparse prior may
possess more similar characteristics with the point-estimate based ARD
solution compared to the posterior mean solution.

%Recently, in connection with sparse linear models, interesting relationship
%has been found between the ARD solution based on point-estimates of the
%relevance hyperparameters and the MAP estimate of the coefficients resulting
%from sparsity-promoting priors (this class of sparse coefficient estimates
%includes methods such as the widely used $\mathcal{L}_1$-norm solution which
%is equivalent to the MAP estimate resulting from independent Laplace priors
%on the coefficients). It has been shown that the ARD solution is exactly
%equivalent to a MAP estimate of the coefficients obtained using a particular
%class of nonfactorial coefficient prior distributions which includes models
%that have desirable advantages over regular MAP estimates
%\citep{Wipf:2008,Wipf:2011}. This connection suggests that the Laplace
%approximation of the input weights with a sparse prior may possess more
%similar characteristics with the point-estimate based ARD solution than with
%the posterior mean solution.

Our aim is to introduce the benefits of the sparse linear models
\citep{Seeger:2008,hernandez:2008} into nonlinear regression by combining the
sparse priors with a two-layer NN in a computationally efficient EP
framework. We propose a logical extension of the linear regression models by
adopting the algorithms presented for sparse linear models to MLPs with a
linear input layer. For this purpose, the main challenge is constructing a
reliable Gaussian EP approximation for the analytically intractable
likelihood resulting from the NN observation model.
%
%that enables computationally efficient moment evaluations of the various
%approximate distributions required for the EP algorithm.
%
Previously, Gaussian approximations for NN models have been formed using the
extended Kalman filter (EKF) \citep{Freitas:1999} and the unscented Kalman
filter (UKF) \citep{Wan:2000}. Alternative mean field approaches possessing
similar characteristic with EP have been proposed by \citet{Opper:1996} and
\citet{Winther:2001}.

We extend the ideas from the UKF approach by utilizing similar decoupling
approximations for the weights as proposed by \citet{Puskorius:1991} for
EKF-based inference and derive a computationally efficient algorithm that
does not require numerical sigma point approximations for multi-dimensional
integrals.
We propose a posterior approximation that assumes the weights associated with
the output-layer and each hidden unit independent. The complexity of the EP
updates in the resulting algorithm scale linearly with respect to the number
of hidden units and they require only one-dimensional numerical quadratures.
The complexity of the posterior computations scale similarly to an ensemble
of independent sparse linear models (one for each hidden unit) with one
additional linear predictor associated with the output layer. It follows that
all existing methodology on sparse linear models (e.g., methods that
facilitate computations with large number of inputs) can be applied
separately on the sparse linear model corresponding to each hidden unit.
Furthermore, the complexity of the algorithm scales linearly with respect to
the number of observations, which is beneficial for large datasets.
The proposed approach can also be extended beyond standard activation
functions and NN model structures, for example, by including a linear hidden
unit or predefined interactions between the linear input-layer models.

%Despite of the explicit independence assumptions between the different hidden
%units and hyperparameters implicit interactions are made possible through the
%observation model and the hierarchical weight priors during the iterative
%algorithm.

%By assuming the hidden units a posterior independent the computational
%complexity of the approach scales similarly to an ensemble of sparse linear
%models which interact through the nonlinear observation model and the
%hierarchical priors defined on the input weights.

In addition to generalizing the standard EP framework for sparse linear
models we introduce an efficient EP approach for inference on the unknown
hyperparameters, such as the noise level and the scale parameters of the
weight priors. This framework enables flexible definition of different
hierarchical priors, such as one common scale parameter for all input
weights, or a separate scale parameter for all weights associated with one
input variable (i.e., an integrated ARD prior).
For example, assigning independent Laplace priors on the input weights with a
common unknown scale parameter does not produce very sparse approximate
posterior mean solutions, but, if required, more sparse solutions can be
obtained by adjusting the common hyperprior of the scale parameters with the
ARD specification.
%
%Additional benefits are achieved by placing a sparse prior also on the output
%layer, which enables automatic control of the model complexity by effectively
%removing unnecessary nonlinear hidden units from the model. This can be
%utilized further in speeding up the computations, as will be described later.
%
We show that by making independent approximations for the hyperparameters,
the inference on them can be done simultaneously within the EP algorithm for
the network weights, without the need for separate optimization steps which
is common for many EP approaches for sparse linear models and GPs
\citep{Rasmussen+Williams:2006,Seeger:2008}, as well as combined EKF and
expectation maximization (EM) algorithms for NNs \citep{Freitas:1999}.
%
%Because there is no convergence guarantee for the standard EP algorithm and
%the derivatives of the EP marginal likelihood approximation can be calculated
%efficiently using explicit hyperparameter dependencies only after convergence
%\citep{Opper:2005,Seeger:2005}, we propose
%
Additional benefits are achieved by introducing left-truncated priors on the
output weights which prevent possible convergence problems in the EP
algorithm resulting from inherent unidentifiability in the MLP network
specification.

The main contributions of the paper can be summarized as:
\begin{itemize}
  \item An efficiently scaling EP approximation for the non-Gaussian
      likelihood resulting from the MLP-model that requires numerical
      approximations only for one-dimensional integrals.
      We derive closed-form solutions for the parameters of the site term
      approximations, which can be interpreted as pseudo-observations of
      a linear model associated with each hidden unit (Sections
      \ref{sec_approximation}--\ref{sec_implementation}
      and Appendices \ref{sec_cavity}--\ref{sec_site_updates}).
  \item An EP approach for integrating over the posterior uncertainty of
      the input weights and their hierarchical scale parameters assigned
      on predefined weight groups (Sections \ref{sec_approximation} and
      \ref{sec_ep_weight_priors}). The approach could be applied also for
      sparse linear models to construct factorized approximations for
      predefined weight groups with shared hyperparameters.
  \item Approximate integration over the posterior uncertainty of the
      observation noise simultaneously within the EP algorithm for
      inference on the weights of a MLP-network (see Appendix
      \ref{sec_qtheta_tilted}). Using factorizing approximations, the
      approach could be extended also for approximate inference on other
      hyperparameters associated with the likelihood terms and could be
      applied, for example, in recursive filtering.
\end{itemize}

Key properties of the proposed model are first demonstrated with three
artificial case studies in which comparisons are made with a neural network
with infinitely many hidden units implemented as a GP regression model with a
NN covariance function and an ARD prior \citep{Williams:1998,
Rasmussen+Williams:2006}. Finally, the predictive performance of the proposed
approach is assessed using four real-world data sets and comparisons are made
with two alternative models with ARD priors:
a GP with a NN covariance function where point estimates of the relevance
hyperparameters are determined by optimizing their marginal posterior
distribution, and a NN where approximate inference on all unknowns is done
using MCMC \citep{Neal:1996a}.

\section{The Model} \label{sec_model}
We focus on two layer NNs where the unknown function value $f_i=f(\x_i)$
related to a $d$-dimensional input vector $\x_i$ is modeled as
\begin{equation} \label{eq_fx_mlp}
  f(\x_i) = \sum_{k=1}^K v_k g(\w_k^{\Tr} \x_i) + v_0,
\end{equation}
where $g(x)$ is a nonlinear activation function, $K$ the number of hidden
units, and $v_0$ the output bias. Vector $\w_k =[w_{k,1}, w_{k,2},...,
w_{k,d}]^{\Tr}$ contains the input layer weights related to hidden unit $k$
and $v_k$ is the corresponding output layer weight. Input biases can be
introduced by adding a constant term to the input vectors $\x_k$. In the
sequel, all weights are denoted by vector $\z=[\w^{\Tr},\vv^{\Tr}]^{\Tr}$,
where $\w=[\w_1^{\Tr},..., \w_K^{\Tr}]^{\Tr}$, and $\vv=[v_1,..., v_K,
v_0]^{\Tr}$.

In this work, we use the following activation function:
\begin{equation} \label{eq_gx}
  g(x) = \frac{1}{\sqrt{K}} \erf \left( \frac{x}{\sqrt{2}} \right)
  = \frac{2}{\sqrt{K}} \left( \Phi(x) -0.5 \right),
\end{equation}
where $\Phi(x)=\int_{-\infty}^x \N(t|0,1) dt$, and the scaling by
$1/\sqrt{K}$ ensures that the prior variance of $f(\x_i)$ does not increase
with $K$ assuming fixed Gaussian priors on $v_k$ and $w_{kj}$.
We focus on regression problems with Gaussian observation model
$p(y_i|f_i,\sigma^2) =\N(y_i|f_i,\sigma^2)$, where $\sigma^2$ is the noise
variance. However, the proposed approach could be extended for other
activation functions and observations models, for example, the probit model
for binary classification.

\subsection{Prior Definitions}

To reduce the effects of irrelevant features, independent zero-mean Laplace
priors are assigned for the input layer weights:
%$p(v_k) = \N(0,\sigma_v^2)$.
%
\begin{equation} \label{eq_prw}
  p(w_{j}| \lambda_{l_j} ) = \frac{1}{2\lambda_{l_j}}
  \exp\left( -\frac{1}{ \lambda_{l_j} } |w_{j}| \right),
\end{equation}
where $w_{j}$ is the $j$:th element of $\w$, and $\lambda_{l_j} = 2^{-1/2}
\exp( \phi_{l_j}/2) $ is a joint hyperparameter controlling the prior
variance of all input weights belonging to group $l_j \in \{1,...,L \}$, that
is, $\var(w_{j}|\lambda_{l_j}) =2 \lambda_{l_j}^2$. Here index variable $l_j$
defines the group in which the weight $\w_j$ belongs to. The EP approximate
inference is done using the transformed scale parameters $\phi_l= \log(2
\lambda_l^2) \in \mathbb{R}$.
%
%The presented approach is applicable also for other types of weight prior
%distributions, such as, $p(w_{kj}|\lambda_w)=\N(w_{kj}|0,\lambda_w^l)$
%
The grouping of the weights can be chosen freely and also other weight prior
distributions can be used in place of the Laplace distribution
\eqref{eq_prw}.
By defining a suitable prior on the unknown group scales $\phi_l$ useful
hierarchical priors can be implemented on the input layer. Possible
definitions include one common scale parameter for all inputs ($L=1$), and a
separate scale parameter for all weights related to each feature, which
implements an ARD prior ($L=d$). To obtain the traditional ARD setting the
Laplace priors \eqref{eq_prw} can be replaced with Gaussian distributions
$p(w_{j}| \lambda_{l_j} ) = \N(w_j|0,\exp(\phi_{l_j}) )$.
When the scale parameters $\{ \phi_l \}_{l=1}^L$ are considered unknown,
Gaussian hyperpriors are assigned to them:
\begin{equation} \label{eq_pruw}
  \phi_l= \log(2 \lambda_l^2) \sim \N(\mu_{\phi,0},\sigma_{\phi,0}^2),
\end{equation}
where a common mean $\mu_{\phi,0}$ and variance $\sigma_{\phi,0}^2$ have been
defined for all groups $l=1,...,L$. Definition \eqref{eq_pruw} corresponds to
a log-normal prior on the associated prior variance $2 \lambda_{l}^2 = \exp(
\phi_{l})$ for the weights in group $l$.

%
%The log-normal prior is set on the variance instead of $\lambda_l$ to help
%defining the prior uncertainty.
%
%To control the complexity of the network by reducing the effects of
%unnecessary hidden units, we assume independent zero-mean Laplace priors with
%a common scale parameter $\sigma_{v,0}^2$  for the output layer weights $v_k$.
%

Because of the symmetry $g(x) = -g(-x)$ of the activation function, changing
the signs of output weight $v_k$ and the corresponding input weights $\w_k$
results in the same prediction $f(\x)$. This unidentifiability may cause
converge problems in the EP algorithm: if the approximate posterior
probability mass of output weight $v_k$ concentrates too close to zero,
expected values of $v_k$ and $\w_k$ may start fluctuating between small
positive and negative values. Therefore the output weights are constrained to
positive values by assigning left-truncated heavy-tailed priors to them:
\begin{equation} \label{eq_prv}
  %p(v_k|\sigma_{v,0}^2) = 2 \N(v_k| 0,\sigma_{v,0}^2),
  p(v_k|\sigma_{v,0}^2) = 2 t_{\nu} (v_k| 0,\sigma_{v,0}^2),
\end{equation}
where $v_k \ge 0$, $k=1,...,K$, and $t_{\nu} (v_k| 0,\sigma_{v,0}^2)$ denotes
a Student-$t$ distribution with degrees of freedom $\nu$, mean zero, and
scale parameter $\sigma_{v,0}^2$.
%
% \lambda_v -> \sigma_{v,0}^2
%
The prior scale is fixed to $\sigma_{v,0}^2=1$ and the degrees of freedom to
$\nu=4$, which by experiments was found to produce sufficiently large
posterior variations of $f(\x)$ when the activation function is scaled
according to \eqref{eq_gx} and the observations are normalized to zero mean
and unit variance. The heavy-tailed prior \eqref{eq_prv} enables very large
output weights if required, for example, when some hidden unit is forming
almost a linear predictor (see, e.g, Section \ref{sec_case2}).
A zero-mean Gaussian prior is assigned to the output bias: $p(v_0|
\sigma_{v_0,0}^2) = \N(0,\sigma_{v_0,0}^2)$, where the scale parameter is
fixed to $\sigma_{v_0,0}^2=1$ because it was also found to be a sufficient
simplification with the same data normalization conditions.
%
%With the same conditions a Gaussian prior with variance fixed to one was
%found sufficient for the output bias: $p(v_0) = \N(0,1^2)$.
%
The noise level $\sigma^2$ is assumed unknown and therefore a log-normal
prior is assigned to it:
\begin{equation} \label{eq_prth}
  \theta=\log(\sigma^2) \sim \N(\mu_{\theta,0},\sigma_{\theta,0}^2)
\end{equation}
with prior mean $\mu_{\theta,0}$ and variance $\sigma_{\theta,0}^2$.

%\subsection{General properties of the model} \ref{}

% Ehkä jotain yleistä lineaarimalleista w'*x

The values of the hyperparameters $\lambda_l = 2^{-1/2} \exp( \phi_{l}/2)$
and $\sigma_{v,0}^2$ affect the smoothness properties of the model in
different ways.
In the following discussion we first assume that there is only one input
scale parameter $\lambda_l$ ($L=1$) for clarity.
Choosing a smaller value for $\lambda_l$ penalizes more strongly for larger
input weights and produces smoother functions with respect to changes in the
input features.
More precisely, in the two-layer NN model \eqref{eq_fx_mlp} the magnitudes of
the input weights (or equivalently the ARD scale parameters) are related to
the nonlinearity of the latent function $f(\x)$ with respect to the
corresponding inputs $\x$. Strong nonlinearities require large input weights
whereas almost a linear function is obtained with a very large output weight
and very small input weights for a certain hidden unit (see Section
\ref{sec_case2} for illustration).

Because the values of the activation function $g(x)$ are constrained to the
interval $[-1,1]$, hyperparameter $\sigma_{v,0}^2$ controls the overall
magnitude of the latent function $f(\x)$. Larger values of $\sigma_{v,0}^2$
increase the magnitude of the steps the hidden unit activation $v_k
g(\w_k^{\Tr} \x)$ can take in the direction of weight vector $\w_k$ in the
feature space. Choosing a smaller value for $\sigma_{v,0}^2$ can improve the
predictive performance by constraining the overall flexibility of the model
but too small value can prevent the model from explaining all the variance in
the target variable $y$.
%
%This effect can be compensated to some extent by increasing $\lambda_l$.
%
In this work, we keep $\sigma_{v,0}^2$ fixed to a sufficiently large value
and infer $\lambda_l$ from data promoting simultaneously smoother solutions
with the prior on $\phi_l= \log(2\lambda_l^2)$.
If only one common scale parameter $\phi_l$ is used, the sparsity-inducing
properties of the prior depend on the shape of the joint distribution
$p(\w|\bm{\lambda}) = \prod_{j} p(w_{j}| \lambda_l)$ resulting from the
choice of the prior terms \eqref{eq_prw}.
By decreasing $\mu_{\phi,0}$, we can favor smaller input weight values
overall, and with $\sigma_{\phi,0}^2$, we can adjust the thickness of the
tails of $p(\w|\bm{\lambda})$.
On the other hand, if individual scale parameters are assigned for all inputs
according to the ARD setting, a family of sparsity-promoting priors is
obtained by adjusting $\mu_{\phi,0}$ and $\sigma_{\phi,0}^2$. If
$\mu_{\phi,0}$ is set to a small value, say 0.01, and $\sigma_{\phi,0}^2$ is
increased, more sparse solutions are favored by allocating increasing amounts
of prior probability on the axes of the input weight space.
%
%Therefore, keeping $\sigma_{v,0}^2$ fixed and choosing a smaller $\lambda_l$
%constrains the nonlinearity of the input layer and consequently also the
%overall flexibility of the network.
%
A sparse prior could be introduced also on the output weights $v_k$ to
suppress redundant hidden units but this was not found necessary in the
experiments because the proposed EP updates have fixed point at $\E(v_k)=0$
and $\E(\w_k)=\bm{0}$ and during the iterations unused hidden units are
gradually driven towards zero (see Section \ref{sec_algorithm}).
%
%if the number of hidden units is much less than the number of observations
%
%Decreasing $\lambda_l$ increases also the sparsity promoting property of the
%joint prior, $p(\w)= \prod_{j} p(w_{j}| \lambda_l)$, which helps to control
%the confounding effects that irrelevant inputs cause to the predictions.
%
%It should be noted that the Laplace prior leads to strictly sparse estimates
%only when the MAP estimator of $\z$ is considered: It is well known that the
%posterior mean estimate is not that sparse \citep[for discussion with linear
%models see,][]{Seeger:2008}.

\section{Approximate Inference}

\begin{figure}[t]
  \centering
  \includegraphics[width=0.9\textwidth]{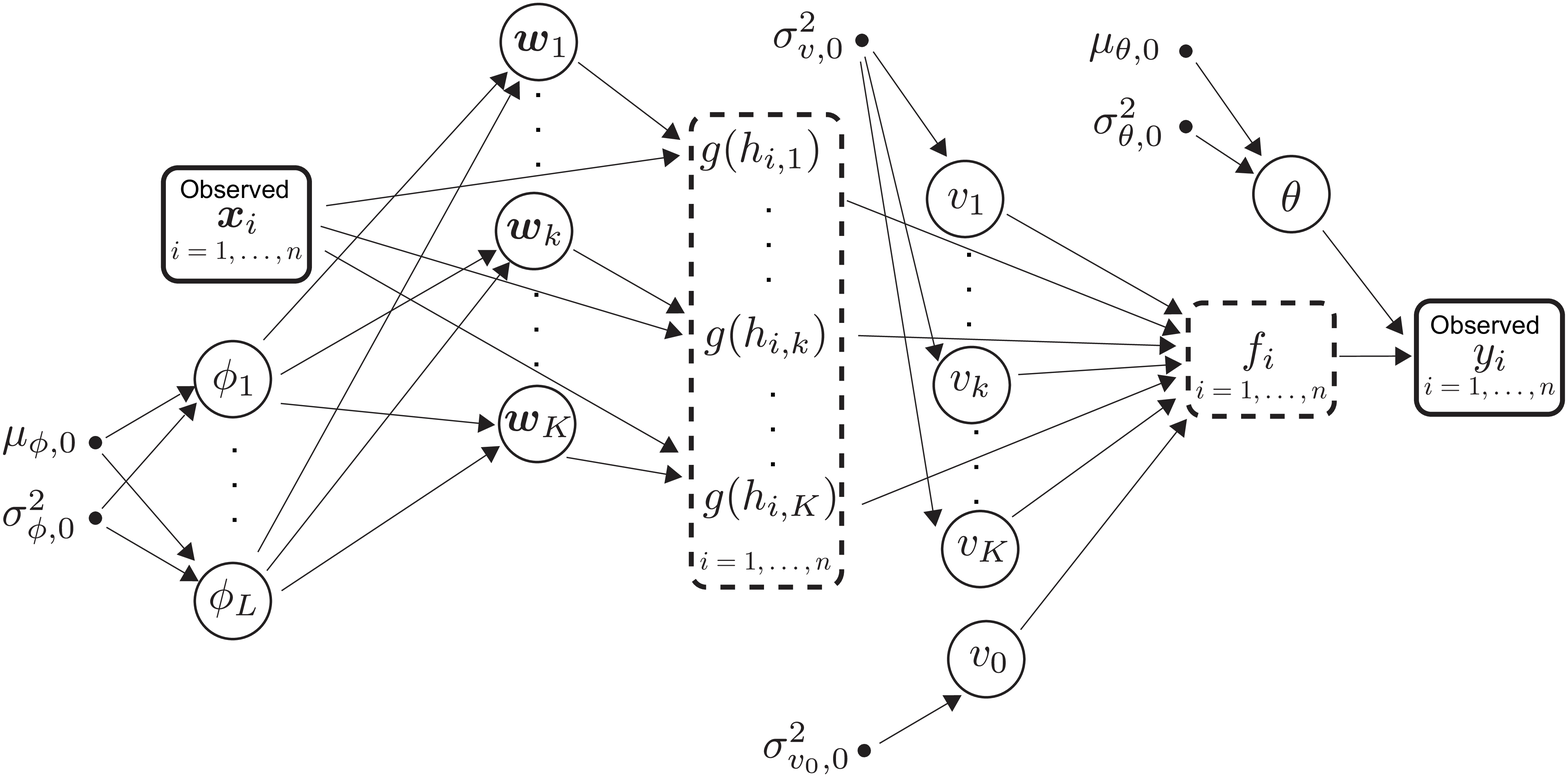}
  \caption{A directed graph representing the joint distribution of all the
  model parameters written in equation \eqref{eq_MLP_post} resulting from the
  observation model and prior definitions summarized in Section \ref{sec_model}. Observed
  variables indexed with $i=1,...,n$ are denoted with boxes, unobserved
  random variables are denoted with circles, and fixed prior parameters are
  denoted with dots. For each input $\x_i$, $i=1,...,n$, two intermediate
  random variables are visualized: The linear hidden unit activations defined
  as $h_{i,k} = \w_k^\Tr \x_i$ and the latent function value given by $f_i=
  \sum_{k=1}^K v_k g(h_{i,k}) +v_0$.} \label{fig_model_graph}
\end{figure}

In this section, we describe how approximate Bayesian inference on the
unknown model parameters $\w$, $\vv$, $\theta$, and $\bm{\phi} =[\phi_1,...,
\phi_L]^\Tr$ can be done efficiently using expectation propagation.
First, the structure of the approximation and the expressions of the
approximate site terms are presented in Section \ref{sec_approximation}. A general
description of
the EP algorithm for determining the parameters of the site approximations is
given in Section \ref{sec_ep} and approximations for the non-Gaussian hierarchial
weight priors are described in Section \ref{sec_ep_weight_priors}. The
various computational blocks required in the EP algorithm are discussed first
in Section \ref{sec_implementation} and detailed descriptions of the methods are given in
Appendices \ref{sec_cavity}--\ref{sec_site_updates}. Finally, an algorithm
description with references to the different building blocks is given in
\ref{sec_algorithm}.

%, by taking advantage of the structure of \eqref{eq_fx_mlp}.

\subsection{The Structure of the Approximation} \label{sec_approximation}

Given a set of $n$ observations $\D=\{\X,\y \}$, where
$\y=[y_1,...,y_n]^{\Tr}$, $\X=[\x_1,...,\x_n]^{\Tr}$, the posterior
distribution of all the unknowns is given by
%
% approximated as
%
%\begin{align}\label{eq_MLP_post}
%  p(\z,\theta|\D) = & Z^{-1} \prod_{i=1}^n p(y_i|f_i,\theta)
%  \prod_{kj} p(w_{kj}|\phi)
%  \prod_{k} p(v_{k}|\lambda_v)
%  p(\theta) p(\phi) \nonumber\\
%  \approx & Z_{EP}^{-1} \prod_{i=1}^n \tilde{t}_{z,i}(\z| \Nut_i,\Tt_i)
%  \tilde{t}_{\theta,i}(\theta| \mut_{\theta,i},\sigmat_{\theta,i}^2)
%  \prod_{kj} \tilde{t}_{w,kj}(w_{kj}| \mut_{w,kj},\sigmat_{w,kj}^2) \nonumber\\
%  &\cdot \prod_{k} \tilde{t}_{v,k}(v_{k}| \mut_{v,k},\sigmat_{v,k}^2)
%  p(\theta) p(\phi),
%\end{align}
%
\begin{align}\label{eq_MLP_post}
  p(\z,\theta,\bm{\phi}|\D) = & Z^{-1} \prod_{i=1}^n p(y_i|f_i,\theta)
  \prod_{j=1}^{Kd} p(w_j|\phi_{l_j})
  \prod_{k=0}^K p(v_{k}|\sigma_{\vv,0}^2) \prod_{l=1}^L p(\phi_l) p(\theta),
\end{align}
%
%\approx & Z_{EP}^{-1} \prod_{i=1}^n \tilde{t}_{z,i}(\z| \Nut_i,\Tt_i)
%\tilde{t}_{\theta,i}(\theta| \mut_{\theta,i},\sigmat_{\theta,i}^2)
%\prod_{kj} \tilde{t}_{w,kj}(w_{kj}| \mut_{w,kj},\sigmat_{w,kj}^2)
%\tilde{t}_{\phi,kj}(\phi| \mut_{\phi,kj},\sigmat_{\phi,kj}^2) \nonumber\\
%&\cdot \prod_{k} \tilde{t}_{v,k}(v_{k}| \mut_{v,k},\sigmat_{v,k}^2)
%\tilde{t}_{v,k}(v_{k}| \mut_{v,k},\sigmat_{v,k}^2)
%p(\theta) p(\phi),
%
% $\bm{\phi}=[\phi_1,...,\phi_L]^\Tr$
%
where $\sigma_{\vv,0}^2 =\{\sigma_{v,0}^2, \sigma_{v_0,0}^2\}$ and $Z=p(\y|
\X,\sigma_{\vv,0}^2)$ is the marginal likelihood of the observed data
conditioned on all fixed hyperparameters (in this case $\sigma_{\vv,0}^2$).
Figure \ref{fig_model_graph} shows a directed graph representing the joint
distribution \eqref{eq_MLP_post} where we have also included intermediate
random variables $h_{i,k}=\w_k^\Tr \x_i$ and $f_i$ related to each data point
to clarify the upcoming description of the approximate inference scheme.
%
%is approximated with $Z_{EP}$, and the non-Gaussian terms are approximated by
%Gaussian site functions ("sites" for short): $\tilde{t}_{z,i}$,
%$\tilde{t}_{\theta,i}$, $\tilde{t}_{w,kj}$, and $\tilde{t}_{v,kj}$
%
To form an analytically tractable approximation, all non-Gaussian terms are
approximated with unnormalized Gaussian site functions, which is a suitable
approximating family for random vectors defined in the real vector space.

We first consider different possibilities for approximating the likelihood
terms $p(y_i|f_i,\theta)$ which depend on the unknown noise parameter $\theta
=\log \sigma^2$ and the unknown weight vectors $\w$ and $\vv$ through the
latent function value $f_i$ according to:
\begin{equation} \label{eq_fx_vec}
  f_i = \vv^\Tr \g( \tilde{\x}_i^{\Tr} \w ) = \vv^\Tr \g(\h_i),
\end{equation}
where $\tilde{\x}_i = \I_K \otimes \x_i$ is a $Kd \times K$ auxiliary matrix
formed as Kronecker product. It can be used to write all the linear input
layer activations $\h_i$ of observation $\x_i$ as
$\h_i=\h(\x_i)=\tilde{\x}_i^{\Tr} \w$.
The vector valued function $\g(\h_i)$ applies the nonlinear transformation
\eqref{eq_gx} on each component of $\h_i$ according to
$\g(\h_i)=[g(\h_{i,1}), g(\h_{i,2}),..., g(\h_{i,K}), 1]^\Tr$ (the last
element corresponds to the output bias $v_0$). If we approximate the
posterior distribution of all the weights $\z=[\w^\Tr, \vv^\Tr]^\Tr$ with a
multivariate Gaussian approximation that is independent of all the other
unknowns including $\theta$, the resulting EP algorithm requires fast
evaluation of the means and covariances of tilted distributions of the form
\begin{equation} \label{eq_tiltedz_gen}
  \hat{p}_i(\z) \propto p(y_i|\vv^T \g(\h_i),\theta) \N(\z|\mq_\z,\Sq_\z),
\end{equation}
where $\mq_\z$ is a known mean vector, and $\Sq_\z$ a known covariance
matrix, and $\theta$ is assumed fixed. Because the non-Gaussian likelihood
term depends on $\z$ only through linear transformation $\h_i$, it can be
shown (by differentiating \eqref{eq_tiltedz_gen} twice with respect to
$\mq_\z$) that the normalization term, mean and covariance of $\hat{p}_i(\z)$
can be exactly determined by using the corresponding moments of the
transformed lower dimensional random vector $\uu_i = \B_i^\Tr \z = [\w^\Tr
\tilde{\x}_i, \vv^\Tr]^\Tr$, where the transformation matrix $\B_i$ can be
written as
\begin{equation}
  \B_i =
  \begin{bmatrix}
    \tilde{\x}_i & \mb{0} \\
    \mb{0} &  \I_{K+1} \\
  \end{bmatrix}.
\end{equation}
This results in significant computational savings because the size of $\B_i$
is $d_z \times d_u$, where we have denoted the dimensions of $\uu_i$ and $\z$
with $d_u=2K+1$ and $d_z=Kd+K+1$ respectively.
It follows that the EP algorithm can be implemented by propagating the
moments of $\uu_i$ using, for example, the general algorithm described by
\citet[][appendix C]{Cseke:2011a}. The same principle has been utilized to
form computationally efficient algorithms also for linear binary
classification \citep{Minka:2001a,Qi:2004}.

Independent Gaussian posterior approximations for both $\z$ and $\theta$ can
be obtained by approximating the likelihood terms by a product of two
unnormalized Gaussian site functions:
\begin{equation*}
  p(y_i|f_i,\theta) \approx \tilde{Z}_{y,i}
  \tilde{t}_{\z,i}(\z)\tilde{t}_{\theta,i}(\theta),
\end{equation*}
where $\tilde{Z}_{y,i}$ is a scalar scaling parameter.
Because of the previously described property, the first likelihood site
approximation depends on $\z$ only through transformation $\B_i^\Tr \z$
\citep{Cseke:2011a}:
%
%we have which means that $\z=[\w^\Tr \vv^\Tr]^\Tr$ and $\theta$ are assumed a
%posteriori independent.
%
% \tilde{Z}_{\z,i}
\begin{equation} \label{eq_tz}
  \tilde{t}_{\z,i}(\z) = \exp
  \left( -\frac{1}{2} \z^{\Tr} \B_i \Tt_i \B_i^{\Tr} \z + \z^{\Tr} \B_i \bt_i \right),
\end{equation}
where $\bt_i$ is a $d_u \times 1$ vector of location parameters, and $\Tt_i$
a $d_u \times d_u$ site precision matrix.
The second likelihood site term dependent on the scalar $\theta =\log
\sigma^2$ is chosen to be an unnormalized Gaussian
\begin{equation} \label{eq_ttheta}
  \tilde{t}_{\theta,i}(\theta)
  = \exp \left(
  -\frac{1}{2} \sigmat_{\theta,i}^{-2} \theta^2
  + \mut_{\theta,i} \sigmat_{\theta,i}^{-2} \theta \right)
  \propto \N(\theta| \mut_{\theta,i},\sigmat_{\theta,i}^2),
\end{equation}
%= \tilde{Z}_{\theta,i} \N(\theta| \mut_{\theta,i},\sigmat_{\theta,i}^2),
%
where the site parameters $\mut_{\theta,i}$ and $\sigmat_{\theta,i}^2$
control the location and the scale of site function, respectively. Combined
with the known Gaussian prior term on $\theta$ this results in a Gaussian
posterior approximation for $\theta$ that corresponds to a log-normal
approximation for $\sigma^2$.

The prior terms of the output weights $v_k$, for $k=1,...,K$, are
approximated with
\begin{equation} \label{eq_tvk}
  p(v_k| \sigma_{v,0}^2) \approx \tilde{Z}_{v,k} \tilde{t}_{v,k}(v_k)
  \propto \N(v_k| \mut_{v,k},\sigmat_{v,k}^2),
\end{equation}
%
%\begin{equation} \label{eq_tvk}
%  p(v_k| \sigma_{v,0}^2) \approx \tilde{t}_{v,k}(v_k)
%  = \tilde{Z}_{v,k} \N(v_k| \mut_{v,k},\sigmat_{v,k}^2),
%\end{equation}
%
where $\tilde{Z}_{v,k}$ is a scalar scaling parameter, $\tilde{t}_{v,k}(v_k)$
has a similar exponential form as \eqref{eq_ttheta}, and the site parameters
$\mut_{v,k}$ and $\sigmat_{v,k}^2$ control the location and scale of the
site, respectively.
If the prior scales $\phi_l$ are assumed unknown, the prior terms of the
input weights $\{ w_j| j=1,...,Kd\}$, are approximated with
\begin{equation} \label{eq_twk}
  p(w_j| \phi_{l_j} )
  \approx \tilde{Z}_{w,j} \tilde{t}_{w,j}(w_j) \tilde{t}_{\phi,j}(\phi_{l_j})
  \propto \N(w_j| \mut_{w,j},\sigmat_{w,j}^2)
  \N(\phi_{l_j} | \mut_{\phi,j},\sigmat_{\phi,j}^2),
\end{equation}
%
%\begin{equation} \label{eq_twk}
%  p(w_j| \phi_{l_j} )
%  \approx \tilde{t}_{w,j}(w_j) \tilde{t}_{\phi,j}(\phi_{l_j})
%  = \tilde{Z}_{w,j} \N(w_j| \mut_{w,j},\sigmat_{w,j}^2)
%  \N(\phi_{l_j} | \mut_{\phi,j},\sigmat_{\phi,j}^2),
%\end{equation}
%
where a factorized site approximation with location parameters $\mut_{w,j}$
and $\mut_{\phi,j}$, and scale parameters $\sigmat_{w,j}^2$ and
$\sigmat_{\phi,j}^2$, is assumed for weight $w_j$ and the associated scale
parameter $\phi_{l_j}$, respectively. A similar exponential form to equation
\eqref{eq_ttheta} is assumed for both $\tilde{t}_{w,j}(w_j)$ and
$\tilde{t}_{\phi,j}(\phi_{l_j})$. This factorizing site approximation results
in independent posterior approximations for $\w$ and each component of
$\bm{\phi}$ as will be described shortly.

The actual numerical values of the normalization parameters
$\tilde{Z}_{y,i}$, $\tilde{Z}_{v,k}$, and $\tilde{Z}_{w,j}$ are not required
during the iterations of the EP algorithm but their effect must be taken into
account if one wishes to form an EP approximation for the marginal likelihood
$Z=p(\y|\X)$ (see Appendix \ref{sec_marg_likelih}). This estimate could be
utilized to compare models or to alternatively determine
type-II MAP estimates for hyperparameters $\theta$, $\{ \phi_l \}_{l=1}^L$,
and $\sigma_{v,0}^2$ in case they are not inferred within the EP framework.

\subsubsection{Fully-coupled approximation for the network weights}

Multiplying the site approximations together according to \eqref{eq_MLP_post}
and normalizing the resulting expression gives the following Gaussian
posterior approximation:
\begin{equation} \label{eq_qep}
  p(\z,\theta,\bm{\phi}|\D,\sigma_{\vv,0}^2) \approx q(\z) q(\theta) \prod_{l=1}^L q(\phi_l),
\end{equation}
where $q(\z)=\N(\z|\mq,\Sq)$, $q(\theta)=\N(\theta | \mu_{\theta}^2,
\sigma_{\theta}^2 )$, and $q(\phi_l) = \N(\phi_l | \mu_{\phi_l}^2,
\sigma_{\phi_l}^2 )$ are the approximate posterior distributions of the
weights $\z$, the noise parameter $\theta=\log \sigma^2$, and the input
weight scale parameter $\phi_l$, respectively. The mean vector and covariance
matrix of $q(\z)$, are given by
\begin{equation} \label{eq_qep_z}
  \Sq =\left( \sum_i^n \B_i \Tt_i \B_i^{\Tr} + \Sq_0^{-1} \right)^{-1}
  \quad \textrm{and} \quad
  \mq = \Sq \left( \sum_i^n \B_i \bt_i + \Sq_0^{-1} \mq_0 \right),
\end{equation}
where the parameters of the prior term approximations \eqref{eq_tvk} and
\eqref{eq_twk} are collected together in $\Sq_0=\text{diag}([ \sigmat_{w,1}^2
,..., \sigmat_{w,Kd}^2,\sigmat_{v,1}^2,...,\sigmat_{v,K}^2 ])$ and
$\mq_0=[\mut_{w,1}, ..., \mut_{w,Kd}, \mut_{v,1},..., \mut_{v,K}]^{\Tr}$.
The parameters of $q(\theta)$ are given by
\begin{eqnarray} \label{eq_qep_theta}
  \sigma_{\theta}^2 = \left( \sum_i^n \sigmat_{\theta,i}^{-2}
  + \sigma_{\theta,0}^{-2} \right)^{-1}
  \quad \textrm{and} \quad
  \mu_{\theta} = \sigma_{\theta,0}^2 \left( \sum_i^n \sigmat_{\theta,i} ^{-2}
  \mut_{\theta,i} +\sigma_{\theta,0}^{-2} \mu_{\theta,0} \right).
\end{eqnarray}
The approximate mean and variance of $q(\phi_l)$ can be computed analogously
to \eqref{eq_qep_theta}.
The key property of the approximation \eqref{eq_qep} is that if we can
incorporate the information provided by the data point $y_i$ in the
parameters $\Tt_i$ and $\bt_i$, for all $i=1,\ldots,n$, the approximate
inference on the non-Gaussian priors $p(v_k)$ and $p(w_{kj})$ is
straightforward by adopting the methods described by \citep{Seeger:2008}. In
the following sections we will show how this can be done by approximating
%
%$p(f_i,\h_i,\vv|y_1,...,y_{i-1},y_{i+1},...,y_n)$
%
the joint distribution of $f_i$, $\h_i$ and $\vv$, given $\y_{-i}
=[y_1,...,y_{i-1},y_{i+1},...,y_n]$, with a multivariate Gaussian and using
it to estimate the parameters $\Tt_i$ and $\bt_i$ one data point at a time
within the EP framework.

\subsubsection{Factorizing approximation for the network weights}

A drawback with the approximation \eqref{eq_qep_z} is that the evaluation of
the covariance matrix $\Sq$ scales as $\mathcal{O}( \min (Kd+K+1,n)^3 )$
which may not be feasible when the number of inputs $d$ is large.
Another difficulty arises in determining the mean and covariance of $\uu_i =
\B_i \z = [\h_i^\Tr, \vv^\Tr]^\Tr$ when $\z$ is distributed according to
\eqref{eq_tiltedz_gen} because during the EP iterations $\Sq_\z$ has similar
correlation structure with $\Sq$. If the observation model is Gaussian and
$\theta$ is fixed, this requires at least $K$-dimensional numerical
quadratures (or other alternative approximations) that may quickly become
infeasible as $K$ increases.
By adopting suitable independence assumptions between $\vv$ and the input
weights $\w_k$ associated with the different hidden units, the mean and
covariance of $\uu_i$ can be approximated using only 1-dimensional numerical
quadratures as will be described in Section \ref{sec_implementation}.

The structure of the correlations in the approximation $\eqref{eq_qep_z}$ can
be studied by dividing $\Tt_i$ into four blocks as follows:
\begin{equation}
  \Tt_i = \left[
         \begin{array}{cc}
           \Tt_{\h_i \h_i} & \Tt_{\h_i \vv} \\
           \Tt_{\h_i \w} & \Tt_{i,\vv\vv} \\
         \end{array}
       \right],
\end{equation}
where $\Tt_{\h_i \h_i}$ is a $K \times K$ matrix, $\Tt_{\h_i \vv}$ a $K
\times K+1$ matrix, and $\Tt_{i,\vv \vv}$ a $K+1 \times K+1$ matrix.
The element $[\Tt_{\h_i \h_i}]_{k,k'}$ contributes to the approximate
posterior covariance between $\w_k$ and $\w_{k'}$, and the sub-matrix
$\Tt_{\h_i \vv}$ contributes to the approximate covariance between $\w$ and
$\vv$. To form an alternative computationally more efficient approximation we
propose a simpler structure for $\Tt_i$.
First, we approximate $\Tt_{\h_i \h_i}$ with a diagonal matrix, that is,
$\Tt_{\h_i \h_i} = \diag(\Tauwl_i)$, where only the $k$:th component of the
vector $\Tauwl_i$ contributes to the posterior covariance of $\w_k$.
Secondly, we set $\Tt_{\h_i \vv}=\bm{0}$ and approximate $\Tt_{i,\vv \vv}$
with an outer-product of the form $\Tt_{i,\vv \vv} = \Tauvl_i \Tauvl_i^\Tr$.
With this precision structure the site approximation \eqref{eq_tz} can be
factorized into terms depending only on the output weights $\vv$ or the input
weights $\w_k$ associated with the different hidden units $k=1,...,K$:
%
% \tilde{Z}_{\z,i}
\begin{align} \label{eq_tz_fact}
  \tilde{t}_{\z,i}(\z) &=
  \exp \left( -\frac{1}{2} (\Tauvl_i^{\Tr} \vv)^2 + \vv^{\Tr} \Nuvl_i \right)
  \prod_{k=1}^K
  \exp \left( -\frac{1}{2} \tauwl_{i,k} (\x_i^\Tr \w_k)^2 +
  \nuwl_{i,k} \w_k^{\Tr} \x_i \right) \\ \nonumber
  &= \tilde{t}_{\vv,i}(\vv) \prod_{k=1}^K \tilde{t}_{\w_k,i}(\w_k),
\end{align}
where the site location parameters $\nuwl_{i,k}$ now correspond to the first
$K$ elements of $\bt_i$ in equation \eqref{eq_tz}, that is, $\Nuwl_i
=[\nuwl_{i,1},...\nuwl_{i,K}]^\Tr =[\bt_{i,1},...,\bt_{i,K}]^\Tr$.
Analogously, the site location vector $\Nuvl_i$ corresponds to the last $K+1$
entries of $\bt_i$, that is, $\Nuvl_i=[\bt_{i,K+1},...,\bt_{i,2K+1}]^\Tr$.

The factored site approximation \eqref{eq_tz_fact} results in independent
posterior approximations for the output weights $\vv$ and the input weights
$\w_k$ associated with different hidden units:
\begin{equation} \label{eq_qep_fact}
  q(\z) = q(\vv) \prod_{k=1}^K q(\w_k),
\end{equation}
where $q(\vv) =\N(\mq_v,\Sq_v)$ and $q(\w_k) =\N(\mq_{\w_k},\Sq_{\w_k})$. The
approximate mean and covariance of $\w_k$ is given by
\begin{equation} \label{eq_qep_w}
  \Sq_{\w_k} =\left( \X^\Tr \Tt_{\w_k} \X + \Sq_{\w_k,0}^{-1} \right)^{-1}
  \quad \textrm{and} \quad
  \mq_{\w_k} = \Sq_{\w_k} \left( \X^\Tr \Nuwl_{\w_k} +
  \Sq_{\w_k,0}^{-1} \mq_{\w_k,0} \right),
\end{equation}
where the diagonal matrix $\Tt_{\w_k} = \diag(\Taut_{\w_k})$ and the vector
$\Nut_{\w_k}$ collect all the site parameters related to hidden unit $k$:
$\Tauwl_{\w_k}= [\tauwl_{1,k},...,\tauwl_{n,k}]^\Tr$ and $\Nuwl_{\w_k}=
[\nuwl_{1,k},...,\nuwl_{n,k}]^\Tr$.
%
%where the site location parameters $\nut_{i,k}$ related to observation $i$
%now correspond to the first $K$ elements of $\bt_i$: $\Nut_i
%=[\nut_{i,1},...\nut_{i,K}] =[\bt_{i,1},...,\bt_{i,K}]^\Tr$.
%
The diagonal matrix $\Sq_{\w_k,0}$ and the vector $\mq_{\w_k,0}$ contain the
prior site scales $\sigmat_{w,j}^2$ and the location variables $\mut_{w,j}$
related to the hidden unit $k$.
The approximate mean and covariance of the output weights $\vv$ are given by
\begin{equation} \label{eq_qep_v}
  \Sq_{\vv} =\left(\sum_{i=1}^{n} \Tauvl_i \Tauvl_i^\Tr + \Sq_{\vv,0}^{-1} \right)^{-1}
  \quad \textrm{and} \quad
  \mq_{\vv} = \Sq_{\vv} \left( \sum_{i=1}^{n} \Nuvl_i
  +\Sq_{\vv}^{-1} \mq_{\vv,0} \right),
\end{equation}
%
%where the site location vector $\Nuvl_i$ now corresponds to the last $K+1$
%entries of $\bt_i$ in equation \eqref{eq_tz}, that is,
%$\Nuvl_i=[\bt_{i,K+1},...,\bt_{i,2K+1}]^\Tr$.
%
where the diagonal matrix $\Sq_{\vv,0}$ and the vector $\mq_{\vv,0}$ contain
the prior site scales $\sigmat_{v,j}^2$ and the location variables
$\mut_{v,j}$.

For each hidden unit $k$, approximations \eqref{eq_qep_fact} and
\eqref{eq_qep_w} can be interpreted as independent linear models with
Gaussian likelihood terms $\N(\tilde{y}_{i,k} | \x_i^\Tr
\w_k,\tauwl_{i,k}^{-1})$, where the pseudo-observations are given by
$\tilde{y}_{i,k}=\tauwl_{i,k}^{-1} \nuwl_{i,k}$. The approximation for the
output weights \eqref{eq_qep_v} has no explicit dependence on the input
vectors $\x_i$ but later we will show that the independence assumption
results in a similar dependence on expected values of $\g_i$ taken with
respect to approximate leave-one-out (LOO)
distributions of $\w$ and $\vv$. %given $\y_{-i}$.

%Our strategy is to approximate the joint distribution of $f_i$, $\h_i$ and
%$\vv$, given $y_1,...,y_{i-1},y_{i+1},...,y_n$, by a multivariate Gaussian
%and use it to estimate the parameters $\Tt_i$ and $\Nut_i$ one data point at
%time within the EP framework as will be described in the following sections.

%%%%%%%ADD ONE SENTENCE WHAT IS OUR OVERALL
%%%%%%%%STRATEGY FOR INCORPORATING THE INFORMATION
%%%%%%%%%PROVIDED BY THE DATA POINT

%  \approx \N(\mq,\Sq) \N(\theta|\mut_{\theta},\sigmat_{\theta}) \B_i

\subsection{Expectation Propagation} \label{sec_ep}

The parameters of the approximate posterior distribution \eqref{eq_qep} are
determined using the EP algorithm \citep{Minka:2001b}. The EP algorithm
updates the site parameters
%$\tilde{Z}=\{ \tilde{Z}_{z,i}, \tilde{Z}_{\theta,i}, \tilde{Z}_{\theta,i} \}$,
%$\mut_i$ and $\sigmat_i^2$
and the posterior approximation $q(\z,\theta,\bm{\phi})$ sequentially. In the
following, we briefly describe the procedure for updating the likelihood
sites $\tilde{t}_{\z,i}$ and $\tilde{t}_{\theta,i}$ and assume that the prior
sites \eqref{eq_tvk} and \eqref{eq_twk} are kept fixed. Because the
likelihood terms $p(y_i|f_i,\theta)$ do not depend on $\bm{\phi}$ and
posterior approximation is factorized, that is $q(\z,\theta,\bm{\phi}) =q(\z)
q(\theta) q(\bm{\phi})$, we need to consider only the approximations for $\z$
and $\theta$. Furthermore, independent approximations for $\w_k$ and $\vv$
are obtained by using \eqref{eq_tz_fact} and \eqref{eq_qep_fact} in place
$\tilde{t}_{\z,i}$ and $q(\z)$, respectively.

At each iteration, first a proportion $\eta$ of the $i$:th site term is
removed from the posterior approximation to obtain a cavity distribution:
\begin{equation}\label{eq_EP_cavity}
  q_{-i}(\z,\theta) = q_{-i}(\z)q_{-i}(\theta)
  \propto q(\z) q(\theta)
  \tilde{t}_{\z,i}(\z)^{-\eta}
  \tilde{t}_{\theta,i}(\theta)^{-\eta},
\end{equation}
where $\eta \in (0,1]$ is a fraction parameter that can be adjusted to
implement fractional (or power) EP updates \citep[][]{Minka:2004,Minka:2005}.
When $\eta=1$, the cavity distribution \eqref{eq_EP_cavity} can be thought of
as a LOO posterior approximation where the contribution of the $i$:th
likelihood term $p(y_i|f_i,\theta)$ is removed from $q(\z,\theta)$.
Then, the $i$:th site is replaced with the exact likelihood term to form a
tilted distribution
\begin{equation}\label{eq_EP_tilted}
\hat{p}_i (\z,\theta) = \hat{Z}_i^{-1} q_{-i}(\z,\theta)p(y_i|\z,\theta,\x)^\eta,
\end{equation}
where $\hat{Z}_i$ is a normalization constant, which in this case can also be
thought of as an approximation for the LOO predictive density of the excluded
data point $y_i$. The tilted distribution can be regarded as a more refined
non-Gaussian approximation to the true posterior distribution.
Next, the algorithm attempts to match the approximate posterior distribution
$q(\z,\theta)$ with $\hat{p}_i(\z,\theta)$ by finding first a Gaussian
$\hat{q}_i(\z,\theta)$ that satisfies
\begin{equation*}
  \hat{q}_i(\z,\theta) = \argmin_{q_i}
  \KL \left(\hat{p}_i(\z,\theta)||q_i(\z,\theta)\right),
\end{equation*}
%% \N(\theta| \hat{\mq}_i, \hat{\Sq}_i^2)
%
%$\hat{q}_i(\z,\theta) = \argmin_{q_i} \KL\left(\hat{p}_i(\z,\theta)||
%q_i(\z,\theta)\right)$,
%
where KL denotes the Kullback-Leibler divergence. When $q(\z,\theta)$ is
chosen to be a Gaussian distribution this is equivalent to determining the
mean vector and the covariance matrix of $\hat{p}_i$ and matching them with
the mean and covariance of $\hat{q}_i$.
Then, the parameters of the $i$:th site terms are updated so that the moments
of $q(\z,\theta)$ match with $\hat{q}(\z,\theta)$:
\begin{equation} \label{eq_EP_moment_matching}
  \hat{q}_i(\z,\theta) \equiv
  q(\z,\theta) \propto q_{-i}(\z) q_{-i}(\theta) \tilde{t}_{\z,i}(\z)^\eta
  \tilde{t}_{\theta,i}(\theta)^\eta.
\end{equation}
Finally, the posterior approximation $q(\z,\theta)$ is updated according to
the changes in the site parameters. These steps are repeated for all sites in
some suitable order until convergence.

From now on, we refer to the previously described EP update scheme as
sequential EP. If the update of the posterior approximation $q(\z,\theta)$ in
the last step is done only after new parameter values have been determined
for all sites (in this case the $n$ likelihood term approximations), we refer
to parallel EP \citep[see, e.g.,][]{Gerven:2009}. Because in our case the
approximating family is Gaussian and each likelihood term depends on a linear
transformation of $\z$, one sequential EP iteration requires (for each of the
$n$ sites) either one rank-$(2K+1)$ covariance matrix update with the
fully-correlated approximation \eqref{eq_qep_z}, or $K+1$ rank-one covariance
matrix updates with the factorized approximation (\ref{eq_qep_w},
\ref{eq_qep_v}). In parallel EP these updates are replaced with a single
re-computation of the posterior representation after each sweep over all the
$n$ sites. In practice, one parallel iteration step over all the sites can be
much faster compared to a sequential EP iteration, especially if $d$ or $K$
are large, but parallel EP may require larger number of iterations for
overall convergence.

Setting the fraction parameter to $\eta=1$ corresponds to regular EP updates
whereas choosing a smaller value produces a slightly different approximation
that puts less emphasis on preserving all the nonzero probability mass of the
tilted distributions \citep{Minka:2005}. Consequently, setting $\eta<1$ tries
to represent possible multimodalities in \eqref{eq_EP_tilted} but ignores
modes far away from the main probability mass, which results in tendency to
underestimate variances. However, in practice decreasing $\eta$ can improve
the overall numerical stability of the algorithm and alleviate convergence
problems resulting from possible multimodalities in case the unimodal
approximation is not a good fit for the true posterior distribution
\citep{Minka:2005, Seeger:2008, Jylanki:2011}.

There is no theoretical convergence guarantee for the standard EP algorithm
but damping the site parameter updates can help to achieve convergence in
harder problems \citep{Minka:2002,Heskes:2002}.\footnote{Alternative provably
convergent double-loop algorithms exist but usually they require more
computational effort in the form of additional inner-loop iterations
\citep{Minka:2001b, Heskes:2002, Opper:2005, Seeger:2011}.}
In damping, the site parameters are updated to a convex combination of the
old values and the new values resulting from \eqref{eq_EP_moment_matching}
defined by a damping factor $\delta \in (0,1]$. For example, the precision
parameter of the likelihood site term $\tilde{t}_{\w_k,i}$ is updated as
$\tauwl_{i,k} = (1-\delta) \tauwl_{i,k}^{\text{old}} + \delta
\tauwl_{i,k}^{\text{new}}$ and a similar update using the same $\delta$-value
is done on the corresponding location parameter $\nuwl_{i,k}$.
The convergence problems are usually seen as oscillations over iterations in
the site parameter values and they may occur, for example, if there are
inaccuracies in the tilted moment evaluations, or if the approximate
distribution is not a suitable proxy for the true posterior, for example, due
to multimodalities.

\subsubsection{EP approximation for the weight prior terms}
\label{sec_ep_weight_priors}

Assuming fixed site parameters for the likelihood approximation
\eqref{eq_tz_fact}, or \eqref{eq_tz} in the case of full couplings, the EP
algorithm for determining the prior term approximations \eqref{eq_tvk} and
\eqref{eq_twk} can be implemented in the same way as with sparse linear
models \citep{Seeger:2008}.

To derive EP updates for the non-Gaussian prior sites of the output weights
$\vv$ assuming the factorized approximation, we need to consider only the
prior site approximations \eqref{eq_tvk} and the approximate posterior
$q(\vv) = \N(\vv| \mq_{\vv}, \Sq_{\vv})$ defined in equation
\eqref{eq_qep_v}. All approximate posterior information from the observations
$\D =\{\y,\X \}$ and the priors on the input weights $\w$ are conveyed in the
parameters $\{\Tauvl_i, \Nuvl_i \}_{i=1}^n$ determined during the EP
iterations for the likelihood sites. The EP implementation of
\citet{Seeger:2008} can be readily applied by using $\sum_i^n \Tauvl_i
\tauvl_i^{\Tr}$ and $\sum_i^n \Nuvl_i$ as a Gaussian pseudo likelihood as
discussed in Appendix \ref{sec_site_updates}.
Because the prior terms $p(v_k|\sigma_{v,0}^2)$ depend only on one random
variable $v_k$, deriving the parameters of the cavity distributions
$q_{-k}(v_k) \propto q(v_k) \tilde{t}_{v,k} (v_k| \mut_{v,k}, \sigmat_{v,k}^2
)^{-\eta}$ and updates for the site parameters $\mut_{v,k}$ and
$\sigmat_{v,k}^2$ require only manipulating univariate Gaussians.
The moments of the tilted distribution $\hat{p}_k(v_k) \propto q_{-k}(v_k)
p(v_k| \sigma_{v,0}^2)^\eta$ can be computed either analytically or using a
one-dimensional numerical quadrature depending on the functional form of the
exact prior term $p(v_k|\sigma_{v,0}^2)$.

To derive EP updates for the non-Gaussian hierarchical prior sites of the
input weights $\w$ assuming the factorized approximation \eqref{eq_qep_fact},
we can consider the approximate posterior distributions $q(\w_k) = \N(\w_k|
\mq_{\w_k}, \Sq_{\w_k})$ from equation \eqref{eq_qep_w} separately with the
corresponding prior site approximations \eqref{eq_twk} related to the $d$
components of $\w_k$.
All approximate posterior information from the observations $\y$ is conveyed
by the site parameters $\{ \Tauwl_{\w_k}, \Nuwl_{\w_k} \}$ that together with
the input features form a Gaussian pseudo likelihood with a precision matrix
$\X^\Tr \diag(\Tauwl_{\w_k}) \X$ and location term $\X^\Tr \Nuwl_{\w_k}$
(compare with equation \ref{eq_qep_w}).
It follows that the EP implementation of \citet{Seeger:2008} can be applied
to update the approximations $q(\w_k)$ but, in addition, we have to derive
site updates also for the scale parameter approximations $q(\phi_{l_j})$.
EP algorithms for sparse linear models that operate on exact site terms
depending on a nonlinear combination of multiple random variables have been
proposed by \citet{hernandez:2008} and \citet{Gerven:2009}.

Because the $j$:th exact prior term \eqref{eq_prw} depends on both the weight
$w_j$ and the corresponding log-transformed scale parameter $\phi_{l_j}$, the
$j$:th cavity distribution is formed by removing a fraction $\eta$ of both
site approximations $\tilde{t}_{w,j}(w_j)$ and
$\tilde{t}_{\phi,j}(\phi_{l_j})$:
\begin{equation} \label{eq_cavity_pw}
  q_{-j}(w_j, \phi_{l_j} ) = q_{-j}(w_j) q_{-j}(\phi_{l_j} ) \propto
  q(w_j) q(\phi_{l_j} ) \tilde{t}_{w,j}(w_j)^{-\eta}
  \tilde{t}_{\phi,j}(\phi_{l_j})^{-\eta},
\end{equation}
where $q(w_j)$ is the $j$:th marginal approximation extracted from the
corresponding approximation $q(\w_k)$, and the approximate posterior for
$\phi_{l_j}$ is formed by combining the prior \eqref{eq_pruw} with all the
site terms $\tilde{t}_{\phi,i}(\phi_{l_i})$ that depend on $\phi_{l_i}$:
\begin{align*}
  q(\phi_{l_j}) \propto p(\phi_{l_j}) \prod_{i=1,l_i = l_j}^{Kd}
  \tilde{t}_{\phi,i}(\phi_{l_i}).
\end{align*}
The $j$:th tilted distribution is formed by replacing the removed site terms
with a fraction $\eta$ of the exact prior term $p(w_j| \phi_{l_j})$:
\begin{equation} \label{eq_tilted_pw}
  \hat{p}_j (w_j, \phi_{l_j} )
  = \hat{Z}_{w_j}^{-1} q_{-j}(w_j) q_{-j}(\phi_{l_j} ) p(w_j| \phi_{l_j})^\eta
  \equiv \hat{q} (w_j, \phi_{l_j} ),
\end{equation}
where $\hat{q} (w_j, \phi_{l_j} )$ is a Gaussian approximation formed by
determining the mean and covariance of $\hat{p}_j (w_j, \phi_{l_j} )$.
The site parameters are updated so that the resulting posterior approximation
is consistent with the marginal means and variances of $\hat{q} (w_j,
\phi_{l_j} )$:
\begin{equation} \label{eq_mm_pw}
  \hat{q}_j (w_j) \hat{q}_j (\phi_{l_j} ) =
  q_{-j}(w_j) q_{-j}(\phi_{l_j} )
  \tilde{t}_{w,j}(w_j)^{\eta} \tilde{t}_{\phi,j}(\phi_{l_j})^{\eta}.
\end{equation}
Because of the factorized approximation, the cavity computations
\eqref{eq_cavity_pw} and the site updates \eqref{eq_tilted_pw} require only
scalar operations similar to the evaluations of $q_{-i}(h_{i,k})$ and to the
updates of $\{\Tauwl_i, \Nuwl_i \}$ in equations \eqref{eq_cavity_w} and
\eqref{eq_sites_w} respectively (see Appendix \ref{sec_cavity} and
\ref{sec_site_updates}).

Determining the moments of \eqref{eq_tilted_pw} can be done efficiently using
one-dimensional quadratures if the means and variances of $w_j$ with respect
to the conditional distribution $\hat{p}_j (w_j| \phi_{l_j} )$ can be
determined analytically. This can be readily done when $p(w_j|\phi_{l_j})$ is
a Laplace distribution or a finite mixture of Gaussians. Note also that if we
wish to implement an ARD prior we can choose simply $p(w_j|\phi_{l_j}) =
\N(w_j|0, \phi_{l_j})$, where $\phi_{l_j}$ is a common scale parameter for
all weights related to the same input feature, that is, weights $\{ w_j,
w_{j+d},..., w_{j+(K-1)d} \}$, for each $j \in \{1,2,...,d\}$, share the same
scale $\phi_j$.
The marginal tilted distribution for $\phi_{l_j}$ is given by
\begin{align} \label{eq_qphi_tilted}
   \hat p(\phi_{l_j}) &= \hat{Z}_{w_j}^{-1}
   \int q_{-j}(w_j) q_{-j}(\phi_{l_j} ) p(w_j| \phi_{l_j})^\eta dw_j
   = \hat{Z}_{w_j}^{-1} Z(\phi_{l_j},\eta) q_{-j}(\phi_{l_j})
   \nonumber\\
   &\approx \N(\phi_{l_j}| \hat{\mu}_{\phi, l_j}, \hat{\sigma}_{\phi, l_j}^2),
\end{align}
where it is assumed that $Z(\phi_{l_j},\eta) = \int q_{-j}(w_j) p(w_j|
\phi_{l_j})^\eta dw_j$ can be calculated analytically. The normalization term
$\hat{Z}_{w_j}^{-1}$, the marginal mean $\hat{\mu}_{\phi, l_j}$, and the
variance $\hat{\sigma}_{\phi, l_j}^2$ can be determined using numerical
quadratures.

The marginal tilted mean and variance of $w_j$ can be determined by
integrating numerically the conditional expectations of $w_j$ and $w_j^2$
over $\hat{p}_j(\phi_{l_j})$:
\begin{align} \label{eq_qwj_tilted}
   \E (w_j) &= \hat{Z}_{w_j}^{-1}
   \int w_j \hat{p}_j(w_j| \phi_{l_j}) Z(\phi_{l_j},\eta) q_{-j}(\phi_{l_j} )
   dw_j d\phi_{l_j}
   = \int \E(w_j|\phi_{l_j},\eta) \hat{p}_j(\phi_{l_j}) d\phi_{l_j} \nonumber\\
   \var (w_j)
   &= \int \E(w_j^2|\phi_{l_j},\eta) \hat{p}_j(\phi_{l_j}) d\phi_{l_j} -\E(w_j)^2,
\end{align}
where $\hat{p}_j(w_j| \phi_{l_j}) = Z(\phi_{l_j},\eta)^{-1} q_{-j}(w_j)
p(w_j| \phi_{l_j})^\eta$, and it is assumed that the conditional expectations
$\E(w_j|\phi_{l_j},\eta)$ and $\E(w_j^2|\phi_{l_j},\eta)$ can be calculated
analytically. For prior distributions $p(w_j|\phi_{l_j})$ belonging to the
exponential family, the exponentiation with $\eta$ results in a distribution
of the same family multiplied by a function of $\eta$ and $\phi_{l_j}$.
%
%For example, with a Gaussian prior $p(w_j|\phi_{l_j})
%=\N(w_j|0,\exp(\phi_{l_j}))$
%
%we get $p(w_j|\phi_{l_j})^\eta =\N(w_j|0,\exp(\phi_{l_j})/\eta)$ and
%
%$Z(\phi_{l_j},\eta) = \N(\mu_{-j}|0, \sigma_{-j}^2 + \eta^{-1}
%\exp(\phi_{l_j}) ) (2\pi \exp(\phi_{l_j}))^{(1-\eta)/2} \eta^{-1/2}$,
%
%where $q_{-j}(w_j) = \N(\mu_{-j},\sigma_{-j}^2)$.
%
%$(2\pi \exp(\phi_{l_j}))^{(1-\eta)/2} \eta^{-1/2}$
%
Evaluating the marginal moments according to equations \eqref{eq_qphi_tilted}
and \eqref{eq_qwj_tilted} requires a total of five one-dimensional quadrature
integrations but in practice this can be done efficiently by utilizing the
same function evaluations of $\hat{p}_j(\phi_{l_j})$ and taking into account
the prior specific forms of $\E(w_j|\phi_{l_j},\eta)$ and
$\E(w_j^2|\phi_{l_j},\eta)$.

% the conditional normalization constant

%
%\begin{equation*} \label{eq_qc_w}
%  p(w_j| \phi_{l_j} )
%  \approx \tilde{Z}_{w,j} \tilde{t}_{w,j}(w_j) \tilde{t}_{\phi,j}(\phi_{l_j})
%  \propto \N(w_j| \mut_{w,j},\sigmat_{w,j}^2)
%  \N(\phi_{l_j} | \mut_{\phi,j},\sigmat_{\phi,j}^2),
%\end{equation*}

\subsection{Implementing the EP Algorithm}
\label{sec_implementation}

In this section, we describe the computational implementation of the EP
algorithm resulting from the choice of the approximating family described in
Section \ref{sec_approximation}. Because the non-Gaussian likelihood term in
the tilted distribution \eqref{eq_EP_tilted} depends on $\z = [\w^\Tr,
\vv^\Tr]^\Tr$ only through the linear transformation $\uu_i = [\h_i^\Tr,
\vv^\Tr]^\Tr = \B_i^\Tr \z$, the EP algorithm can be more efficiently
implemented by iteratively determining and matching the moments of the
lower-dimensional random vector $\uu_i$ instead of $\z$ \citep[][appendix
C]{Cseke:2011a}.
The computations can be further facilitated by using the factorized
approximation \eqref{eq_qep_fact}: Because the hidden activations $h_{i,k} =
\x_i^\Tr \w_k$ related to different hidden units $k=1,...,K$ are independent
of each other and $\vv$, it is only required to propagate the marginal means
and covariances of $h_{i,k}$ and $\vv$ to determine the new site parameters.
%
%also the cavity distribution resulting from \eqref{eq_EP_cavity} can be
%factored similarly.
%
This results in computationally more efficient determination of the cavity
distributions \eqref{eq_EP_cavity}, the tilted distributions
\eqref{eq_EP_tilted}, and the new site parameter from
\eqref{eq_EP_moment_matching}. Details of the computations required for
updating the likelihood site approximations are presented in Appendices
\ref{sec_cavity}--\ref{sec_site_updates}. The main properties of the
procedure can be summarized as:
\begin{itemize}
  \item Appendix \ref{sec_cavity} presents the formulas for computing the
      parameters of the cavity distributions \eqref{eq_EP_cavity}. The
      factorized approximation \eqref{eq_qep_fact} leads to efficient
      computations, because the cavity distribution can be factored as
      $q_{-i}(\z) = q_{-i}(\,\vv) \prod_{k=1}^K q_{-i}(\w_k)$. The
      parameters of $q_{-i} (h_{i,k})$ resulting from the transformation
      $h_{i,k} = \x_i^\Tr \w_k$ can be computed using only scalar
      manipulations of the mean and covariance of $q(h_{i,k}) =
      \N(\x_i^\Tr \mq_{\w_k}, \x_i^\Tr \Sq_{\w_k} \x_i)$. Because of the
      outer-product structure of $\tilde{t}_{\vv,i}(\vv)$ in equation
      \eqref{eq_tz_fact}, also the parameters of $q_{-i}(\vv)$ can be
      computed using rank-one matrix updates.
  \item Appendix \ref{sec_qv_tilted} describes how the marginal mean and
      covariance of $\vv$ with respect to the tilted distribution
      \eqref{eq_EP_tilted} can be approximated efficiently using a
      similar approach as in the UKF filter \citep{Wan:2000}. Because of the
      factorized approximation \eqref{eq_qep_fact} only one-dimensional
      quadratures are required to compute the means and variances of
      $g(h_{i,k})$ with respect to $q_{-i}(h_{i,k})$ and no multivariate
      quadrature or sigma-point approximations are needed.
  \item Appendix \ref{sec_qh_tilted} presents a new way to approximate
      the marginal distribution of $\hat{p}_i (h_{i,k})$ resulting from
      \eqref{eq_EP_tilted}. In preliminary simulations we found that a
      more simple approach based on the unscented transform and the
      approximate linear filtering paradigm did not capture well the
      information from the left-out observation $y_i$. This behavior was
      more problematic when there was a large discrepancy between the
      information provided by the likelihood term through the marginal
      tilted distribution $\hat{p}_i(y_i|h_{i,k}) = \int
      p(y_i|f_i,\theta)^\eta q_{-i}(\vv) q_{-i} (\h_{i,-k}) d\vv
      d\h_{i,-k}$ and the cavity distribution $q_{-i}(h_{i,k})$, where
      $\h_{i,-k}$ includes all components of $\h_i$ except
      $h_{i,k}$.\footnote{The UKF approach was found to perform better
      with smaller values $\eta$ because then a fraction of the site
      approximation from the previous iteration is left in the cavity,
      which can reduce the possible multimodality of the tilted
      distribution.}
      The improved numerical approximation of $\hat{p}_i (h_{i,k})$ is
      obtained by approximating $q_{-i}(f_i|h_{i,k})$, that is, the
      distribution of the latent function value $f_i = \sum_{k=1}^K
      g(h_{i,k}) + v_0$ resulting from $q_{-i}(\h_{i,-k},\vv| h_{i,k}) =
      q_{-i}(\vv) \prod_{k' \neq k} q_{-i}(h_{i,-k})$, with a Gaussian
      distribution and carrying out the integration over $f_i$
      analytically. According to the central limit theorem we expect this
      approximation to get more accurate as $K$ increases.
  \item Appendix \ref{sec_qtheta_tilted} generalizes the tilted moment
      estimations of Appendices \ref{sec_qv_tilted} and
      \ref{sec_qh_tilted} for approximate integration over the posterior
      uncertainty of $\theta=\log \sigma^2$. Computationally convenient
      marginal mean and covariance estimates for $\vv$,
      $\{h_{i,k}\}_{k=1}^K$, and $\theta$ can be obtained by assuming an
      independent posterior approximation for $\theta$ and making a
      similar Gaussian approximation for $q_{-i}(f_i)$ as in Appendix
      \ref{sec_qh_tilted}.
      Compared to the tilted moments approximations of $\vv$ and $\h_i$
      with fixed $\theta$, the approach requires five additional
      univariate quadratures for each likelihood term, which can be
      facilitated by utilizing the same function evaluations.
  \item Appendix \ref{sec_site_updates} presents expressions for the new
      site parameters obtained by applying the results of Appendices
      \ref{sec_cavity}--\ref{sec_qtheta_tilted} in the moment matching
      condition \eqref{eq_EP_moment_matching}.
      Because of the factorization assumption in \eqref{eq_qep_fact} and
      the UKF-style approximation in the tilted moment estimations
      (Appendix \ref{sec_qv_tilted}), the parameters of the likelihood
      term approximations related to $\vv$ can be written as $\Tauvl_i
      =\m_{\g_i} \sigmat_{\vv,i}^{-1}$ and $\Nuvl_i =\m_{\g_i}
      \sigmat_{\vv,i}^{-2} \mut_{\vv,i}$, where $[\m_{\g_i}]_k = \int
      g(h_{i,k}) q_{-i}(h_{i,k}) d h_{i,k}$ and $\mut_{\vv,i}$ can be
      interpreted as Gaussian pseudo-observations with noise variances
      $\sigmat_{\vv,i}^{2}$ (compare with equation \eqref{eq_sites_tauv}
      and \eqref{eq_sites_nuv}). By comparing the parameter expressions
      with \eqref{eq_qep_v}, the output-layer approximation $q(\vv)$ can
      be interpreted as a linear model where the cavity expectations of
      the hidden unit outputs $g(h_{i,k})= g(\w_k^\Tr \x_i)$ are used as
      input features.
      %
%      Damping the site updates of $\tauvl_i$ and $\Nuvl_i$ requires an
%      approximate approach, because straightforward damping of the
%      canonical parameters of $\tilde{t}_{\vv,i}(\vv)$ in
%      \eqref{eq_tz_fact} would break the outer product form of the site
%      precision matrix $\Tt_{i,\vv\vv} = \Tauvl_i \Tauvl_i^\Tr$.
      %
      The EP updates for the site parameters $\tauwl_{i,k}$ and
      $\nuwl_{i,k}$ related to the input weight approximations $q(\w_k)$
      require only scalar operations similarly to other standard EP
      implementations \citep{Minka:2001a,Rasmussen+Williams:2006}.
\end{itemize}

Appendix \ref{sec_predictions} describes how the predictive distribution
$p(y_*|\x_*)$ related to a new input vector $\x_*$ can be approximated
efficiently using $q(\vv)$, $\{ q(\w_k) \}_{k=1}^K$, and $q(\theta)$. Note
that the prior scale approximations $\{ q(\phi_l) \}_{l=1}^L$ are not needed
in the predictions because information from the hierarchical input weight
priors is approximately absorbed in $\{ q(\w_k) \}_{k=1}^K$ during the EP
iterations.
Appendix \ref{sec_marg_likelih} shows how the EP marginal likelihood
approximation, $\log Z_{\text{EP}} \approx \log p(\y|\X)$, conditioned on
fixed hyperparameters (in this case $\sigma_{\vv,0}^2$), can be computed in a
numerically efficient and stable manner. The marginal likelihood estimate can
be used to monitor convergence of the EP iterations, to determine marginal
MAP estimates of the fixed hyperparameters, and to compare different model
structures.
%
%In the following section we give a general algorithm description which puts
%together all the previously described computational techniques.

\begin{algorithm}[t]
  \label{alg_mlp_ep}
  \DontPrintSemicolon
  %\linesnumbered
  %\KwIn{
  %\textbf{Init:}
  Initialize approximations
    $\mq_\vv$, $\Sq_\vv$, $\{ \mq_{\w_k}, \Sq_{\w_k} \}_{k=1}^K$,
    $ \mu_{\theta}$, $\sigma^2_{\theta}$,
    and $\{ \mu_{\phi_l}, \sigma^2_{\phi_l} \}_{l=1}^L$ using $\X$,
    $\{ \Tauwl_i, \Nuwl_i, \Tauvl_i, \Nuvl_i,
    \tilde{\mu}_{\theta,i}, \tilde{\sigma}^2_{\theta,i} \}_{i=l}^n$,
    $\{ \tilde{\mu}_{w,j}, \tilde{\sigma}^2_{w,j},
     \tilde{\mu}_{\phi,j}, \tilde{\sigma}^2_{\phi,j} \}_{j=1}^{Kd}$,
    and $\{ \tilde{\mu}_{v,k}, \tilde{\sigma}^2_{v,k} \}_{k=1}^{K}$.\;
    % (Section \ref{sec_algorithm})
  %}
  %\KwOut{$y$, the net activation}
  %\Begin{
  %The main iterations:\;
  %\For{iter $\leftarrow 1$ \KwTo maxiter}{
  \Repeat{convergence or maximum number of iterations exceeded}{
    \If{sufficient convergence in $\{ \Tauwl_i, \Nuwl_i, \Tauvl_i, \Nuvl_i,
    \tilde{\mu}_{\theta,i}, \tilde{\sigma}^2_{\theta,i} \}_{i=l}^n$ and
    $\{ \tilde{\mu}_{v,k}, \tilde{\sigma}^2_{v,k} \}_{k=1}^{K}$}
      {
      \nl Run the EP algorithm to update the parameters $\{ \tilde{\mu}_{w,j},
      \tilde{\sigma}^2_{w,j}, \tilde{\mu}_{\phi,j}, \tilde{\sigma}^2_{\phi,j}
      \}_{j=1}^{Kd}$ of the prior site approximations \eqref{eq_twk}
      associated with the input weights $\w$ and the scale parameters $\bm{\phi}$
      (Section \ref{sec_ep_weight_priors}). \;
    }
    Loop over the non-Gaussian likelihood terms:\;
    \For{$i\leftarrow 1$ \KwTo $n$}{
      %{\bf 1.}
      \nl Compute the means and covariances of the cavity distributions:
      $\{ q_{-i}(h_{i,k}) \}_{k=1}^K$ and $q_{-i}(\vv)$
      using equations \eqref{eq_cavity_w} and \eqref{eq_cavity_v}.\;

      %\If{unknown $\theta$}{
      If $\theta$ unknown, compute the cavity distribution $q_{-i}(\theta)$
      (Appendix \ref{sec_cavity}).\;
      %}

      %{\bf 2.}
      \nl Compute the means and covariances of the tilted
      distributions $\hat{q}_i(\vv) = \N(\hat{\mq}_i,\hat{\Sq}_i)$ and
      $\hat{q}_i(h_{i,j}) = \N(\hat{m}_i,\hat{V}_i)$ for $k=1,...,K$: \;
      %
      %\textbf{If} $\theta$ \emph{known} \textbf{then} use \eqref{eq_tilted_v} and
      %\eqref{eq_qh_tilted} \;
      %\textbf{else} use \eqref{eq_mq_tilted_v2} and \eqref{eq_Sq_tilted_v2}, and
      %compute $\hat{q}(\theta) = \N(\hat{\mu}_i,\hat{\sigma}_i^2)$ from
      %\eqref{eq_qtheta_tilted}. \;

      If $\theta$ known, use \eqref{eq_tilted_v} and \eqref{eq_qh_tilted}. \;
      Otherwise, use \eqref{eq_mq_tilted_v2}, \eqref{eq_Sq_tilted_v2}, and
      \eqref{eq_qh_tilted2}, and compute $\hat{q}_i(\theta) =
      \N(\hat{\mu}_i,\hat{\sigma}_i^2)$ from \eqref{eq_qtheta_tilted}. \;

%      \eIf{fixed $\theta$}{
%        $\hat{q}(\vv) = \N(\hat{\mq}_i (\theta),\hat{\Sq}_i (\theta))$
%        (Section \ref{sec_qv_tilted})\;
%
%        $\hat{q}(h_{i,j}) = \N(\hat{m}_i (\theta), \hat{V}_i(\theta) )$ for $k=1,...,K$
%        (Section \ref{sec_qh_tilted})\;
%      }{
%        $\hat{q}(\vv) = \N(\hat{\mq}_i,\hat{\Sq}_i)$ (Section \ref{sec_qtheta_tilted})\;
%
%        $\hat{q}(h_{i,j}) = \N(\hat{m}_i,\hat{V}_i)$ for $k=1,...,K$
%        (Section \ref{sec_qtheta_tilted})\;
%
%        $\hat{q}(\theta) = \N(\hat{\mu}_i,\hat{\sigma}_i^2)$
%        (Section \ref{sec_qtheta_tilted})\;
%      }

      %{\bf 3.}
      \nl Update the site parameters $\Tauwl_i$, $\Nuwl_i$, $\Tauvl_i$,
      $\Nuvl_i$  using \eqref{eq_siteupdate_tauv}, \eqref{eq_siteupdate_nuv},
      and \eqref{eq_sites_w}. \;
      %\ref{sec_site_updates})\;
      If $\theta$ unknown, update $\tilde{\mu}_{\theta,i}$,
      $\tilde{\sigma}^2_{\theta,i}$.

      \If{sequential updates}{
        \nl Rank-1 updates for $\{ q(\w_k) \}_{k=1}^K$ according to
        the changes in $\{ \tauwl_{i,k}, \nuwl_{i,k} \}_{k=1}^K$. \;
        %and rank-2 update for $q(\vv)$.
        If $\theta$ unknown, update the mean and covariance of $q(\theta)$.
        %
%      \If{unknown $\theta$}{
%        Update the mean and covariance of the posterior approximation $q(\theta)$.
%      }
    }
    }
    \If{parallel updates}{
        \nl Recompute the posterior approximations $\{ q(\w_k) \}_{k=1}^K$
        using $\{\Tauwl_i,\Nuwl_i\}_{i=1}^n$.

      %\If{unknown $\theta$}{
        If $\theta$ unknown, recompute the mean and covariance of $q(\theta)$.
      %}
    }
    \nl Update $q(\vv)$  using $\{\Tauvl_i,\Nuvl_i\}_{i=1}^n$. \;
    \If{sufficient data fit}{
      \nl Run the EP algorithm to update the parameters $\{ \tilde{\mu}_{v,k},
      \tilde{\sigma}^2_{v,k} \}_{k=1}^{K}$ of the prior site
      approximations \eqref{eq_tvk} related to the output weights $\vv$
      (Section \ref{sec_ep_weight_priors}).}
  }
  %Compute predictions using \eqref{eq_pred}.
  \caption{An EP algorithm for a two-layer MLP-network with non-Gaussian
  hierarchical priors on the weights.}
\end{algorithm}

\subsubsection{General algorithm description and other practical considerations}
\label{sec_algorithm}

Algorithm \ref{alg_mlp_ep} collects together all the computational components
described in Section \ref{sec_ep_weight_priors} and Appendices
\ref{sec_cavity}--\ref{sec_site_updates} into a single EP algorithm. In this
section we will discuss the initialization, the order of updates between the
different term approximations, and the convergence properties of the
algorithm.

In practice, we combined the EP algorithms for the likelihood sites
\eqref{eq_tz_fact} and the weight prior sites of $\vv$ \eqref{eq_tvk} and
$\w$ \eqref{eq_twk} by running them in turn (in respective order, see lines
2-7, 8, and 1 in Algorithm \ref{alg_mlp_ep}). Because all information from
the observations $\y$ is conveyed by the likelihood term approximations, it
is sensible to iterate first the parameters $\Tauwl_i$, $\Nuwl_i$,
$\Tauvl_i$, and $\Nuvl_i$ to obtain a good data fit while keeping the prior
term approximations \eqref{eq_tvk} and \eqref{eq_twk} fixed so that all the
output weights remain effectively positive and all the input weights have
equal prior distributions.
Otherwise, depending on the scales of the priors, many hidden units and input
weights could be effectively pruned out of the model by the prior site
parameters $\{ \tilde{\mu}_{v,k}, \tilde{\sigma}^2_{v,k} \}_{k=1}^{K}$ and
$\{ \tilde{\mu}_{w,j}, \tilde{\sigma}^2_{w,j}, \tilde{\mu}_{\phi,j},
\tilde{\sigma}^2_{\phi,j} \}_{j=1}^{Kd}$:
for example, the prior means $\tilde{\mu}_{w,j}$ would push the approximate
marginal distribution $q(w_j)$ towards zero and the scale parameter
$\tilde{\sigma}^2_{w,j}$ would adjust the variance of $q(w_j)$ to the level
reflecting the fixed scale prior definition $p(\phi_{l_j}) =
\N(\mu_{\phi,0},\sigma^2_{\phi,0})$.
During the iterations, the data fit can be assessed by monitoring the
convergence of the approximate LOO predictive density $\log Z_{\text{LOO}} =
\sum_i \log p(y_i|\y_{-i},\X) \approx \sum_i \log \hat{Z}_i$ that usually
increases steadily in the beginning of the learning process as the model
adapts to the observations $\y$. In contrast, the approximate marginal
likelihood $\log Z_{\text{EP}} \approx \log p(\y|\X)$ depends more on the
model complexity and usually fluctuates more during the learning process
because many different model structures can produce the same predictions.

We initialized the algorithm with 10-20 iterations over the likelihood sites
with $\theta$ fixed to a sufficiently small value, such as
$\sigma^2=\exp(\theta)=0.3^2$, and the input weight priors set to $
\tilde{\mu}_{w,j}=0$ and $\tilde{\sigma}^2_{w,j}=0.5$, where we have assumed
that the target variables $\y$ and the columns of $\X$ containing the input
variables are normalized to zero mean and unit variance. For the input bias
term (the last column of $\X$), a larger variance
$\tilde{\sigma}^2_{w,d}=2^2$ can be used so that the network is able to
produce step functions at different locations of the input space.
The prior means of the output weights $\tilde{\mu}_{v,k}$ were initialized
with linear spacing in some appropriate interval, for example $[1,2]$, and
the prior variances were set to sufficiently small values such as
$\tilde{\sigma}^2_{v,k}=0.2^2$ so that the elements of the approximate mean
vector $\mq_\vv$ remain positive during the initial iterations.

We initialized the parameters $\{ \Tauwl_i, \Nuwl_i, \Tauvl_i, \Nuvl_i
\}_{i=1}^n$ to zero, which means that in the beginning all hidden units
produce zero expected outputs $\m_{\g_i}=\mb{0}$ resulting into zero messages
to the output weights $\vv$ in equations \eqref{eq_sites_tauv} and
\eqref{eq_sites_nuv}. However, because of the initialization of
$\tilde{\mu}_{v,k}$ and $\tilde{\sigma}^2_{v,k}$, the initial approximate
means of the output weights $[\mq_{\vv}]_k = \tilde{\mu}_{v,k}$ will be
positive and nonidentical.
%$[\mq_{\vv}]_k = \tilde{\mu}_{v,k}>0$.
%
It follows that different nonzero messages will be sent to the input weights
according to \eqref{eq_sites_w} because the tilted moments $\hat{m}_{i,k}$
and $\hat{V}_{i,k}$ from \eqref{eq_qh_tilted} will differ from the
corresponding marginal moments $m_{i,k}=\x_i^\Tr \mq_{\w_k}$ and
$V_{i,k}=\x_i^\Tr \Sq_{\w_k} \x_i$.
If in the beginning all the hidden units were updated simultaneously with the
same priors for the output weights, they would get very similar approximate
posteriors. In this case all the computational units do more or less the same
thing but sufficiently many iterations would eventually result in different
values for all the input weight approximations $q(\w_k)$. This learning
process can be accelerated by the previously described linearly spaced prior
means $\tilde{\mu}_{v,j}$ or by updating only one hidden unit in the
beginning and increasing the number of updated units one by one after every
few iterations. The rationale behind the latter incremental scheme is that
the first unit will usually explain the dominant linear relationships between
$\x$ and $y$ and the remaining units will fit to more local nonlinearities.

The positive Gaussian output weight priors defined at the initialization of
$\tilde{\mu}_{v,k}$ and $\tilde{\sigma}_{v,k}^2$ can be relaxed after the
initial iterations by running the EP algorithm on the term approximations
\eqref{eq_tvk} for the truncated prior terms $\eqref{eq_prv}$ (line 8 in
Algorithm \ref{alg_mlp_ep}).
%
%in connection with each parallel update of the output weights
%as described in Section \ref{sec_ep_weight_priors}.
%
The EP updates for the observation noise $\theta$ can be started after the
initial iterations (lines 2-5 in Algorithm \ref{alg_mlp_ep}). We initialized
the site parameters $\{ \tilde{\mu}_{\theta,i}, \tilde{\sigma}^2_{\theta,i}
\}_{i=1}^n$ to zero, and at the first iteration for $\theta$ we also kept
parameters $\Tauwl_i$, $\Nuwl_i$, $\Tauvl_i$, and $\Nuvl_i$ fixed so that the
initial fluctuations of $\tilde{\mu}_{\theta,i}$ and
$\tilde{\sigma}^2_{\theta,i}$ do not affect the approximations $q(\vv)$ and
$q(\w_k)$.
After sufficient convergence is obtained in the EP iterations on the
parameters of the likelihood sites $\{ \Tauwl_i, \Nuwl_i, \Tauvl_i, \Nuvl_i,
\tilde{\mu}_{\theta,i}, \tilde{\sigma}^2_{\theta,i} \}_{i=l}^n$ and the
parameters of the output weight prior sites $\{ \tilde{\mu}_{v,k},
\tilde{\sigma}^2_{v,k} \}_{k=1}^{K}$, the EP updates can be started on the
parameters $\{ \tilde{\mu}_{w,j}, \tilde{\sigma}^2_{w,j},
\tilde{\mu}_{\phi,j}, \tilde{\sigma}^2_{\phi,j} \}_{j=1}^{Kd}$ of the prior
term approximations \eqref{eq_twk} (line 1 in Algorithm \ref{alg_mlp_ep}).

If all the prior term approximations together with $\{ \Tauwl_i, \Nuwl_i
\}_{i=1}^n$ are kept fixed, that is, $q(\w_k)$ are not updated, the EP
algorithm for the parameters $\Tauvl_i$ and $\Nuvl_i$ related to $q(\vv)$
converges typically in 5-10 iterations. In addition, if $\Tauwl_{\w_k}$ and
$\Nuwl_{\w_k}$ related to only one hidden unit $k$ are updated, the algorithm
will typically converge in less than 10 iterations.
The fast convergence is expected because in both cases the iterations can be
interpreted as a standard EP algorithm on a linear model with known input
variables.
However, updating only one hidden unit at a time will induce moment
inconsistencies between the approximations and the corresponding tilted
distributions of the other $K-1$ hidden unit activations $h_{i,k}$ and the
output weights $\vv$.
This means that such update scheme would require many separate EP runs for
each hidden unit and $\vv$ to achieve overall convergence and in practice it
was found more efficient to update all of them together simultaneously with a
sufficient level of damping.
The updates on $\Tauvl_i$ and $\Nuvl_i$ were damped more strongly by $\delta
\in 0.2$ so that subsequent changes in $q(\vv)$ would not inflict unnecessary
fluctuations in the parameters of $q(\w_k)$, which are more difficult to
determine and converge more slowly compared with $q(\vv)$. In other words, we
wanted to change the output weight approximations more slowly so that
messages have enough time to propagate between the hidden units.
For the same reason, on the line 7 of Algorithm \ref{alg_mlp_ep} parallel
updates are done on $q(\vv)$ whereas the user can choose between sequential
and parallel updates for $q(\w_k)$ (lines 5 and 6).
With sequential posterior updates for $q(\w_k)$, damping the updates of
$\Tauwl_i$ and $\Nuwl_i$ with $\delta \in [0.5,0.8]$ was found sufficient
whereas with parallel updates $\delta < 0.5$ was often required. If there are
large number of input features, it may be more efficient to use parallel
updates for $q(\w_k)$ with larger amount of damping in a similar framework as
described by \citet{Gerven:2009}.

The EP updates for the prior terms of $\vv$ and $\w_k$ are computationally
less expensive and converge faster compared with the likelihood term
approximations. With fixed values of $\{ \Tauwl_i, \Nuwl_i, \Tauvl_i, \Nuvl_i
\}_{i=1}^n$ typically 5-10 iterations were required for convergence of the
updates on the prior term approximations related to $\vv$ in line 8 of
Algorithm \ref{alg_mlp_ep}. The relative time required for computations is
negligible compared with lines 2-7 because the output weights are allowed to
change relatively slowly by damping the updates on $\Tauvl_i$ and $\Nuvl_i$
in line 4.
For this reason we ran the EP algorithm for the prior term approximations of
$\vv$ to convergence after each parallel update of $q(\vv)$ on line 7 to make
sure that components of $\vv$ are distributed at positive values at all
times.
%
%Furthermore, with the parallel updates on $q(\vv)$ there is no need to run
%the EP algorithm on the term approximations of $p(\vv_k|\sigma_{v,0}^2)$ at each
%$i$ to ensure that the output weights are distributed at positive values at
%all times.
%
Because of the propagation of information between approximations $q(\w_k)$
via the hierarchial scale parameter approximations $q(\phi_l)$, larger number
of iterations (typically 10-40) were required for convergence of the updates
on the hierarchical prior term approximations related to $\w$ in line 1 of
Algorithm \ref{alg_mlp_ep}.
At least two sensible update schemes can be considered for EP on the input
weight priors after sufficient convergence is achieved with the initial
Gaussian priors defined using $\tilde{\mu}_{w,j}$ and
$\tilde{\sigma}^2_{w,j}$:
1) The EP algorithm in line 1 is run only once until convergence and then the
other parameters $\{ \Tauwl_i, \Nuwl_i, \Tauvl_i, \Nuvl_i,
\tilde{\mu}_{\theta,i}, \tilde{\sigma}^2_{\theta,i} \}_{i=l}^n$ and $\{
\tilde{\mu}_{v,k}, \tilde{\sigma}^2_{v,k} \}_{k=1}^{K}$ are iterated to
convergence with fixed $\{ \tilde{\mu}_{w,j}, \tilde{\sigma}^2_{w,j}
\}_{j=1}^{Kd}$
or
2) the EP algorithm in line 1 is run once until convergence and after that
only one inner iteration is done on $\{ \tilde{\mu}_{w,j},
\tilde{\sigma}^2_{w,j}, \tilde{\mu}_{\phi,j}, \tilde{\sigma}^2_{\phi,j}
\}_{j=1}^{Kd}$ in line 1.
In the first scheme a fixed sparsity-favoring Gaussian prior is constructed
using the current likelihood term approximations and in the latter scheme the
prior is iterated further within the EP algorithm for the likelihood terms.
The latter scheme usually converges more slowly and requires more damping.
Damping the updates by $\delta \in [0.5, 0.7]$ and choosing a fraction
parameter $\eta \in [0.7, 0.9]$ resulted in numerically stable updates and
convergence for the EP algorithms on the prior term approximations.

The fraction parameter $\eta$ used in updating the likelihood term
approximations according to equations
\eqref{eq_EP_cavity}--\eqref{eq_EP_moment_matching} has a significant effect
on the behavior of the algorithm. Because the approximate tilted
distributions \eqref{eq_qh_tilted} and \eqref{eq_qh_tilted2} are often
multimodal when the prediction resulting from the cavity distributions
$q_{-i}(\vv)$ and $q_{-i}(\h_i)$ does not fit well the left out observation
$y_i$, the value of $\eta$ affects significantly the Gaussian approximation
$\hat{q}(h_{i,k}) = \N(h_{i,k})|\hat{m}_{i,k}, \hat{V}_{i,k})$.
When $\eta$ is close to one and the discrepancy between $y_i$ and the cavity
prediction is large, the resulting multimodal tilted distribution is
represented with a very wide Gaussian distribution.
If there are no other data points supporting the deviating information
provided by $y_i$, the model may simply attempt to widen the predictive
distribution at $\x_i$. Consequently, the updates on the sites with large
discrepancies are often more difficult because of large changes to $\Tauwl_i$
and $\Nuwl_i$. Furthermore, the approximation may not fit well the training
data if there are isolated data points that cannot be considered as outliers.
If $\eta$ is smaller, for example $\eta \in [0.4, 0.7]$, a fraction $1-\eta$
of the site approximation $\tilde{t}_{\w_k,i} (\w_k|\Tauwl_i,\Nuwl_i)$ from
the previous iteration is left in the cavity distribution and the discrepancy
between the cavity prediction and $y_i$ is usually small. Consequently, the
model fits more accurately to the training data, the EP updates are
numerically more robust, and usually less damping is required. However, in
the experiments we found that with smaller values of $\eta$ the model can
also overfit more easily which is why we set $\eta = 0.95$.

%Approximations $q(\vv)$ are relatively unaffected by changes in $\eta$
%because given current input weight approximations $\{ q(\w_k) \}_{k=1}^K$ the
%output weights identify well.

%sufficient data fit with respect to $\log Z_{LOO} \approx \sum_i \log
%\hat{Z}_i$

%In practical implementations all normalization terms such as $\hat{Z}_i$ in
%equation \eqref{eq_qtheta_tilted} should be computed and stored in their
%logarithmic forms to avoid numerical problems, which can arise, for example,
%from small values of the cavity predictive variance $V_{y_i}=V_{f_i}+
%\eta^{-1} \exp(\theta)$ or large difference between the predictive mean
%$m_{f_i}$ and the observation $y_i$. Because larger values of $V_{y_i}$
%usually alleviate such problems, it is also evident that decreasing $\eta$
%improves the numerical stability of the tilted moment evaluations.

%%
%Independent posterior approximations are assumed for the input weights of
%each hidden unit which is why the output of the model can be interpreted as a
%nonlinear combination of sparse linear models. Note that evaluating the
%tilted moments of $\vv$ is not necessary because the site parameters can be
%solved directly as described in Section \ref{sec_site_updates}.

\section{Experiments}

First, three case studies with simulated data were carried out to illustrate
the properties of the proposed EP-based neural network approach with sparse
priors (NN-EP).
Case 1 compares the effects of integration over the uncertainty resulting
from a sparsity-favoring prior with a point-estimate based ARD solution.
Case 2 illustrates the benefits of sparse ARD priors on regularizing the
proposed NN-EP solution in the presence of irrelevant features and various
input effects with different degree of nonlinearity.
Case 3 compares the parametric NN-EP solution to an infinite Gaussian process
network using observations from a discontinuous latent function.
In cases 1 and 3, comparisons are made with an infinite network (GP-ARD)
implemented using a Gaussian process with a neural network covariance
function and ARD-priors with separate variance parameters for all input
weights \citep{Williams:1998, Rasmussen+Williams:2006}.
The neural-network covariance function for the GP-prior can be derived by
letting the number of hidden units approach infinity in a 2-layer MLP network
that has cumulative Gaussian activation functions and fixed zero-mean
Gaussian priors with separate variance (ARD) parameters on the input-layer
weights related to each input variable \citep{Williams:1998}.
Point estimates for the ARD parameters, the variance parameter of the output
weights, and the noise variance were determined by optimizing the marginal
likelihood with uniform priors on the log-scale.
Finally, the predictive accuracy of NN-EP is assessed with four real-world
data sets and comparisons are made with a neural network GP with a single
variance parameter for all input features (GP), a GP with ARD priors
(GP-ARD), and a neural network with hierarchical ARD priors (NN-MC) inferred
using MCMC as described by \citet{Neal:1996a}.

\subsection{Case 1: Overfitting of the ARD} \label{sec_case1}

%\begin{figure*}[!t]
%  \centering
%  \subfigure[]{\includegraphics[scale=0.32]{case1_gp.eps} \label{fig_case1_a} }
%  \subfigure[]{\includegraphics[scale=0.32]{case1_mlp.eps} \label{fig_case1_b} }
%  \subfigure[]{\includegraphics[scale=0.33]{case1_mlp_weights.eps} \label{fig_case1_c} }
%  \caption{Case 1: An example of the overfitting of the ARD on a simulated
%  data set with two relevant input features.
%  (a) A GP model with a neural network covariance function and point-estimates
%  for the ARD parameters.
%  (b) An EP approximation for a neural network with 10 hidden units and
%  Laplace priors on the weights.
%  (c) The 95 \% approximate marginal posterior probability intervals for all the
%  weights of the EP-based neural network.}
%  \label{fig_case1}
%\end{figure*}

\begin{figure*}[!t]
  \centering \subfigure[]{\includegraphics[scale=0.32]{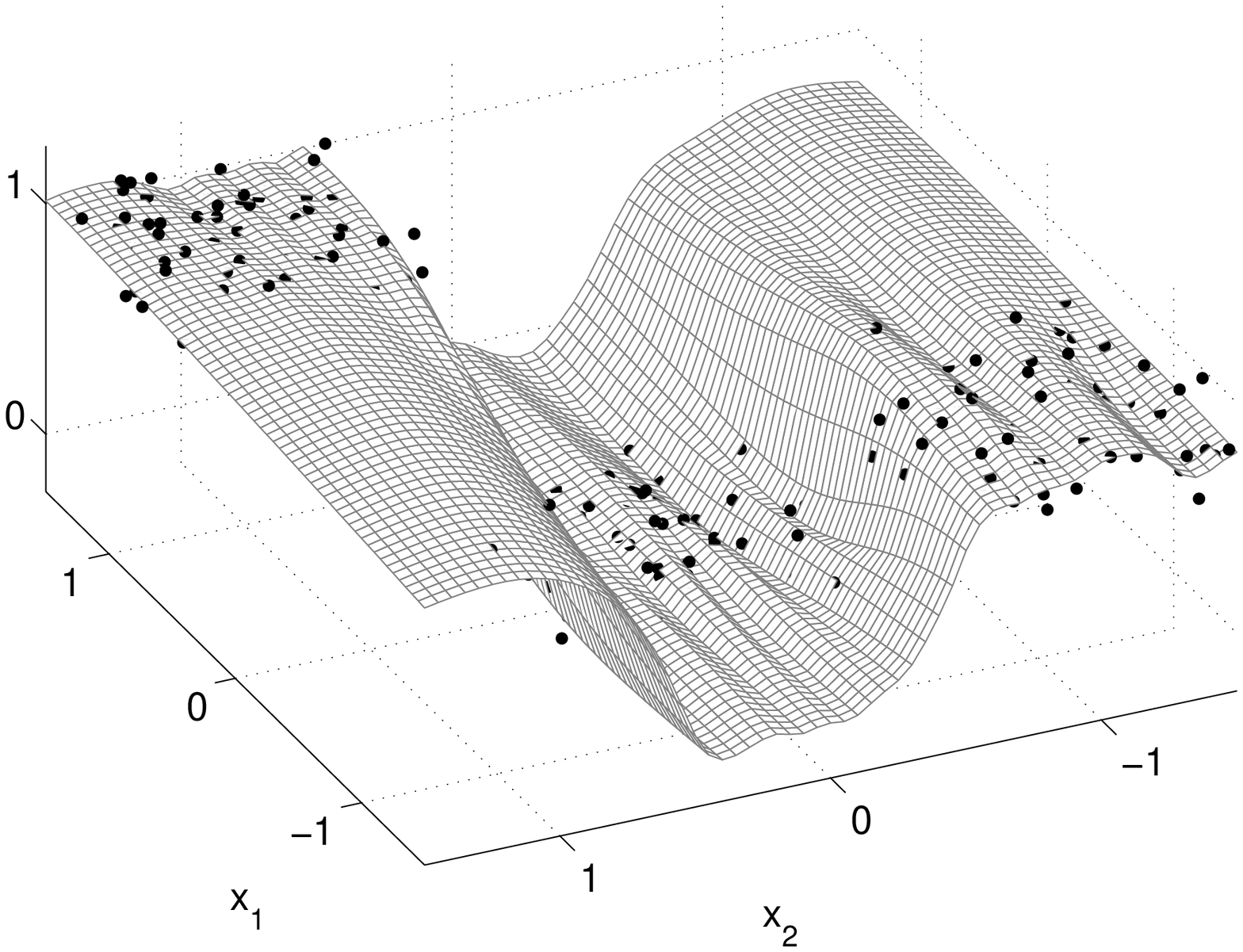}
  \label{fig_case1_a} }
  \subfigure[]{\includegraphics[scale=0.32]{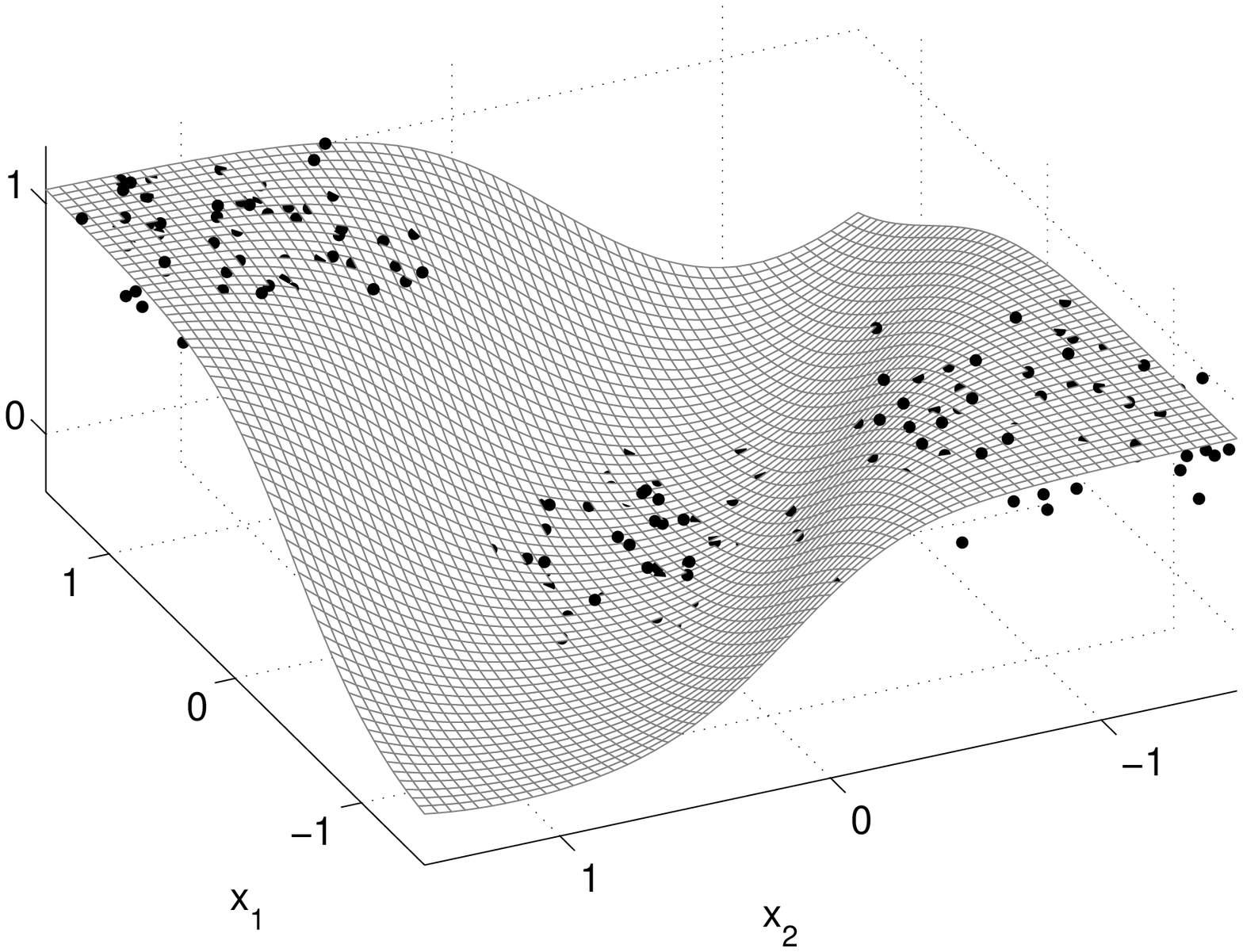}
  \label{fig_case1_b} }
  \subfigure[]{\includegraphics[scale=0.34]{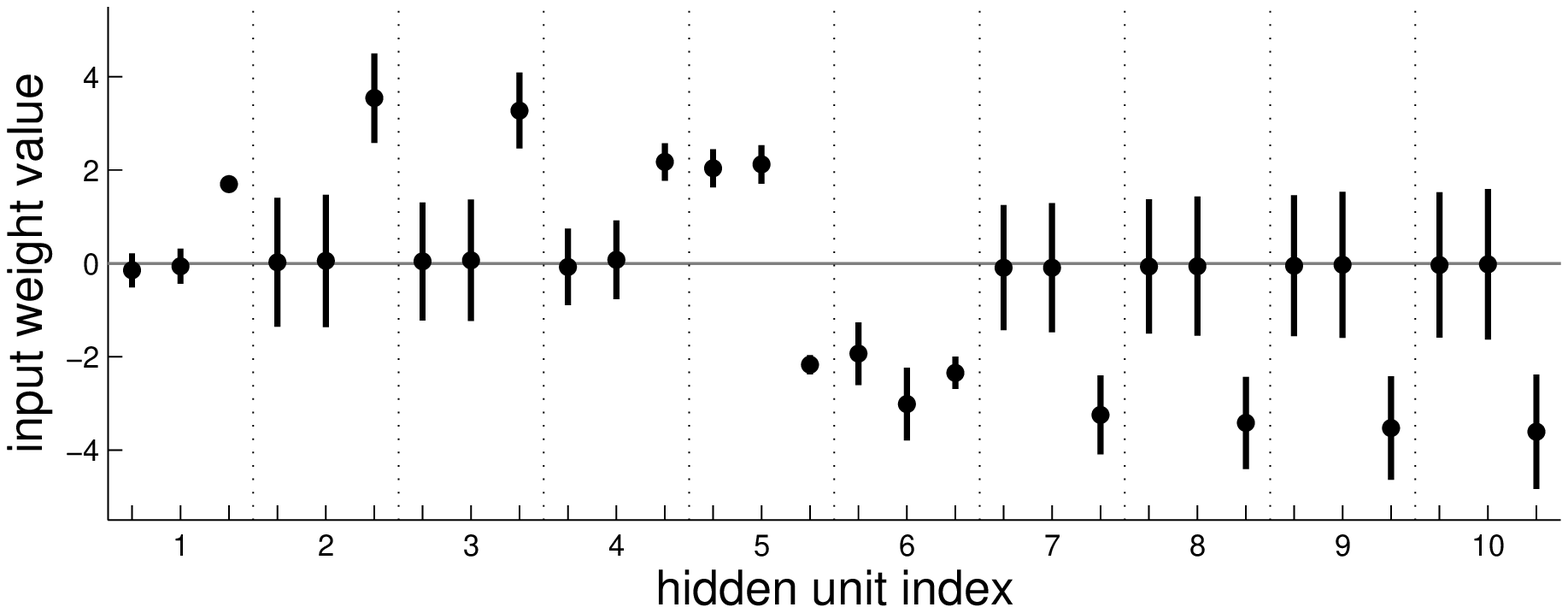}
  \label{fig_case1_c} }
  \subfigure[]{\includegraphics[scale=0.34]{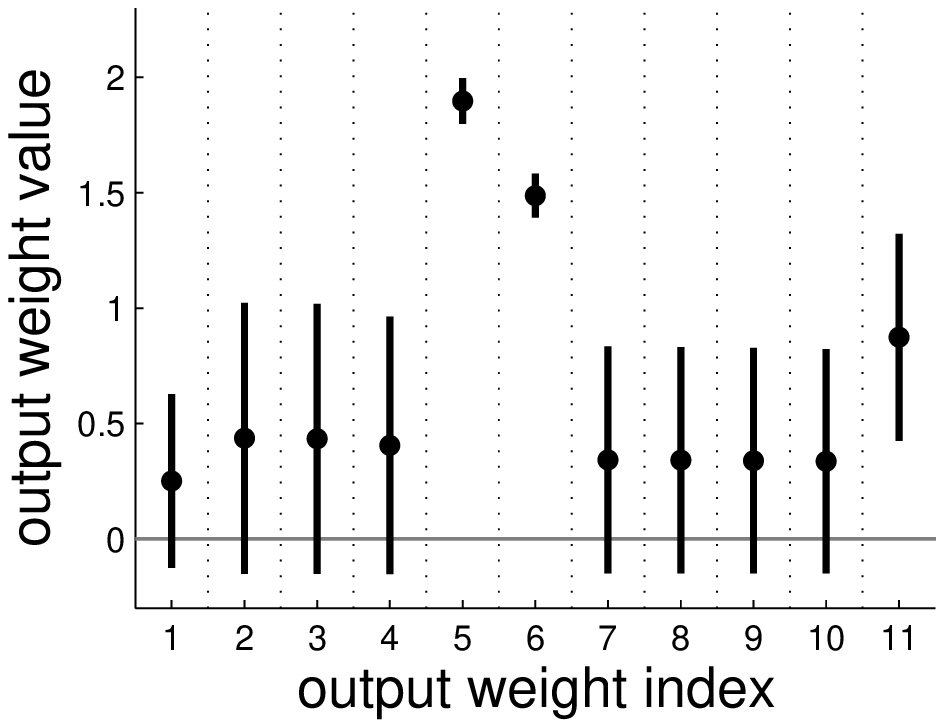}
  \label{fig_case1_d} }
  \caption{Case 1: An example of the overfitting of the
  point-estimate based ARD on a simulated data set with two relevant input
  features. (a) A GP model with a neural network covariance function and
  point-estimates for the ARD parameters. (b) An EP approximation for a
  neural network with 10 hidden units and independent Laplace priors
  with one common unknown scale parameter $\phi$ on the input weights.
  (c) and (d) The 95 \% approximate marginal posterior probability intervals for the
  input weights and the output weights of the EP-based neural network.}
  \label{fig_case1}
\end{figure*}

The first case illustrates the overfitting of ARD with a similar example as
presented by \citet{Qi:2004}. Figure \ref{fig_case1} shows a two-dimensional
regression problem with two relevant inputs $x_1$ and $x_2$.
%
%The data points are obtained only from two of the four quadrants, $\{\x \in
%\mathbb{R}^2 | x_1>0, x_2>0\}$ and $\{\x \in \mathbb{R}^2 | x_1<0, x_2<0\}$,
%so that there is large uncertainty on the relevance of the input variables.
%
The data points are obtained from three clusters, $\{f(\x)=1 | x_1>0.5,
x_2>0.5\}$, $\{f(\x)=0 | 0.5>x_1>-0.5, 0.5>x_2>-0.5\}$, and $\{f(\x)=0.8 |
x_1<-0.5, x_2<-0.5\}$. The noisy observations were generated according to
$y=f(\x) +\epsilon$, where $\epsilon \sim \N(0,0.1^2)$.
%
%so that there is large uncertainty on the relevance of the input variables.
%
The observations can be explained by using a combination of two step
functions with only either one of the input features but a more robust model
can be obtained by using both of them.

Subfigure \subref{fig_case1_a} shows the predictive mean of the latent
function $f(\x)$ obtained with the optimized GP-ARD solution.
%
%Input $x_2$ is effectively pruned out and a very sharp step function is
%obtained with respect to input $x_1$.
%
Input $x_2$ is effectively pruned out and almost a step function is obtained
with respect to input $x_1$.
Subfigure \subref{fig_case1_b} shows the NN-EP solution with $K=10$ hidden
units and Laplace priors with one common unknown scale parameter $\phi_1$ on
the input weights $\w$.
The prior for $\phi_1$ was defined as $ \phi_1 \sim \N(\mu_{\phi,0},
\sigma_{\phi,0}^2)$, where $\mu_{\phi,0}=2 \log (0.1)$ and $\sigma_{\phi,0}^2
= 1.5^2$.
The noise variance $\sigma^2$ was inferred using the same prior definition
for both models: $\theta = \log(\sigma^2) ~ \sim \N(\mu_{\theta,0}
,\sigma_{\theta,0}^2)$, where $\mu_{\theta,0}=2 \log (0.05)$ and
$\sigma_{\theta,0}^2 = 1.5^2$.
NN-EP produces a much smoother step function that uses both of the input
features.
Despite of the sparsity favoring Laplace prior, the NN-EP solution preserves
the uncertainty on the input variable relevances. This shows that the
approximate integration over the weight prior can help to avoid pruning out
potentially relevant inputs.
Subfigure \subref{fig_case1_c} shows the 95\% approximate marginal posterior
probability intervals derived from the Gaussian approximations $q(\w_k)$ with
the same ordering of the weights as in vector $\z^\Tr=[\w_1^\Tr, ...,
\w_K^\Tr]$ (every third weight corresponds to the input bias term). The
vertical dotted lines separate the input weights associated with the different
hidden units.
Subfigure \subref{fig_case1_d} shows the same marginal posterior intervals
for the output weights computed using $q(\vv)$.
%
%Only the first hidden unit has active input weights at indices 1 and 2. The
%input weights related to the other hidden units $2,..., K$ are distributed
%around zero and they have negligible effect on the predictions because also
%the corresponding output weights $v_2,..., v_K$ are distributed close to
%zero.
%
Only hidden units 5 and 6 have clearly nonzero output weights and input
weights corresponding to the input variables $x_1$ and $x_2$ (see the first
two weight distributions in triplets 5 and 6 in panel \subref{fig_case1_c}).
For the other hidden units, the input weights related to $x_1$ and $x_2$ are
distributed around zero and they have negligible effect on the predictions.
In panel \subref{fig_case1_c}, the third input weight distribution
corresponding to the bias term in each triplet are distributed in nonzero
values for many unused hidden units but these bias effects affect only
the mean level of the predictions. These nonzero bias weight values may
be caused by the observations not being normalized to zero mean.
%
%The effects are small on average because the corresponding output
%weights are distributed in small values.
%
The weights corresponding to hidden unit 1 differ from the other unused
units, because a linear action function was assigned to it for illustration
purposes.
If required, a truly sparse model could be obtained by removing the unused
hidden units and running additional EP iterations until convergence.

%The corresponding output weight is active at the index 36 and the output bias
%is active at index 41. No Laplace priors were assigned for the input-bias
%weights and therefore they have larger posterior uncertainty intervals.

%This can be though of as an illustration of the feature selection problem in a very
%high dimensional space where the data points are sparsely distributed

% The predictive relevances of variables 4-10 are similar but
% nonlinearity increases

\subsection{Case 2: The Effect of Sparse Priors in a Regression Problem
Consisting of Additive Input Effects with Different Degree of Nonlinearity}
\label{sec_case2}

The second case study illustrates the effects of sparse priors using a
similar regression example as considered by \citet{Lampinen:2001}.
In our experiments we found two main effects from applying sparsity-promoting
priors with adaptive scale parameters $\bm{\phi}=[\phi_1,...,\phi_L]$ on the
input-layer.
Firstly, the sparse priors can help to suppress the effects of irrelevant
features and protect from overfitting effects in input variable relevance
determination as illustrated in Case 1 (Section~\ref{sec_case1}).
Secondly, sparsity-promoting priors with adaptive prior scale parameters
$\bm{\phi}$ can mitigate the effects of
%
% too aggressive initial iteration steps (too large damping factor $\delta$)
%
unsuitable initial Gaussian prior definitions on the input layer (too large
or too small initial prior variances $\tilde{\sigma}_{w,j}^2$, see Section
\ref{sec_algorithm} for discussion on the initialization).
%
%In the second case, the sparse priors can help to regularize the EP solution
%if the initial learning phase results in unnecessarily strong nonlinear
%effects (the model overfits in the data) because of too loose initial
%Gaussian prior definitions on the input layer (large initial prior variances
%$\tilde{\sigma}_{w,j}^2$, see Section \ref{sec_algorithm} for details). On
%the other hand, too small initial prior variances for the input-layer can
%prevent the model from fitting to the data properly.
%
More precisely, the sparse priors with adaptive scale parameters can help to
obtain better data fit and more accurate predictions by shrinking the
uncertainty on the weights related to irrelevant features towards zero and by
allowing the relevant input weights to gain larger values which are needed in
modeling strongly nonlinear (or step) functions.
Placing very large initial prior variances $\tilde{\sigma}_{w,j}^2$ on all
weights enables the model to fit strong nonlinearities but the initial
learning phase is more challenging and prone to end up in poor local minima.
%
%Furthermore, if the number of unknown model parameters is large compared to
%$n$ the posterior uncertainties are likely to remain large.
%
In this section, we demonstrate that switching to Gaussian ARD priors with
adaptive scale parameter $\phi_1,...,\phi_d$ after a converged EP solution is
obtained with fixed Gaussian priors can reduce the effects of irrelevant
features, decrease the posterior uncertainties on the predictions on $f(\x)$,
and enable the model to fit more accurately latent nonlinear effects.

%if the initial learning phase results in unnecessarily strong nonlinear
%effects (the model overfits in the data) because of too loose initial
%Gaussian prior definitions on the input layer (large initial prior variances
%$\tilde{\sigma}_{w,j}^2$, see Section \ref{sec_algorithm} for details). On
%the other hand, too small initial prior variances for the input-layer can
%prevent the model from fitting to the data properly.

\begin{figure*}[t]
  \centering
  \subfigure[]{\includegraphics[width=0.48\textwidth]{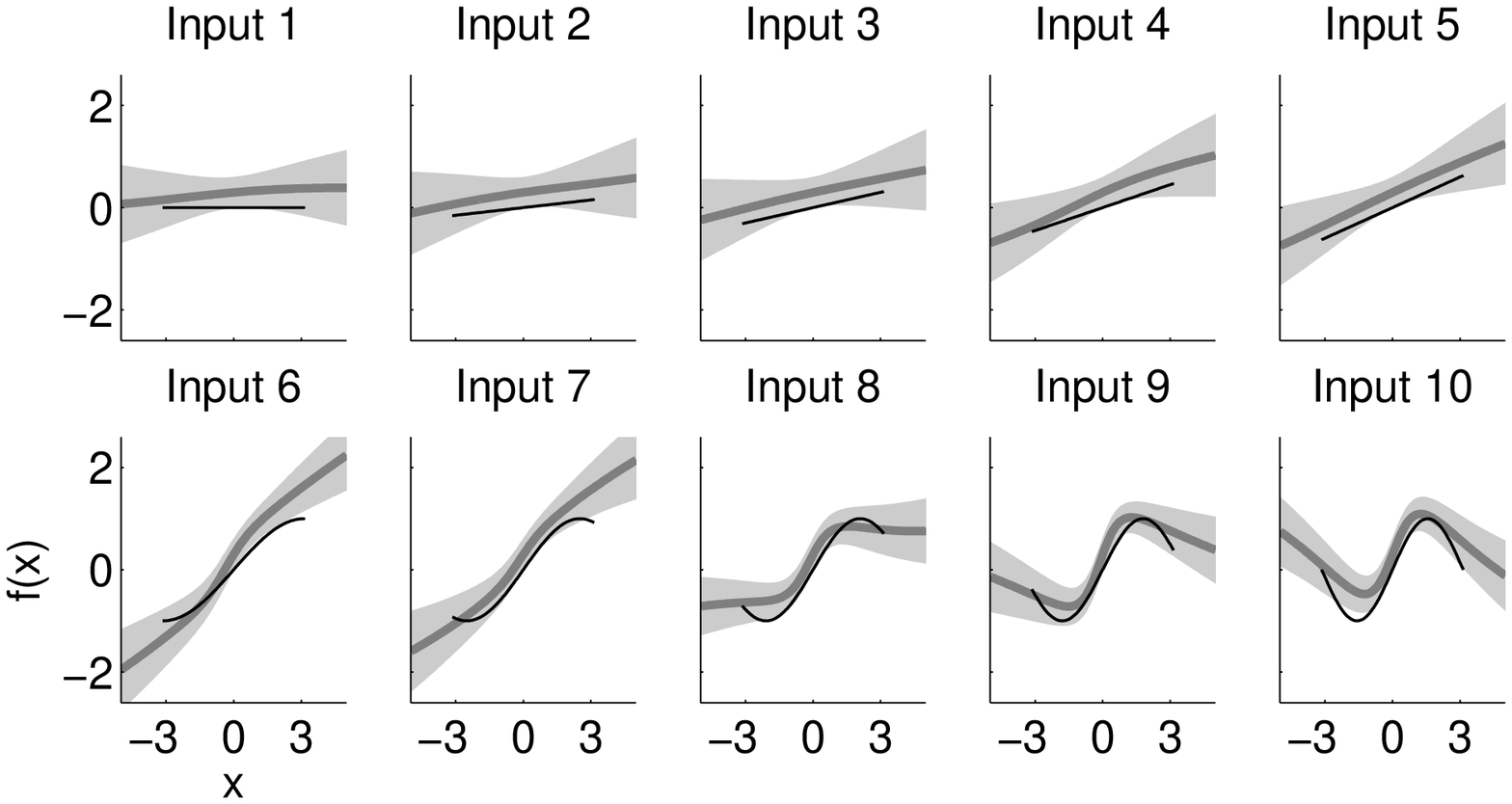}
  \label{fig_case2_a}}
  \subfigure[]{\includegraphics[width=0.48\textwidth]{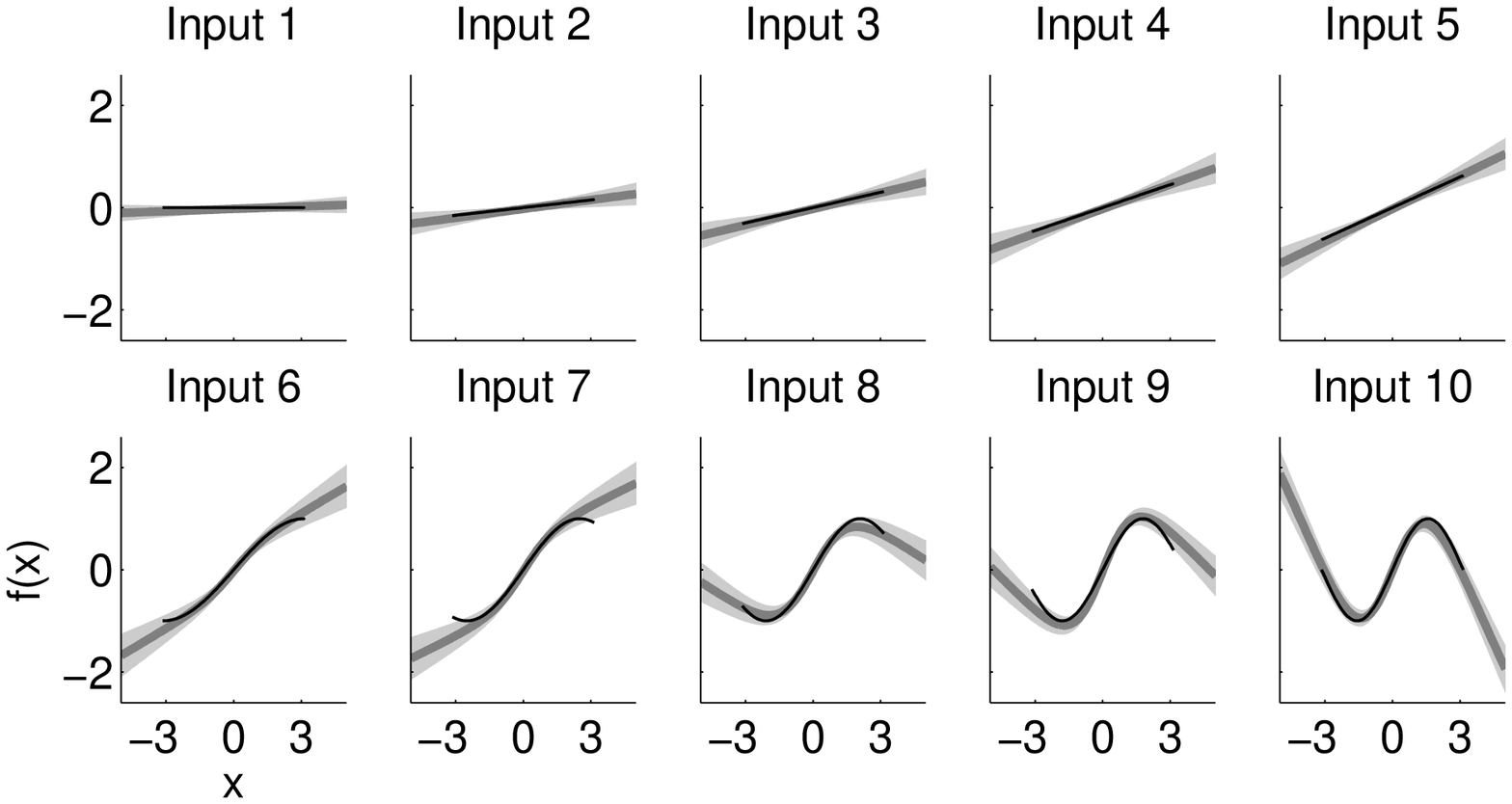}
  \label{fig_case2_b}}
  \caption{Case 2: An artificial regression problem where the observations are formed
  as a sum of additive input effects dependent on ten input features. The true
  additive effects are shown with black lines and the estimated mean predictions
  with dark grey lines. The 95\% posterior predictive intervals are shaded with light grey.
  (a) A converged EP approximation for a neural network with ten hidden units
  and fixed zero-mean Gaussian priors on the input weights. (b) The resulting
  EP approximation when the Gaussian priors of the network in panel (a) are
  replaced with Gaussian ARD priors with separate scale parameters
  $\phi_1,...,\phi_d$ for all input variables, and additional EP iterations
  are done until a new converged solution is obtained. Figure
  \ref{fig_case2_weights} visualizes the approximate posterior distributions
  of the parameters of the ARD network from panel (b).} \label{fig_case2}
\end{figure*}

A data set with 200 observations and ten input variables with different
additive effects on the target variable was simulated. The black lines in
Figure \ref{fig_case2} show the additive effects as a function of each input
variable $x_{i,j}$. The targets $y_i$ were calculated by summing the additive
effects together and adding Gaussian noise with a standard deviation of 0.2.
The first input variable is irrelevant and variables 2-5 have an increasing
linear effect on the target. The effects of input variables 6-10 are
increasingly nonlinear and the last three of them require at least three
hidden units for explaining the observations.

Figure \ref{fig_case2_a} shows the converged NN-EP solution with fixed
zero-mean Gaussian priors on the input weights. The number of hidden units
was set to $K=10$ and the noise variance $\sigma^2$ was inferred using the
prior definition $\mu_{\theta,0}=2 \log (0.05)$ and $\sigma_{\theta,0}^2 =
2^2$. The Gaussian priors were defined by initializing the prior site
parameters of the input weights as $\{\tilde{\mu}_{w,j}=0,
\tilde{\sigma}_{w,j}^2 = 0.4^2 \}_{j=1}^{Kd}$.
The dark grey lines illustrate the posterior mean predictions and the shaded
light gray area the 95\% approximate posterior predictive intervals of the
latent function $f(\x)$. The graphs are obtained by changing the value of
each input in turn from $-5$ to $5$ while keeping the others fixed at zero.
The training observations are obtained by sampling all input variables
linearly from the interval $x_{i,j} \in [-\pi,\pi]$.
Panel \subref{fig_case2_b} shows the resulting NN-EP solution when the
Gaussian priors of the network in panel \subref{fig_case2_a} are replaced
with Gaussian ARD priors with adaptive scale parameter $\phi_1,...,\phi_d$
and additional EP iterations are done until convergence.
Prior distributions for the scale parameters were defined as $ \phi_l \sim
\N(\mu_{\phi,0}, \sigma_{\phi,0}^2)$, where $\mu_{\phi,0}=2 \log (0.01)$ and
$\sigma_{\phi,0}^2 = 2.5^2$. This prior definitions favors small input
variances close to $0.01$ but enables also larger values around one. It
should be noted that the actual variance parameters $\tilde{\sigma}_{w,j}^2$
of the prior site approximations can attain much larger values from the EP
updates.

%
%A one common fixed scale parameter $\lambda_{1} = 2^{-1/2} \exp( \phi_{1}/2)
%$ was selected so that the prior variances of the Laplace priors were equal
%with the variances of the Gaussian priors.
%
With the Gaussian priors (Figure \ref{fig_case2_a}), the predictions do not
capture the nonlinear effects very accurately and the model produces a small
nonzero effect on the irrelevant input 1.
Applying the ARD priors (Figure \ref{fig_case2_b}) with additional iterations
produces clearly more accurate predictions on the latent input effects and
effectively removes the predictive effect of input 1. The overall approximate
posterior uncertainties on the latent function $f(\x)$ are also smaller
compared with the initial Gaussian priors.
We should note that the result of panel (a) depends on the initial Gaussian
prior definitions and choosing a smaller $\tilde{\sigma}_{w,j}^2 = 0.2^2$ or
optimizing it could give more accurate predictions compared with the solution
visualized in panel (a).

\begin{figure*}[t]
  \centering
  \subfigure[]{\includegraphics[scale=0.38]{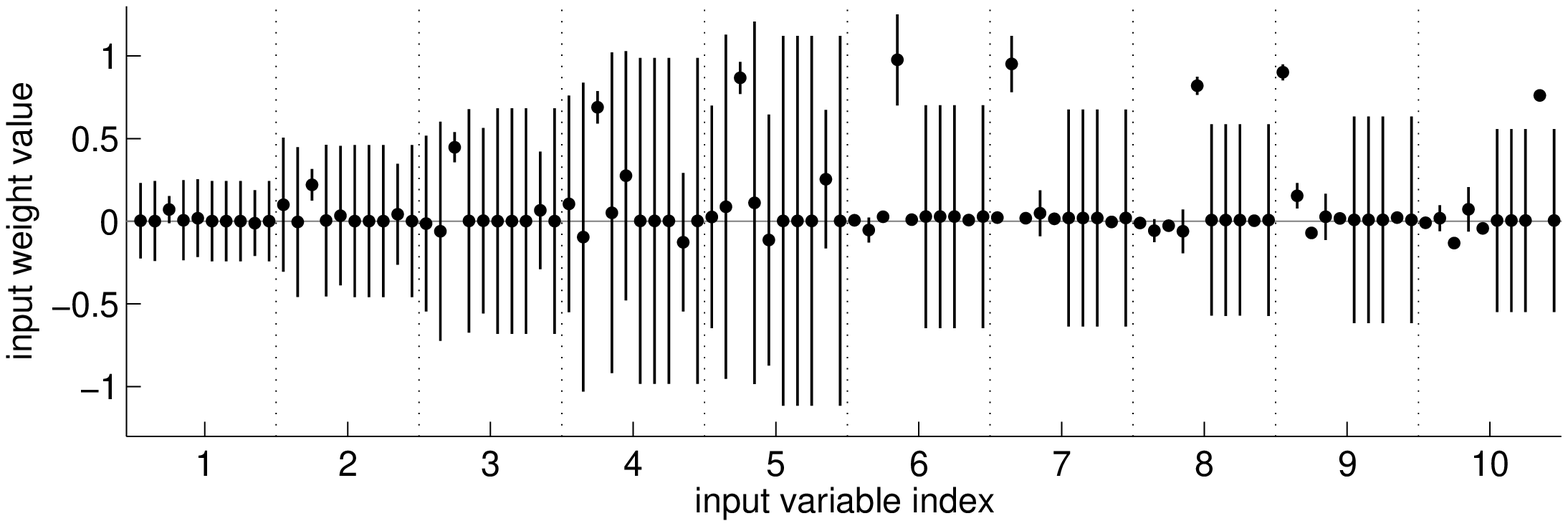}
  \label{fig_case2_w}}
  \subfigure[]{\includegraphics[scale=0.38]{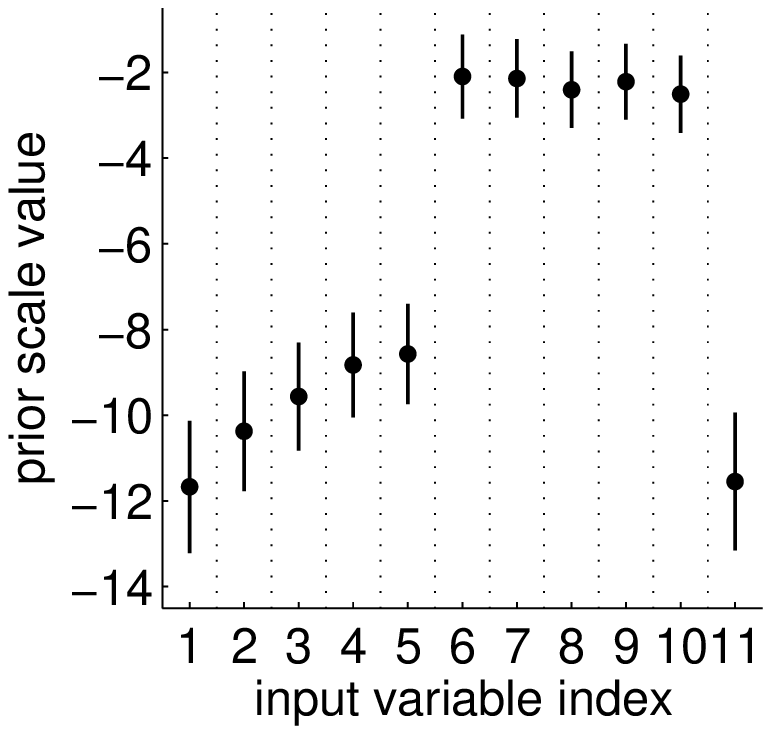}
  \label{fig_case2_phi}}
  \subfigure[]{\includegraphics[scale=0.38]{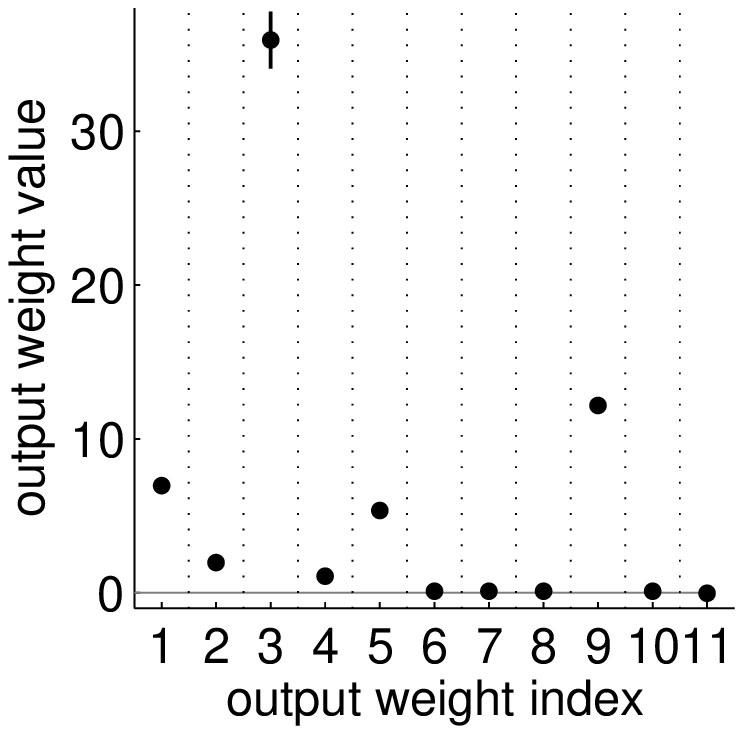}
  \label{fig_case2_v}} \caption{Case 2: Visualization of the model parameters
  related to the artificial regression problem shown in Figure \ref{fig_case2}.
  Panels (a), (b), and (c) show the 95\% marginal posterior credible
  intervals for the input weights $\w$, the scale parameters
  $\phi_1,...,\phi_d$, and the output weights $\vv$ of the neural network
  with Gaussian ARD priors from Figure \ref{fig_case2_b}.
  In panel (a) the input weights associated with each additive input effect
  (1-10) are grouped together (the bias terms are not shown). The weight
  distributions related to the linear input effects 1--5 are much smaller
  compared with the nonlinear effects 6--10, which is why they are scaled by 40
  for better illustration in panel (a).} \label{fig_case2_weights}
\end{figure*}

Figure \ref{fig_case2_weights} shows the 95\% posterior credible intervals
for the input weights $\w$ (a), the prior scale parameters
$\phi_1,...,\phi_d$ (b), and the output weights $\vv$ (c) of the NN-EP
approximation with ARD priors visualized in Figure \ref{fig_case2_b}. In
panel \subref{fig_case2_w} the input weights from the different hidden units
are grouped together according to the different additive input effects 1--10,
and the weights related to the linear effects 1--5 are scaled by 40 for
illustration purposes, because they are much smaller compared with the
weights associated with the nonlinear input effects 6--10.
%
%and panel \ref{fig_case2_v} visualizes the marginal posterior uncertainty on
%the output weights $v_k$ related to each hidden unit.
%
From panels \subref{fig_case2_w} and \subref{fig_case2_v} we see that only
hidden units are 1--5 and 9 have clearly non-zero effect on the predictions.
The linear effects of inputs 1--5 are modeled by unit 3 that has very small
but clearly nonzero input weights in panel \subref{fig_case2_w} and a very
large output weight in panel \subref{fig_case2_w}.
The input weights related to the irrelevant input 1 are all zero in the 95\%
posterior confidence level.
By comparing panels \subref{fig_case2_w} and \subref{fig_case2_v} we can also
see that hidden units 1, 2, 4, 5, and 9 are most probably responsible for
modeling the nonlinear input effects 6-7 because of large input weights
values.
Panel \subref{fig_case2_phi} gives further evidence on this interpretation
because the scale parameters associated with the nonlinear input effects
6--10 are clearly larger compared to effects 1--5.
The scale parameters associated with the linear input effects 1--5 increase
steadily as the magnitudes of the true effects increase.
These results are congruent with the findings of \citet{Lampinen:2001} who
showed by MCMC experiments that with MLP models the magnitudes of the ARD
parameters and the associated input weights also reflect the degree of
nonlinearity associated with the latent input effects, not only the relevance
of the input features.

%From panel \subref{fig_case2_a} we can see that the posterior uncertainty on
%the input weights related to the inactive units 1,6, and 7 is much larger in
%each input group than the uncertainty on the active input weights and this
%uncertainty is controlled by the magnitude of the scale parameter
%$\lambda_l$.
%
%From panel \subref{fig_case2_w} we can see that the input weights of the
%irrelevant input 1 are distributed around zero whereas the almost linear
%predictions for effects 2-5 are done using 1-2 small input weights.
%%
%The more strongly nonlinear predictions for effects 6-10 are made using 2-3
%much larger input weights.

%The nonlinear effects of variables 7,9, and 10, use clearly at least two
%hidden units, and the rest of the nonlinearity is modeled as an average of
%smaller contributions from several hidden units.
%The nonlinearity of the related additive effect increases both the posterior
%weights although the predictive relevances of variables 4-10 are similar.

\subsection{Case 3: Comparison of a Finite vs. Infinite Network with Observations from
  a Latent Function with a Discontinuity} \label{sec_case3}

\begin{figure}[t]
  \centering \subfigure[]{\includegraphics[width=0.48\textwidth]{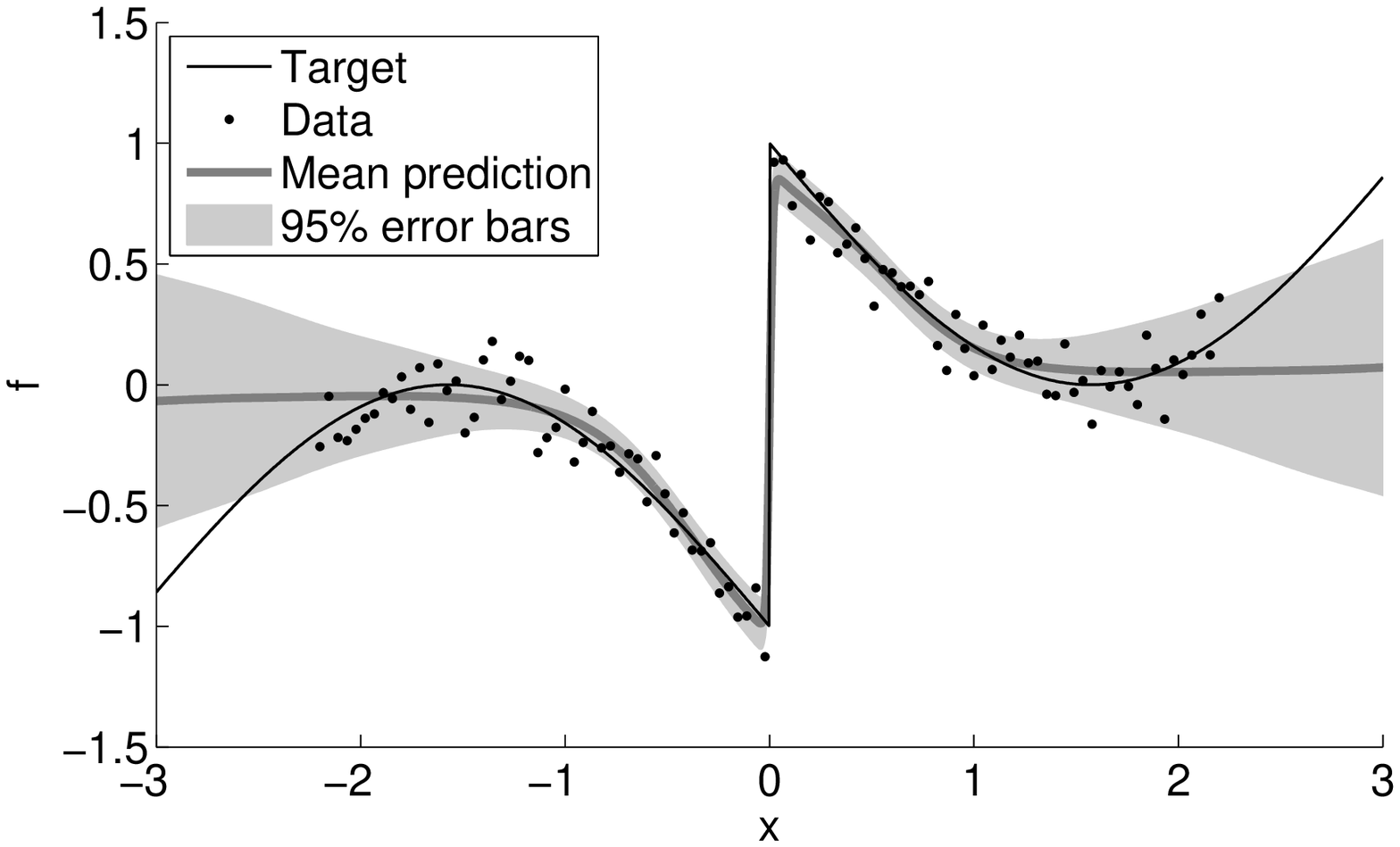}
  \label{fig_case3_a}}
  \subfigure[]{\includegraphics[width=0.48\textwidth]{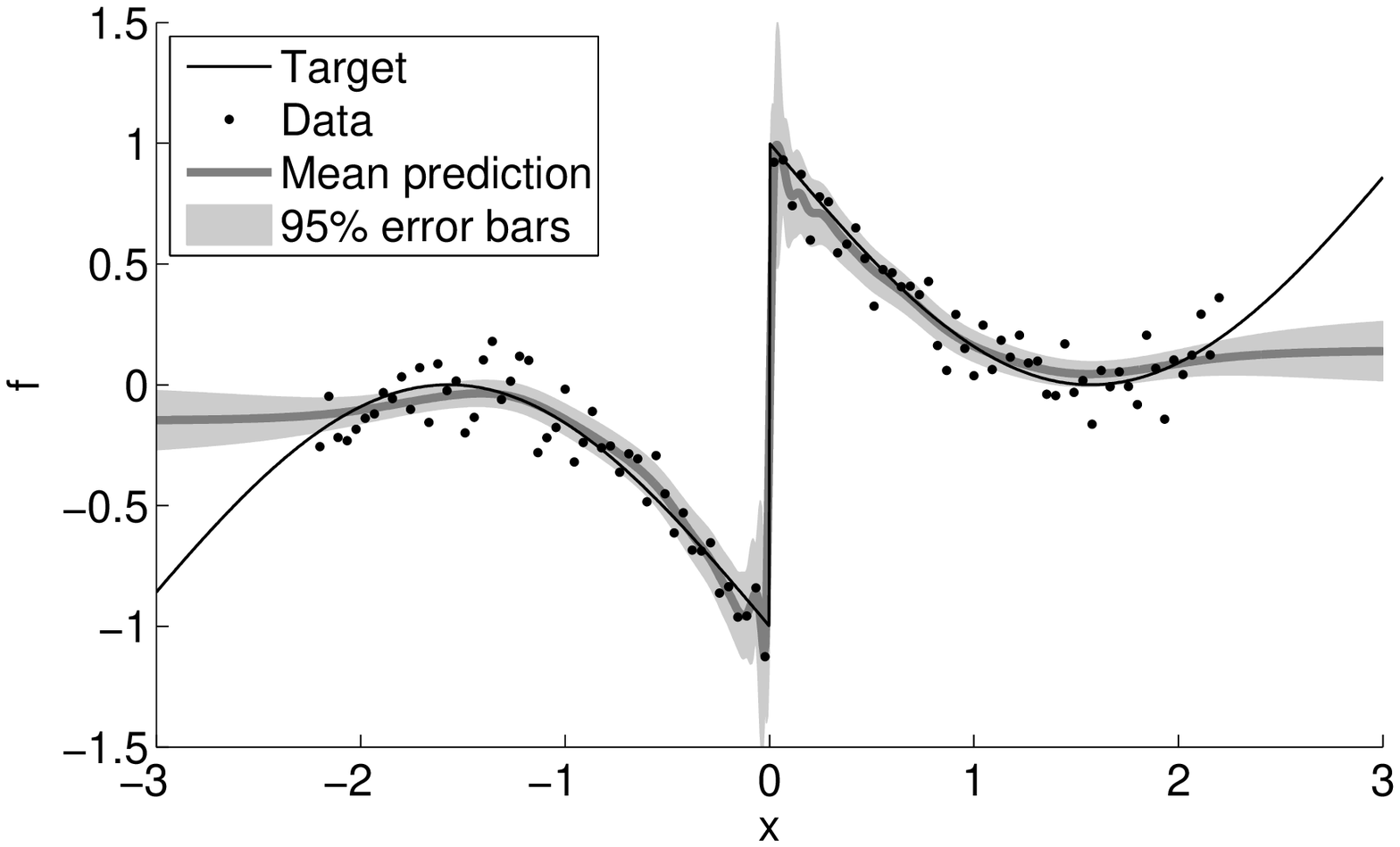}
  \label{fig_case3_b}}
  \caption{Case 3: An artificial regression problem consisting of noisy
  observations (black dots) generated from a latent function (black lines)
  that has a discontinuity at zero.
  Panel (a) shows the mean predictions (dark grey line) and the 95\%
  credible intervals (light gray shaded area) obtained using the proposed EP
  approach for a NN with ten hidden units and Laplace priors with one common scale
  parameter $\phi$ on the input weights.
  Panel (b) visualizes the corresponding predictive distribution obtained using
  a GP with a neural network covariance function.}
  \label{fig_case3}
\end{figure}

The third case study compares the performance of the finite NN-EP network
with an infinite GP network in a one-dimensional regression problem with a
strong discontinuity. Figure \ref{fig_case3} shows the true underlying
function (black lines) that has a discontinuity at zero together with the
noisy observations (black dots). Panel \subref{fig_case3_a} shows the
predictive distributions obtained using NN-EP with ten hidden units ($K=10$)
and Laplace priors with one common scale parameter $\phi$.
The prior distribution for the scale parameter was defined with
$\mu_{\phi,0}=2 \log (0.01)$ and $\sigma_{\phi,0}^2 = 2.5^2$, and the noise
variance $\sigma^2$ was inferred from the data using the prior definition
$\mu_{\theta,0}=2 \log (0.05)$ and $\sigma_{\theta,0}^2 = 2^2$.
Panel \subref{fig_case2_b} shows the corresponding predictions obtained using
a GP with a neural network covariance function. With the GP network the noise
variance was optimized together with the other hyperparameters using the
marginal likelihood.
The finite NN-EP network explains the discontinuity with a slightly smoother
step compared to infinite GP network, but the GP mean estimate shows
fluctuations near the discontinuity. It seems that the infinite GP network
fits more strongly to individual observations near the discontinuity. This
shows that a flexible parametric model with a limited complexity may
generalize better with finite amount of observations even though the GP model
includes the correct solution a priori. This is in accordance with the
results described by \citet{Winther:2001}.

\subsection{Predictive Comparisons with Real World Data}

In this section the predictive performance of NN-EP is compared to three
other nonlinear regression methods using the following real-world data sets:
the concrete quality data (Concrete) analyzed by \citet{Lampinen:2001}, the
Boston housing data (Housing) and the unnormalized Communities and Crime data
(Crime) that can be obtained from the UCI data repository \citep{UCI:2010},
and the robot arm data (Kin40k) utilized by
\citet{Schwaighofer:2003}.\footnote{ Kin40k data is based on the same
simulation of the forward kinematics of an 8 link all-revolute robot arm as
the Kin family of data sets available at
\url{http://www.cs.toronto.edu/~delve/} except for lower noise level and
larger amount of observations.}
The number of observations $n$ and the number of input features $d$ are shown
in Table~\ref{table1} for each data set. The Kin40k includes originally only
8 input features but we added 92 irrelevant uniformly sampled random inputs
to create a challenging feature selection problem.
%
% 506/13
% 215/27
% 1993/102
% 5000/100
%
The columns of the input matrices $\X$ and the output vectors $\y$ were
normalized to zero mean and unit variance for all methods.
The predictive performance of the models was measured using the log
predictive densities and the squared errors evaluated with separate test
data. We used 10-fold cross-validation with the Housing, Concrete, and Crime
data, whereas with Kin40k we chose randomly 5000 data points for training and
used the remaining observations for validation.

%
%The KIN8NM data set 4 represents the forward dynamics of an 8 link
%all-revolute robot arm, based on 8192 examples. The goal is to predict the
%distance of the end-effector from a target, given the twist angles of the 8
%links as features. KIN40K represents the same task, yet has a lower noise
%level than KIN8NM and contains 40000 examples.

The proposed NN-EP solution was computed using two alternative prior
definitions: Laplace priors with one common scale parameter $\phi$
(NN-EP-LA), and Gaussian ARD priors with separate scale parameters
$\phi_1,...,\phi_d$ for all inputs including the input bias terms
(NN-EP-ARD).
With both prior frameworks, the hyperpriors for the scale parameters were
defined as $ \phi_l \sim \N(\mu_{\phi,0}, \sigma_{\phi,0}^2)$, where
$\mu_{\phi,0}=2 \log (0.01)$ and $\sigma_{\phi,0}^2 = 2.5^2$. This definition
encourages small input weight variances (around $0.01^2$) but enables also
large input weight values if required for strong nonlinearities assuming the
input variables are scaled to unit variance.
The noise level $\theta=\log(\sigma^2)$ was inferred from data with a prior
distribution defined by $\mu_{\theta,0}=2 \log (0.01)$ and
$\sigma_{\theta,0}^2 = 2^2$, which is a sufficiently flexible prior when the
output variables $\y$ are scaled to unit variance.
The methods used for comparison include an MCMC-based MLP network with ARD
priors (NN-MC) and two GPs with a neural network covariance function: one
with common variance parameter for all inputs (GP), and another with separate
variance hyperparameters for all inputs (GP-ARD). With both GP models the
hyperparameters were estimated by gradient-based optimization of the
analytically tractable marginal likelihood \citep{Rasmussen+Williams:2006}.
For NN-MC and NN-EP, we set the number of hidden units to $K=10$ with the
Housing, Concrete, and Crime data sets. With the Kin40k data, we set $K=30$
because $n$ is large and fewer units were found to produce clearly worse data
fits.

Table \ref{table1} summarizes the means (mean) and standard deviations (std)
of the log predictive densities (LPDs) and the squared errors (SEs). Because
the distributions of the LPD values are heavily skewed towards negative
values, we summarize also the lower 1\% percentiles (prct 1\%). Similarly,
because the SE values are skewed towards positive values we summarize also
the 99\% percentiles (prct 99\%). These additional measures describe the
quality of the worst case predictions of the methods.
Table \ref{table1} summarizes also the average relative CPU times (cputime)
required for parameter estimation and predictions using MATLAB
implementations. The GP models were implemented using the
GPstuff\footnote{\url{http://becs.aalto.fi/en/research/bayes/gpstuff/}}
toolbox and NN-MC was implemented using the
MCMCstuff\footnote{\url{http://becs.aalto.fi/en/research/bayes/mcmcstuff/}}
toolbox.
The CPU times were averaged over the CV-folds and scaled so that the relative
cost for NN-EP is one. These running time measures are highly dependent on
the implementation, the tolerance levels in optimization and iterative
algorithms, and the number of posterior draws, and therefore they are
reported only to summarize the main properties regarding the scalability of
the different methods. When assessing the results with respect to the Housing
and Concrete data sets, it is worth noting that there is evidence that an
outlier-robust observation model is beneficial over the Gaussian model used
in this comparison with both data sets \citep{Jylanki:2011}.

\begin{table}[!ht]
\caption{A predictive assesment of the proposed EP approach for neural
networks with two different prior definitions: Laplace priors with one common
scale parameter $\phi$ (NN-EP-LA) and Gaussian ARD priors with separate scale
parameters $\phi_1,...,\phi_d$ for all inputs (NN-EP-ARD). The methods used
for comparison include a neural network with ARD priors inferred using MCMC
(NN-MC), and two GPs with a neural network covariance: one with a common
variance hyperparameter for all inputs (GP), and another with separate
variance hyperparameters for all inputs (GP-ARD).
The log predictive densities are summarized with their means, standard
deviations (std), and lower 1\% percentiles (prct 1\%). The squared errors
are summarized with their means, standard deviations (std), and upper 99\%
percentiles (prct 99\%).} \label{table1}
\vspace{-0.3cm}
\begin{center}
\begin{tabular}{l|ccc|ccc|r}
Housing & \multicolumn{3}{|c|}{log predictive density (LPD)} &
\multicolumn{3}{|c|}{squared error (SE)} & \\
($n$=506, $d$=13) & mean&std&prct 1\% & mean&std&prct 99\% & cputime\\
\hline
NN-EP-LA  & -0.44 & 1.64 & ~-7.55 & 0.15 & 0.45 & 2.42 & ~~1.0 \\
NN-EP-ARD & -0.50 & 1.66 & ~-6.31 & 0.17 & 0.49 & 1.60 & ~~1.0 \\
NN-MC 	  & -0.08 & 1.17 & ~-4.54 & 0.11 & 0.50 & 1.18 & 110.5 \\
GP 	      & -0.29 & 2.35 & ~-7.57 & 0.13 & 0.53 & 1.98 & ~~0.3 \\
GP-ARD    & -0.20 & 2.00 & -10.71 & 0.10 & 0.37 & 1.53 & ~~1.0 \\
\hline
%\multicolumn{8}{c}{} \\
\multicolumn{2}{l}{Concrete ($n$=215, $d$=27)} & \multicolumn{6}{c}{} \rule{0pt}{3ex} \\
\hline
NN-EP-LA  & ~0.18 & 0.85 &-3.05 & 0.05 & 0.08 & 0.30 & ~~1.0 \\
NN-EP-ARD & ~0.05 & 1.03 &-4.61 & 0.05 & 0.11 & 0.57 & ~~0.8 \\
NN-MC 	  & ~0.22 & 1.52 &-3.62 & 0.04 & 0.08 & 0.28 & 103.0 \\
GP 	      & -0.07 & 1.70 &-5.12 & 0.06 & 0.11 & 0.66 & ~0.03 \\
GP-ARD    & ~0.15 & 1.98 &-4.23 & 0.04 & 0.08 & 0.28 & ~~0.6 \\
\hline
%\multicolumn{8}{c}{} \\
\multicolumn{2}{l}{Crime ($n$=1993, $d$=102)} & \multicolumn{6}{c}{} \rule{0pt}{3ex} \\
\hline
NN-EP-LA  & -0.83 &0.89 &-4.64 & 0.31 &0.55 &2.60 & ~1.0 \\
NN-EP-ARD & -0.84 &0.89 &-4.81 & 0.31 &0.55 &2.75 & ~0.2 \\
NN-MC 	  & -0.80 &0.93 &-4.81 & 0.29 &0.53 &2.60 & 19.8 \\
GP 	      & -0.81 &0.91 &-4.80 & 0.30 &0.54 &2.69 & ~0.2 \\
GP-ARD 	  & -0.81 &1.01 &-5.49 & 0.30 &0.55 &2.75 & ~4.4 \\
\hline
%\multicolumn{8}{c}{} \\
\multicolumn{2}{l}{Kin40k ($n$=5000, $d$=100)} & \multicolumn{6}{c}{} \rule{0pt}{3ex} \\
\hline
NN-EP-LA  & -0.59 &0.89 &-4.27 & 0.19 &0.29 &1.38 & ~1.0 \\
NN-EP-ARD & ~0.27 &1.19 &-4.63 & 0.03 &0.08 &0.37 & ~0.9 \\
NN-MC 	  & ~0.49 &1.51 &-5.37 & 0.02 &0.07 &0.26 & 48.7 \\
GP 	      & -1.15 &0.72 &-4.18 & 0.58 &0.83 &4.06 & ~0.5 \\
GP-ARD 	  & ~0.64 &1.11 &-3.90 & 0.02 &0.05 &0.24 & 32.3 \\
\end{tabular}
\end{center}
\end{table}

Table\ref{table1} shows that NN-EP-LA performs slightly better compared to
NN-EP-ARD in all data sets except in Kin40k, where NN-EP-ARD gives clearly
better results. The main reason for this is probably the stronger sparsity of
the NN-EP-ARD solutions: In Kin40k data there are a large number truly
irrelevant features that should be completely pruned out of the model,
whereas with the other data sets most features have probably some relevance
for predictions or at least they are not generated in a completely random
manner. Further evidence for this is given by the clearly better performance
of GP-ARD over GP with the Kin40k data.

If the mean log predictive densities (MLPDs) are considered, the NN-MC
approach based on a finite network performs best in all data sets except with
Kin40k, where the infinite GP-ARD network is slightly better. The main reason
for this is probably the strong nonlinearity of the true latent mapping,
which requires a large number of hidden units, and consequently the infinite
GP network with ARD priors gives very accurate predictions. In pair-wise
comparisons the differences in MLPDs are significant in 95\% posterior
confidence level only with Housing and Kin40k data sets. In terms of mean
squared errors (MSEs), GP-ARD is best in all data sets except Crime, but with
95\% confidence level the pair-wise differences are significant only with the
Kin40k data.
With the Kin40k data, the performance of NN-MC could probably be improved by
increasing $K$ or drawing more posterior samples, because learning the
nonlinear mapping with a large number of unknown parameters and potentially
multimodal posterior distribution may require a very large number of
posterior draws.

When compared with NN-MC and GP-ARD, NN-EP gives slightly worse MLPD scores
with all data sets except with Concrete. The pair-wise differences in MLPDs
are significant with 95\% confidence level in all cases except with the
Concrete data.
In terms of MSE scores, NN-EP is also slightly but significantly worse with
95\% confidence level in all data sets.
By inspecting the std:s and 1\% percentiles of the LPDs, it can be seen that
NN-EP achieves better or comparable worst case performance when compared to
GP-ARD. In other words, NN-EP seems to make more cautions predictions by
producing less very high or very low LPD values.
One possible explanation for this behavior is that it might be an inherent
property of the chosen approximation. Approximating the possibly multimodal
tilted distribution $\hat{p}(h_{i,k})$, where one mode is near the cavity
distribution $q_{-i}(h_{i,k})$ and another at the values of $h_{i,k}$ giving
the best fit for $y_i$, with an unimodal Gaussian approximation as described
in Appendix \ref{sec_qh_tilted}, may lead to reduced fit to individual
observations.
Another possibility is that the EP-iterations have converged into a
suboptimal stationary solution or the maximum number of iterations has been
exceeded. Doing more iterations or using an alternative non-zero
initialization for the input-layer weights might result in better data fit.
The second possibility is supported by the generally acknowledged benefits
from different initializations, for example, the unsupervised schemes
discussed by \citet{Erhan:2010}, and our experiments using the Kin40k data
without the extra random inputs. We found that initializing the location
parameters $\tilde{\mu}_{v,k}$ and $\tilde{\mu}_{w,j}$ of the prior site
approximations \eqref{eq_tvk} and \eqref{eq_twk} using a gradient-based MAP
estimate of the weights $\w$ and $\vv$, and relaxing the prior site
approximations after initial iterations using the proposed EP framework, can
result in better MSE and MLPD scores.
However, such alternative initialization schemes were left out of these
experiments, because our aim was to test how good performance could be
obtained using only the EP algorithm with the zero initialization described
in Section \ref{sec_algorithm}.

The CPU times of Table \ref{table1} indicate that with small $n$ the
computational cost of NN-EP is larger compared to GP-ARD, which requires only
one $\mathcal{O}(n^3)$ Cholesky decomposition per analytically tractable
marginal likelihood evaluation. However, as $n$ increases GP-ARD becomes
slower, which is why several different sparse approximation schemes have been
proposed \citep[see, e.g,][]{Rasmussen+Williams:2006}.
Furthermore, assuming a non-Gaussian observation model, such as the binary
probit classification model, GP or GP-ARD would require several
$\mathcal{O}(n^3)$ iterations to form Laplace or EP approximations for the
marginal likelihood at each hyperparameter configuration. With NN-EP, probit
or Gaussian mixture models could be used without additional computations.
The computational cost of NN-EP increases linearly with $n$ and $K$, but as
$d$ increases the posterior updates of $q(\w_k)$, which scale as
$\mathcal{O}(Kd^3)$, become more demanding. The results of Table~\ref{table1}
were generated using a sequential scheme for updating $q(\w_k)$ (see
Algorithm~\ref{alg_mlp_ep}), which can be seen as larger computational costs
with respect to NN-MC with the Crime and Kin40k data sets.
One option with larger $d$ is to use parallel EP updates, but this may
require more damping or better initialization for the input weight
approximations.
Another possibility would be to use fully factorized posterior approximations
in place of $q(\w_k)$, or to assign different overlapping subgroups of the
input features into the different hidden units and to place hierarchical
prior scale parameters between the groups.

\section{Discussion}

In this article, we have described how approximate inference using EP can be
carried out with a two-layer NN model structure with sparse hierarchical
priors on the network weights, resulting in a novel method for nonlinear
regression problems.

We have described a computationally efficient EP algorithm that utilizes
independent approximations for the weights associated with the different
hidden units and layers to achieve computational complexity scaling similar
to an ensemble of $K$ sparse linear models.
More generally, our approach can be regarded as a non-linear adaptation of
the various EP methods proposed for sparse linear regression models.
This is achieved by constructing a factorized Gaussian approximation for the
posterior distribution resulting from the nonlinear MLP model structure with
a linear input layer, and adapting the algorithms proposed for sparse linear
models separately on the independent Gaussian approximations for each hidden
unit.
Because of the structure of the approximation, all existing methodology
presented for facilitating the computations in sparse linear models can be
applied on the hidden unit approximations separately.
We have also introduced an EP framework that enables definition of flexible
hierarchial priors using higher level scale parameters that are shared by a
group of independent linear models (in our case the hidden units). The
proposed EP approach enables efficient approximate integration over these
scale parameters simultaneously with the coefficients of the linear models.
We used this framework for inferring the common scale parameter of Laplace
priors assigned on the input weights, and to implement Gaussian ARD priors
for the input-layer.
In this article, we have focused on the Gaussian observation model, but the
method can be readily extended to others as well (e.g., binary probit
classification and robust regression with Gaussian mixture models).

Using simple artificial examples we demonstrated several desirable
characteristics of our approach. The sparsity promoting priors can be used to
suppress the confounding predictive influences of possibly irrelevant
features without the potential risk of overfitting associated with
point-estimate based ARD priors. More precisely, the approximate integration
over the posterior uncertainty helps to avoid pruning out potentially
relevant features in cases with large uncertainty on the input relevances.
Albeit more challenging to estimate, the finite parametric model enables a
posteriori inspection of the model structure and feature relevances using the
hyperparameter and weight approximations. Furthermore, the parametric model
structure can also be used to construct more restricted models by assigning
different input variables into different hidden units, grouping the inputs
using the hierarchical scale priors, using different nonlinear activation
functions for the different hidden units, or using fixed interaction terms
dependent on certain hidden units as inputs for the output-layer.

In deriving the EP algorithm, we have also described different computational
techniques that could be useful in other models and approximation methods.
These include the EP approximation for the hierarchical priors on the scale
parameters of the weights that could be useful in combining sparse linear
models associated with different subjects or measurement instances, the noise
estimation framework that could be used for estimating the likelihood
parameters in sparse linear models or approximate Gaussian filtering methods,
and the proposed approach for approximating the tilted distributions of the
hidden unit activations that could be useful in forming EP approximations for
observation models involving sums of nonlinear functions taken from random
variables with factorized Gaussian posterior approximations.

%The presented framework for sparse nonlinear modeling could be useful in
%problems with a large number of input features compared to $n$, requiring
%efficient adaptive control of model complexity.

%Moreover, the sparse priors on the output-layer weights enable automatic
%control of the model complexity by effectively removing unnecessary hidden
%units. Since the tilted moment evaluations, which require $K(K+1)/2$ smaller
%approximation steps in the more robust second method described in the section
%3.4, are the computationally most demanding part of the proposed algorithm,
%this output layer sparsity can also be utilized to speed up the computations
%by removing the unnecessary hidden units during the EP iterations.

%The challenge with the proposed approach is finding a fast computational
%implementation for the tilted moment integrations, which is a topic requiring
%further investigation. Once more efficient algorithms for this problem are
%developed, our framework could offer interesting alternatives to GPs  also in
%large-scale problems because of the linear scaling with respect to the number
%of observations.

% Acknowledgements should go at the end, before appendices and references

\acks{This research was partially funded by the Academy of Finland (grant
218248).}

%\newpage

\appendix

\section{Cavity Distributions with the Factorized Approximation}
\label{sec_cavity}

Assuming the factorized approximation of equation \eqref{eq_qep_fact} for
$q(\z)$, and applying the transformation $\h_i = \tilde{\x}_i^\Tr \w$ on
\eqref{eq_EP_cavity} results in the following cavity distribution for $\h_i$:
$q_{-i}(\h_i)=\prod_k \N(h_{i,k}| m_{-i,k}, V_{-i,k})$. The scalar cavity
means and variances are given by
\begin{align} \label{eq_cavity_w}
  V_{-i,k} &= (V_{i,k}^{-1} - \eta \tauwl_{i,k})^{-1} \nonumber\\
  m_{-i,k} &= V_{-i,k} (V_{i,k}^{-1} m_{i,k} - \eta \nuwl_{i,k}),
\end{align}
where the mean and variance of $h_{i,k}$ under the current approximation
\eqref{eq_qep_w} are denoted with $m_{i,k} = \x_i^\Tr \mq_{\w_k}$ and
$V_{i,k} = \x_i^\Tr \Sq_{\w_k} \x_i$, respectively.
Using \eqref{eq_qep_v} and \eqref{eq_EP_cavity} the $i$:th cavity
distribution for $\vv$ can be written as $q_{-i}(\vv)= \N(\vv| \mq_{-i},
\Sq_{-i})$ and the cavity mean and covariance are given by
\begin{align} \label{eq_cavity_v}
  \Sq_{-i} &= \Sq_{\vv} + \Sq_{\vv} \Tauvl_i s^{-1} \Tauvl_i^\Tr \Sq_{\vv}
  \nonumber\\
  \mq_{-i} &= \mathbf{a} + \Sq_{\vv} \Tauvl_i s^{-1} \Tauvl_i^\Tr \mathbf{a},
\end{align}
where $s=\eta^{-1} -\Tauvl_i^\Tr \Sq_{\w} \Tauvl_i$ and $\mathbf{a} =
\mq_{\vv}-\eta\Sq_{\vv}\Nuvl_i$.
Using \eqref{eq_cavity_w} and \eqref{eq_cavity_v} the cavity evaluations can
be implemented efficiently: for the input weights $\w_k$ only scalar moments
of $h_{i,k}$ need to be determined, and for the output weights $\vv$ rank-one
matrix updates are required. The cavity computations for the noise level term
approximations \eqref{eq_ttheta} and the weight prior term approximations
(\ref{eq_tvk}, \ref{eq_twk}) require only manipulation of univariate Gaussian
distributions and can be implemented similarly as in \eqref{eq_cavity_w}.

%\subsubsection{Tilted moments with the factorized approximation}
\section{Tilted Moments for the Output Weights} \label{sec_qv_tilted}

To obtain the desired site approximation structure \eqref{eq_tz_fact} and
closed form expressions for the the corresponding site parameters
($\Tauwl_i$, $\Nuwl_i$, $\Tauvl_i$, and $\Nuvl_i$) satisfying the moment
matching condition \eqref{eq_EP_moment_matching} we need to form suitable
approximations for the marginal means and covariances of $h_{i,k}$ and $\vv$
resulting from the tilted distribution $\eqref{eq_EP_tilted}$.
We start by assuming the noise level $\theta$ known and extend the presented
approach for approximate integration over $q_{-i}(\theta)$ later.

We first consider an approximate scheme which has already been utilized in
the unscented Kalman filtering framework for inferring the weights of a
neural network \citep{Wan:2000}.
Adopting the approach to our setting, first a cavity-predictive joint
Gaussian approximation is formed for the random vector $[\uu_i^{\Tr},
\tilde{y}_i]^\Tr = [\h_i^{\Tr}, \vv^{\Tr}, \tilde{y}_i]^\Tr$, which is
distributed according to $q(\h_i, \vv, \tilde{y}_i| \theta) \propto
p(\tilde{y}_i| f_i,\theta)^\eta q_{-i}(\h_i,\vv)$. This is done by
approximating the central moments $\E(\tilde{y}_i| \theta)$,
$\var(\tilde{y}_i| \theta)$, $\cov(\h_i,\tilde{y}_i| \theta)$, and
$\cov(\vv,\tilde{y}_i| \theta)$ using the unscented transform. Approximations
for the mean and covariance of the tilted distribution \eqref{eq_EP_tilted}
can now be determined by conditioning on $\tilde{y}_i$ in the joint Gaussian
approximation of $[\uu_i^{\Tr}, \tilde{y}_i]^\Tr$  to obtain
$\E(\uu_i|\tilde{y}_i, \theta)$ and $\cov(\uu_i|\tilde{y}_i, \theta)$, and
plugging in the observation $y_i$.
In our experiments, this approach was found sufficiently accurate for
approximating the moments of $\vv$, which is probably explained by the
conditional linear dependence of $f_i$ on $\vv$ in the observation model.
Thus, we approximate the marginal tilted distribution of $\vv$ with
$\hat{p}_i(\vv|\theta) \approx \N(\hat{\mq}_i(\theta), \hat{\Sq}_i(\theta))$,
where
\begin{align} \label{eq_tilted_v}
  \hat{\mq}_i (\theta) &= \mq_{-i} + \Sq_{\vv,f_i}  V_{y_i}^{-1} (y_i-m_{f_i})
  \nonumber\\
  \hat{\Sq}_i (\theta) &= \Sq_{-i} - \Sq_{\vv,f_i} V_{y_i}^{-1} \Sq_{\vv,f_i}^{\Tr},
\end{align}
and $V_{y_i} =V_{f_i} +\eta^{-1} \exp(\theta)$. Here we have defined the
required cavity predictive moments in terms of $f_i=\vv^\Tr \g(\h_i)$ instead
of $\tilde{y}_i$ to facilitate the upcoming approximate integration over
$q_{-i}(\theta)$,
%
%and these moments are denoted with $m_{f_i} = \E(f_i)$, $V_{f_i} =
%\var(f_i)$, and $\Sq_{\vv,f_i} =\cov(\vv,f_i)$, where the expectations are
%taken with respect to $q_{-i} (\h_i,\vv)= q_{-i} (\h_i) q_{-i} (\vv)$.
%
%Assuming the factorized posterior approximation \eqref{eq_qep_fact}, the
%cavity predictive moments required in \eqref{eq_tilted_v} can be written as
%
and assuming the factorized posterior approximation \eqref{eq_qep_fact},
these central moments can be written as
\begin{align} \label{eq_qf_fact}
  m_{f_i} &= \E(f_i) = \mq_{-i}^\Tr \m_{\g_i}
  \nonumber\\
  V_{f_i} &= \var(f_i) = \m_{\g_i}^\Tr \Sq_{-i} \m_{\g_i}
  + \V_{g_i}^\Tr( \diag( \Sq_{-i}) + \mq_{-i} \circ \mq_{-i})
  \nonumber\\
  \Sq_{\vv,f_i} &= \cov(\vv,f_i) =  \Sq_{-i} \m_{\g_i},
\end{align}
where $\circ$ denotes the element-wise matrix product, and the $(K+1) \times
1$ vectors $\m_{\g_i} =\E(\g(\h_i))$ and $\V_{\g_i} =\var(\g(\h_i))$ are
formed by computing the means and variances from each component of $\g(\h_i)$
with respect to $q_{-i}(\h_i)$ defined in \eqref{eq_cavity_w}. Note that the
last elements of $\m_{\g_i}$ and $\V_{\g_i}$ are one and zero corresponding
to the output bias term $v_0$.

With the activation function \eqref{eq_gx} the mean $\m_{\g_i}$ can be
computed analytically as
\begin{equation*}
  \E(g(h_{i,k})) = 2 K^{-1/2} \left(
  \Phi \left( m_{-i,k} (1+V_{-i,k})^{-1/2} \right) -0.5 \right),
\end{equation*}
and for computing the variance $\V_{\g_i}$ the following integral has to be
evaluated numerically
\begin{equation*}
  \var(g(h_{i,k})) =  2 (K\pi)^{-1} \int_0^{\sin^{-1} (\rho)}
  \exp \left( -\frac{m_{-i,k}^2}{ (1+V_{-i,k}) (1+\sin(x)) } \right) dx,
\end{equation*}
where $\rho= V_{-i,k} (1+V_{-i,k})^{-1}$. Other activation functions could be
incorporated by using only one-dimensional numerical quadratures whereas with
the full posterior couplings \eqref{eq_qep_z} $K$-dimensional numerical
integrations would be required to approximate $m_{f_i}$, $V_{f_i}$, and
$\Sq_{\vv,f_i}$.

\section{Tilted Moments for the Hidden Unit Activations}
\label{sec_qh_tilted}

The challenge in approximating the mean and variance of $\hat{p}_i(h_{i,k})$
is that this marginal density can have multiple distinct modes, one related
to the cavity distribution $q_{-i}(\h_i)$ and another related to the
likelihood $p(y_i|\vv^\Tr \g(\h_i),\theta)$, that is, to the values of
$h_{i,k}$ that give better fit for the left-out observation $y_i$. In our
numerical experiments, the previously described simple approach based on a
joint Gaussian approximation for $[\h_i^\Tr, \vv^\Tr f_i]$ was found to
underestimate the marginal probability mass of the latter mode related to
$y_i$ especially in cases where the modes were clearly separated from each
other.
This problem was found to be mitigated by decreasing $\eta$, which probably
stems from leaving a fraction of the old site approximation
$\tilde{t}_{\z,i}$ from the previous iteration in the approximation that in
turn shifts the cavity towards the observation $y_i$.
With some difficult data sets, $\eta$-values as small as 0.5 were found
necessary for obtaining a good data fit but usually this also required more
iterations for achieving convergence compared to larger values of $\eta$.

To form a robust approximation for the marginal tilted distributions of the
hidden unit activations $h_{i,k}$ also in case of multimodalities, we propose
an alternative approximate method that enables numerical integration over
$\hat{p}_i(h_{i,k}|\theta)$ using one-dimensional quadratures.
We aim to explore numerically the effect of each hidden activation $h_{i,k}$
on the tilted distributions $\hat{p}_i(\h_i,\vv| \theta) \propto
\N(y_i|\vv^\Tr \g(\h_i), \exp(\theta) )^\eta q_{-i}(\h_i) q_{-i}(\vv)$ when
all other activations $\h_{i,-k}$ and the weights $\vv$ are averaged out.
To approximate the marginalization over $\h_{i,-k}$ and $\vv$ we approximate
$q_{-i}(f_i|h_{i,k})$, that is, the cavity distribution of $f_i$ resulting
from $q_{-i}(\h_i) q_{-i}(\vv)$ by conditioning on $h_{i,k}$, with a
univariate Gaussian given by
\begin{equation} \label{eq_qf_cond}
  q_{-i}(f_i|h_{i,k}) \approx \N \left( f_i| m(h_{i,k}),V(h_{i,k}) \right),
\end{equation}
where $m(h_{i,k})$ and $V(h_{i,k})$ denote the mean and variance of $f_i$
computed with respect to $q_{-i}(\h_{i,-k}, \vv) = q_{-i}(\h_{i,-k})
q_{-i}(\vv)$.
The required conditional moments $m(h_{i,k})$ and $V(h_{i,k})$ can be
calculated using equation \eqref{eq_qf_fact} by modifying the $k$:th element
of $\m_{\g_i}$ and $\V_{\g_i}$ corresponding to the known values of
$h_{i,k}$, that is, $[\m_{\g_i}]_k=g(h_{i,k})$ and $[\V_{\g_i}]_k=0$.
Note that the possible numerical integrations for determining $\E(\g(\h_i))$
and $\var(\g(\h_i))$ need to be computed only once for each site update and
only the terms dependent on $\m_{\g_i,k}$ have to be re-evaluated for each
value of $h_{i,k}$.
The approximation \eqref{eq_qf_cond} can be justified using the central limit
theorem according to which the distribution of the sum in $f_{i} = \sum_{k'}
v_{k'} g(h_{i,k'}) +v_0$ given $h_{i,k}$ approaches a normal distribution as
$K$ increases.

Using equation \eqref{eq_qf_cond}, we can write the following approximation
for the marginal tilted distribution of $h_{i,k}$:
\begin{align} \label{eq_qh_tilted}
  \hat{p}_i(h_{i,k}| \theta) & \propto \int
  \N(y_i|\vv_{-k}^\Tr \g(\h_{i,-k}) +v_k h_{i,k},\exp(\theta) )^\eta
  q_{-i}(\vv) \prod_{k'=1}^{K} q_{-i}(h_{i,k'}) d\vv d\h_{i,-k}
  \nonumber \\
  & =  \int \N \left( y_i| f_i, \exp(\theta) \right)^\eta
  q_{-i}\left( f_i| h_{i,k} ) \right)
  q_{-i}(h_{i,k}) df_i
  \nonumber \\
  & \approx Z(\theta) \N \left( y_i| m(h_{i,k}), V(h_{i,k})+ \eta^{-1} \exp(\theta) \right)
  q_{-i}(h_{i,k}) \nonumber\\
  & \approx \hat{Z}_{i,k}(\theta) \N \big( h_{i,k}|\hat{m}_{i,k}(\theta),
  \hat{V}_{i,k}(\theta) \big),
\end{align}
%
%& \approx \int \N \left( y_i| f_i, \exp(\theta) \right)^\eta
%  \N \left( f_i| m(h_{i,k}),V(h_{i,k}) \right)
%  q_{-i}(h_{i,k}) df_i
%  \nonumber \\
%
where all output weights excluding $v_k$ are denoted by $\vv_{-k}$ and
$\hat{Z}_{i,k} (\theta)$ is a normalizing constant.
%
%an approximation for the normalization constant $\hat{Z}_i$ defined in
%\eqref{eq_EP_tilted}.
%
In the last step we have substituted approximation \eqref{eq_qf_cond} and
carried out the integration over $f_i$ analytically to give $Z(\theta) =
(2\pi \exp(\theta))^{(1-\eta)/2} \eta^{-1/2}$.
Approximation \eqref{eq_qh_tilted} enables numerical inspection for the
possible multimodality of $\hat{p}_i (h_{i,k}|\theta)$, and the conditional
tilted mean $\hat{m}_{i,k} (\theta)$ and variance $\hat{V}_{i,k} (\theta)$
can be determined using a numerical quadrature. Compared with the simple
approach described in Appendix \ref{sec_qv_tilted}, equation
\eqref{eq_qh_tilted} results in more accurate tilted mean estimates in
multimodal cases.

%Instead of mean and variance approximations might be useful,
% for example, only one mode could be summarized.

%This is done approximating the $\E(f_i|h_{i,k})$ and $\E(f_i|h_{i,k})$
%
%approximate $q_{-i}(f_i) =\N(m_{f_i},V_{f_i})$
%
%Condition analytically on $h_{i,k}$ and integrate over $\N (y_i|
%m_{f_i}(h_{i,k}), V_{f_i}(h_{i,k}) + \exp(\theta) )$
%
%we propose to approximate $q_{-i} (f_i|h_{i,k}) \approx \N ( f_i|
%m(h_{i,k}),V(h_{i,k}) )$

\section{Tilted Moments with Unknown Noise Level} \label{sec_qtheta_tilted}

If the noise level $\theta$ is assumed unknown and estimated using the EP,
the marginal mean $\hat{\mu}_{\theta,i}$ and variance
$\hat{\sigma}_{\theta,i}^2$ can be approximated with a similar approach as
was done for $h_{i,k}$ in Appendix \ref{sec_qh_tilted}. We approximate first
the cavity distribution of $f_i$ with $q_{-i}(f_i|\theta) \approx
\N(y_i|m_{f_i},V_{f_i})$, where the mean and variance are computed using
\eqref{eq_qf_fact}. Then, assuming a Gaussian observation model, we can
integrate analytically over $f_i$ to obtain a numerical approximation for the
tilted distribution of $\theta$:
\begin{align} \label{eq_qtheta_tilted}
  \hat{p}_i(\theta) & \propto \int
  \N(y_i|\vv^\Tr \g(\h_{i}),\exp(\theta) )^\eta
  q_{-i}(\vv) q_{-i}(\h_{i}) q_{-i}(\theta) d\vv d\h_i
  \nonumber \\
  & =  \int \N \left( y_i| f_i, \exp(\theta) \right)^\eta
  q_{-i}( f_i ) q_{-i}(\theta) df_i
  \nonumber \\
  & \approx Z(\theta) \N \left( y_i| m_{f_i}, V_{f_i}+ \eta^{-1} \exp(\theta) \right)
  q_{-i}( \theta )
  \approx \hat{Z}_i \N(\theta |\hat{\mu}_{\theta,i},\hat{\sigma}^2_{\theta,i}),
\end{align}
where $Z(\theta) = (2\pi \exp(\theta))^{(1-\eta)/2} \eta^{-1/2}$, and
$\hat{Z}_i$ is an approximation for the normalization term of the tilted
distribution \eqref{eq_EP_tilted}. Using equation \eqref{eq_qtheta_tilted}
the approximate mean $\hat{\mu}_{\theta,i}$, variance
$\hat{\sigma}_{\theta,i}^2$, and normalization term $\hat{Z}_i$ can be
calculated with a numerical quadrature, and if $\theta$ is known or fixed,
the normalization term can by approximated with $\hat{Z}_i(\theta) =
Z(\theta) \N \left( y_i| m_{f_i}, V_{f_i}+ \eta^{-1} \exp(\theta) \right)$.

To approximate the marginal means and covariances of $\vv$ and $h_{i,k}$ with
unknown $\theta$ the conditional approximations of equations
\eqref{eq_tilted_v} and \eqref{eq_qh_tilted} have to be integrated over
$\hat{q}_i(\theta) = \hat{Z}_i^{-1} \hat{Z}_i(\theta) q_{-i}(\theta) \approx
\hat{p}_i(\theta)$
because we have $\hat{p}_{i}(\h_i,\vv,\theta) \approx \hat{Z}_i^{-1}
\hat{p}_{i}(\h_i,\vv|\theta) \hat{Z}_i(\theta) q_{-i}(\theta)$ according to
\eqref{eq_qtheta_tilted}.
In case of the simple joint Gaussian approximation for $\vv$ we can write
\begin{align} \label{eq_mq_tilted_v2}
  \hat{\mq}_i = \E_{\hat{p}_{i}(\vv)} (\vv)
  & = \E_{\hat{p}_i(\theta)} \big(\E_{\hat{p}_{i}(\vv|\theta)}(\vv|\theta) \big)
  \approx \E_{\hat{q}_i(\theta)} (\hat{\mq}_i (\theta) ) \nonumber\\
  &= \mq_{-i} + \Sq_{\vv,f_i} \E_{\hat{q}_i(\theta)}
  \big( V_{y_i}^{-1} \big) (y_i-m_{f_i}),
\end{align}
%
%&\int
%\N(y_i|\vv^\Tr \g(\h_{i}),\exp(\theta) )^\eta
%q_{-i}(\vv) q_{-i}(\h_{i}) q_{-i}(\theta) d\vv d\h_i \\
%
where the conditional mean of $\vv$ with respect to $\hat{p}_{i}(\vv|\theta)$
is approximated using \eqref{eq_tilted_v}, and the integration over
$V_{y_i}^{-1}=(V_{f_i}+ \eta^{-1} \exp(\theta))^{-1}$ can be done using a
one-dimensional quadrature.
Similarly, for the marginal covariance of $\vv$ we can write
\begin{align} \label{eq_Sq_tilted_v2}
  \hat{\Sq}_i = \cov_{\hat{p}_{i}(\vv)} (\vv) & = \E_{\hat{p}_i(\theta)}
  \big( \cov_{\hat{p}_{i}(\vv|\theta)} (\vv|\theta) \big)
  +\cov_{\hat{p}_i(\theta)} \big( \E_{\hat{p}_{i}(\vv|\theta)} (\vv|\theta) \big)
  \nonumber\\
  &\approx \E_{\hat{q}_i(\theta)} \big( \hat{\Sq}_i (\theta) \big)
  +\E_{\hat{q}_i(\theta)} \Big( \big( \hat{\mq}_i(\theta) -\hat{\mq}_i \big)
    ( \hat{\mq}_i(\theta) -\hat{\mq}_i )^\Tr \Big)
  \nonumber\\
  &= \Sq_{-i} - \Sq_{\vv,f_i}  \left( \E_{\hat{q}_i(\theta)} \big( V_{y_i}^{-1} \big)
  -(y_i-m_{f_i})^2 \var_{\hat{q}_i(\theta)} \big( V_{y_i}^{-1} \big)
  \right) \Sq_{\vv,f_i}^{\Tr} ,
\end{align}
where the conditional covariance of $\vv$ with respect to
$\hat{p}_{i}(\vv|\theta)$ is approximated using \eqref{eq_tilted_v} and
$\var_{\hat{q}_i(\theta)} \big( V_{y_i}^{-1} \big) = \E_{\hat{q}_i(\theta)}
\big( V_{y_i}^{-1} -\E_{\hat{q}_i(\theta)} (V_{y_i}^{-1}) \big)^2$ can be
computed with a numerical quadrature.
%
%the outer expectations on the left hand side of  are taken with respect to
%$q_{-i}(\theta)$.

For the output weights $\vv$ the integration over the uncertainty of $\theta$
can be done without significant additional computational cost. The mean and
variance of $V_{y_i}^{-1}$ can be determined using the same function
evaluations that are used in the quadrature integrations required for
computing $\hat{\mu}_{\theta,i}$, $\hat{\sigma}^2_{\theta,i}$, and
$\hat{Z}_i$ according to equation \eqref{eq_qtheta_tilted}.
Approximating the marginal means and covariances of the hidden unit
activations $h_{i,k}$ is more demanding because integration over the
approximate marginal tilted distribution resulting from approximation
\eqref{eq_qh_tilted},
\begin{equation} \label{eq_qh_tilted_2d}
  \hat{p}_i({h_{i,k},\theta}) \approx \hat{Z}_i^{-1} Z(\theta)
  \N \left( y_i| m(h_{i,k}),V(h_{i,k})+ \eta^{-1} \exp(\theta) \right)
  q_{-i}(h_{i,k}) q_{-i}(\theta),
\end{equation}
would require a two-dimensional numerical quadratures for each hidden unit
$K$. To reduce the computational burden, we approximate the probability mass
of $\hat{p}_i({h_{i,k},\theta})$ to be relatively sharply peaked near the
marginal expected value $\hat{\mu}_{\theta,i}$ resulting from
\eqref{eq_qtheta_tilted} yielding
\begin{align} \label{eq_qh_tilted2}
  \hat{p}_i(h_{i,k}) &\approx \hat{Z}_i^{-1} Z(\theta)
  \N \left( y_i| m(h_{i,k}),V(h_{i,k})+ \eta^{-1} \exp(\hat{\mu}_{\theta,i}) \right)
  q_{-i}(h_{i,k})
  \nonumber\\
  &\approx \N \big( h_{i,k}|\hat{m}_{i,k}(\hat{\mu}_{\theta,i}),
  \hat{V}_{i,k}(\hat{\mu}_{\theta,i}) \big).
\end{align}
This approximation does not require additional computational effort compared
to the conditional estimate \eqref{eq_qh_tilted} and the difference in
accuracy compared to the two-dimensional quadrature estimate based on
\eqref{eq_qh_tilted_2d} is small after a few iterations provided that there
are enough observations.

% update q(\theta) with fixed \w first !

\section{Site Parameters and Updates}
\label{sec_site_updates}

In this appendix we present closed form expressions for the parameters of the
site approximations \eqref{eq_tz_fact} resulting from the moment matching
condition \eqref{eq_EP_moment_matching} and the approximate tilted moments
derived in Appendices \ref{sec_qv_tilted} -- \ref{sec_qtheta_tilted}.

Using the moment matching condition
$\hat{\Sq}_i^{-1}= \Sq_{-i}^{-1} + \eta \Tauvl_i \Tauvl_i^\Tr$
resulting from \eqref{eq_EP_moment_matching} and approximations
\eqref{eq_tilted_v} or \eqref{eq_Sq_tilted_v2} we can write the following
expression for the scale parameter vector $\Tauvl_i$ of the $i$:th
approximate site term $\tilde{t}_{\vv,i}$ related to the output weights:
\begin{align} \label{eq_sites_tauv}
  \Tauvl_i = \m_{\g_i} \text{sign}(\hat{a}_i) |\hat{a}_i|^{1/2}
  \left(1 - \hat{a}_i \m_{\g_i}^\Tr \Sq_{-i} \m_{\g_i} \right)^{-1/2} \eta^{-1/2},
\end{align}
where $\hat{a}_i =  \E_{\hat{q}_i(\theta)} \big( V_{y_i}^{-1} \big)
-(y_i-m_{f_i})^2 \var_{\hat{q}_i(\theta)} \big( V_{y_i}^{-1} \big)$ with
unknown $\theta$ and $\hat{a}_i = V_{y_i}^{-1}$ otherwise.
Similarly for the location parameter vector $\Nuvl_i$, equation
\eqref{eq_EP_moment_matching} results in the moment matching condition
$\hat{\Sq}_i^{-1} \hat{\mq}_i= \Sq_{-i}^{-1} \mq_{-i} + \eta \Nuvl_i$
that together with equation \eqref{eq_tilted_v} or \eqref{eq_mq_tilted_v2}
gives
\begin{align} \label{eq_sites_nuv}
  \Nuvl_i = \m_{\g_i} \left(1 - \hat{a}_i \m_{\g_i}^\Tr \Sq_{-i} \m_{\g_i}
  \right)^{-1}
  \left( \hat{a}_i \m_{\g_i}^\Tr \mq_{-i}
  + \hat{b}_i (y_i - m_{f_i}) \right) \eta^{-1}
\end{align}
where $\hat{a}_i$ is defined as in the previous equation, and $\hat{b}_i =
\E_{\hat{q}_i(\theta)} \big( V_{y_i}^{-1}\big)$ when $\theta$ is unknown and
otherwise $\hat{b}_i = V_{y_i}^{-1}$.
By looking at equations \eqref{eq_sites_tauv} and \eqref{eq_sites_nuv} we can
now extend the discussion of the last paragraph of Section \ref{sec_approximation}.
The mean and covariance of the posterior approximation $q(\vv)$ defined in equation
\eqref{eq_qep_v} can be interpreted as the posterior distribution of a linear
model where the input features are replaced with the expected values of the
nonlinearly transformed input layer activations $\m_{\g_i} = \E_{q_{-i}}(\g(
\tilde{\x}_i^\Tr \w))$ and pseudo observations $\tilde{y}_i = \m_{\g_i}^\Tr
\mq_{-i} + \hat{a}_i^{-1} \hat{b}_i (y_i - m_{f_i})$ are made according to an
observation model $\N (\tilde{y}_i| \m_{\g_i}^\Tr \vv, \hat{a}_i^{-1} -
\m_{\g_i}^\Tr \Sq_{-i} \m_{\g_i})$.

Damping the site updates can improve the numerical robustness and convergence
of the EP algorithm, but applying damping on the site precision structure
$\Tt_{i,\vv\vv} =\Tauvl_i \Tauvl_i^\Tr$ resulting from equations
\eqref{eq_qep_v} and \eqref{eq_sites_tauv}, that is,
$\Tt_{i,\vv\vv}^{\text{new}} = (1-\delta) \Tauvl_i^{\text{old}}
(\Tauvl_i^{\text{old}})^\Tr + \delta \Tauvl_i \Tauvl_i^\Tr$,
would break the outer product form of the $i$:th likelihood site
approximation \eqref{eq_tz_fact} and produce a computationally more demanding
rank-$K$ site precision after $K$ iterations.
In case the input weight approximations $q(\w_k)$ were kept fixed while
updating the output weights $\vv$, the expected activations $\m(\g_i)$ would
remain constant and one could consider damping only the scalar terms on the
right hand side of equations \eqref{eq_sites_tauv} and \eqref{eq_sites_nuv}.

In the more general case where also the site parameters $\tauwl_{i,k}$ and
$\nuwl_{i,k}$ related to the input weights are updated simultaneously, we can
approximate the new site precision structure
$\Tt_{i,\vv\vv}^{\text{new}} = \A_i \A_i^\Tr$, where $\A_i =
[(1-\delta)^{1/2} \Tauvl_i^{\text{old}}, \delta^{1/2} \Tauvl_i]$ and
$\Tauvl_i$ is obtained from \eqref{eq_sites_tauv},
with its largest eigenvector at each site update step. This requires solving
the eigenvector $\mb{v}_i$ corresponding to the largest eigenvalue
$\lambda_i$ of the $2 \times 2$ matrix $\A_i^\Tr \A_i \approx \mb{v}_i
\lambda_i \mb{v}_i^\Tr$ after which the new damped site parameter vector can
be approximated as %$\Tauvl_i^{\text{new}} = \mb{A}_i \mb{v}_i$.
\begin{align} \label{eq_siteupdate_tauv}
  \Tauvl_i^{\text{new}} = \mb{A}_i \mb{v}_i.
\end{align}
Damping the site location vector $\Nuvl_i$ is straightforward because update
$\Nuvl_i^{\text{new}} = (1-\delta) \Nuvl_i^{\text{old}} + \delta \Nuvl_i =
\mb{b}_i$, where $\Nuvl_i$ is obtained from \eqref{eq_sites_nuv},
will preserve the structure of the site approximation \eqref{eq_tz_fact}.
However, approximation $\Tauvl_i^{\text{new}} = \mb{A}_i \mb{v}_i$ changes
the moment consistency conditions used in deriving \eqref{eq_sites_nuv} which
is why $\Nuvl_i^{\text{new}}$ has to be modified so that combining it with
$\Tauvl_i^{\text{new}}$ according to the moment matching rule
\eqref{eq_EP_moment_matching} results in the same mean vector $\mq_{\vv}$ as
the rank-2 site $\A_i \A_i^\Tr$ combined with $\mb{b}_i$. In other words, we
approximate the posterior covariance $\Sq_\vv$ resulting from the rank-two
damped update but choose $\Nuvl_i^{\text{new}}$ so that the mean $\mq_\vv$
will be exact.
%
%\begin{align*}
%  (\Sq_{-i}^{-1} + \eta \Tauvl_i^{\text{new}} (\Tauvl_i^{\text{new}})^\Tr )^{-1}
%  (\Sq_{-i}^{-1} \mq_{-i} + \eta \Nuvl_i^{\text{new}})
%  =
%  (\Sq_{-i}^{-1} + \eta \A_i \A_i^\Tr)^{-1} (\Sq_{-i}^{-1} \mq_{-i} +
%  \eta \mb{b}_i)
%\end{align*}
%
This is achieved by updating the site location according to
%
% \Tauvl_i^{\text{new}} &= \A_i \vv_i \nonumber\\
%
\begin{align} \label{eq_siteupdate_nuv}
  \Nuvl_i^{\text{new}} &= \mb{b}_i + \eta^{-1} \A_i (\vv_i \vv_i^\Tr -\mb{I})
  (\A_i^\Tr \Sq_{-i} \A_i + \eta^{-1} \mb{I})^{-1}
  \A_i^\Tr (\mq_{-i} +\eta \Sq_{-i} \mb{b}_i )
\end{align}
where $\mb{b}_i = (1-\delta) \Nuvl_i^{\text{old}} + \delta \Nuvl_i$.

With the factorized posterior approximation \eqref{eq_qep_fact} the
parameters of the likelihood site approximation terms $\tilde{t}_{\w_k,i}$
associated with the input weights decouple over the different hidden units
$k=1,...,K$ and consequently the moment matching condition
\eqref{eq_EP_moment_matching} results in simple scalar site parameter
updates.
Using the moment matching condition with the cavity definitions
\eqref{eq_cavity_w} and the approximation \eqref{eq_qh_tilted} or
\eqref{eq_qh_tilted2} gives the following site updates
\begin{align} \label{eq_sites_w}
  \tauwl_{i,k}^{\text{new}} &= (1-\delta) \tauwl_{i,k}
  + \delta \eta^{-1} (\hat{V}_{i,k}^{-1} - V_{-i,k}^{-1}) =
  \tauwl_{i,k} + \delta \eta^{-1} (\hat{V}_{i,k}^{-1} - V_{i,k}^{-1})
  \\
  \nuwl_{i,k}^{\text{new}} &= (1-\delta) \nuwl_{i,k}
  + \delta \eta^{-1} (\hat{V}_{i,k}^{-1} \hat{m}_{i,k} - V_{-i,k}^{-1} m_{-i,k})
  = \nuwl_{i,k} +\delta \eta^{-1}
  (\hat{V}_{i,k}^{-1} \hat{m}_{i,k} - V_{i,k}^{-1} m_{i,k}), \nonumber
\end{align}
where $\delta \in (0,1]$ is a damping factor and the marginal tilted moments
resulting from either fixed or unknown $\theta$ are denoted simply with
$\hat{m}_{i,k}$ and $\hat{V}_{i,k}$.
Equation \eqref{eq_sites_w} shows that the EP iterations on the input weights
$\w_k$ have converged when the approximate marginal means $m_{i,k}$ and
variances $V_{i,k}$ of the activations $h_{i,k}$ from all hidden units are
consistent with all tilted distributions \eqref{eq_EP_tilted}.

\section{Computing the Predictions} \label{sec_predictions}

The prediction for a new test input $\x_*$ can be computed using
approximations \eqref{eq_qep_theta}, \eqref{eq_qep_w} and \eqref{eq_qep_v},
as follows
\begin{align} \label{eq_pred}
  p(y_*|\x_*) &\approx \int p(y_*|f(\x_*),\theta) q( \vv| \mq_\vv, \Sq_\vv)
  \prod_{k=1}^K q( \w_k| \mq_{\w_k}, \Sq_{\w_k})
  q(\theta| \mu_\theta, \sigma_\theta^2 ) d\vv d\w d\theta
  \nonumber \\
  &\approx \int \N( y_*|f_*,\exp(\theta) ) \N(f_*| m_{f_*}, V_{f_*})
  q(\theta) df_* d\theta
  \nonumber \\
  &= \int \N( y_*| m_{f_*}, V_{f_*} + \exp(\theta)) q(\theta) d \theta,
\end{align}
where the approximate mean $m_{f_*}$ and $V_{f_*}$ of the latent function
value $f_* = \sum_{k=1}^K v_k g(\w_k^\Tr \x_*) +v_0$ is approximated in the
same way as in equation \eqref{eq_qf_fact}. The cavity mean $\mq_{-i}$ and
covariance $\Sq_{-i}$ are replaced with $\mq_\vv$ and $\Sq_\vv$, and the
activation means $\m_{\g_*} = \E (\g(\h_*))$ and variances $\V_{\g_*} = \var
(\g(\h_*))$ are computed with respect to the approximations $q(\w_k)$.
The predictive mean is given by $\E(y_*|\x_*) =  \E ( \E(y_*|\x_*, \theta) )
= m_{f_*}$. The predictive variances $\var(y_*|\x_*) =  \E ( \var(y_*|\x_*,
\theta) ) + \var ( \E(y_*|\x_*, \theta) ) = V_{f_*} + \E( \exp(\theta) )$ and
the predictive densities $p(y_*|\x_*)$, can be approximated either with a
plugin value for $\theta=\mu_\theta$ or by integration over $\theta$ using a
numerical quadrature (in the experiments we used numerical quadratures).

\section{Marginal Likelihood Approximation}
\label{sec_marg_likelih}

An EP approximation for the log marginal likelihood $\log p(\y|\X)$ can be
computed in a numerically stable and efficient manner following the general
EP formulation for Gaussian approximating families summarized by
\citet[][appendix C]{Cseke:2011a}:
\begin{align} \label{eq_marg_likelih}
  \log Z_{\text{EP}} &= \Psi (\mq_\vv,\Sq_\vv)
  +\sum_{k=1}^K \Psi (\mq_{\w_k},\Sq_{\w_k})
  +\Psi (\mu_{\theta},\sigma_{\theta}^2)
  +\sum_{l=1}^{L} \Psi(\mu_{\phi,l},\sigma^2_{\phi,l})
  \nonumber \\
  &+ \frac{1}{\eta} \sum_{i=1}^n \Big( \ln \hat{Z}_i
  +\Psi (\mu_{\theta,-i},\sigma_{\theta,-i}^2)
  -\Psi (\mu_{\theta},\sigma_{\theta}^2) \Big)
  \nonumber \\
  & +\frac{1}{\eta} \sum_{i=1}^n \Big(
  -\frac{1}{2} \big( \ln (s_i\eta) + s_i^{-1} (\mb{a}_i^\Tr \Tauvl_i)^2
  - \eta \Nuvl_i^\Tr(\mq_{\vv}+\mb{a}_i) \big) \Big)
  \nonumber\\
  & +\frac{1}{\eta} \sum_{n=1}^n \sum_{k=1}^K
  \Big( \Psi (m_{i,k},V_{i,k}) - \Psi (m_{-i,k},V_{-i,k}) \Big)
  +\frac{1}{\eta_w}\sum_{j=1}^{Kd} \ln \hat{Z}_{w_j}
  \nonumber\\
  &+\frac{1}{\eta_w} \sum_{j=1}^{Kd} \Big(
  \Psi(\mu_{w,-j},\sigma_{w,-j}^2)
  -\Psi(\mu_{w,j},\sigma_{w,j}^2)
  +\Psi(\mu_{\phi,-j},\sigma_{\phi,-j}^2)
  -\Psi (\mu_{\phi,l_j},\sigma_{\phi,l_j}^2) \Big)
  \nonumber\\
  & +\frac{1}{\eta_v} \sum_{k=1}^{K} \Big( \ln \hat{Z}_{v_k}
  + \Psi (\mu_{v,-k},\sigma_{v,-k}^2)
  -\Psi (\mu_{v,k},\sigma_{v,k}^2)  \Big)
  \nonumber\\
  & -\Psi(\mu_{v_0},\sigma^2_{v_0})
  -\Psi(\mu_{\theta,0},\sigma^2_{\theta,0})
  -\sum_{l=1}^{L} \Psi(\mu_{\phi,0},\sigma^2_{\phi,0}),
\end{align}
where $s_i = \eta^{-1} + \Tauvl_i^\Tr \Sq_{\w_k} \Tauvl_i$, $\mb{a}_i =
\mq_\vv -\eta \Sq_\vv \Nuvl_i$, and the normalization term of the tilted
distribution $\hat{Z}_i \approx \int p(y_i|\vv^\Tr \g(\h_i), \theta)^\eta
q_{-i}(\vv,\h_i,\theta) d\vv d\h_i d\theta$ is computed using
\eqref{eq_qtheta_tilted}.
The normalization terms of the other tilted distributions are defined as
\begin{align*}
\hat{Z}_{v_k} = \int p(v_k|\sigma_{v,0}^2)^{\eta_v} q_{-k}(v_k) d v_k \quad \text{and} \quad
\hat{Z}_{w_j} = \int p(w_j|\phi_{l_j})^{\eta_w} q_{-j}(w_j)
q_{-j}(\phi_{l_j}) d w_j d \phi_{l_j},
\end{align*}
and they can be computed using numerical quadratures.
The normalization terms (also known as log partition functions) related to
the Gaussian cavity and marginal distributions can be computed as
\begin{equation*}
  \Psi(\mq,\Sq) =  \frac{1}{2} \mq^\Tr \bm{\nu}
  +\frac{1}{2} \log |\Sq| + \frac{d}{2} \log(2\pi),
\end{equation*}
where $\mq$ and $\bm{\nu} = \Sq^{-1} \mq$ are $d \times 1$ vectors and $\Sq$
is a $d \times d$ matrix.
In the fifth line of \eqref{eq_marg_likelih} the means and variances of the
$j$:th cavity distribution related to the prior term $p(w_j|\phi_{l_j})$ are
denoted according to (see equation \eqref{eq_cavity_pw})
\begin{equation*}
  q_{-j}(w_j,\phi_{l_j}) = \N (w_j|\mu_{w,-j},\sigma_{w,-j}^2)
  \N(\phi_{l_j}| \mu_{\phi,-j},\sigma_{\phi,-j}^2 ),
\end{equation*}
and the corresponding approximate marginal distributions are defined as
\begin{equation*}
  q(w_j,\phi_{l_j}) = \N (w_j|\mu_{w,j},\sigma_{w,j}^2)
  \N(\phi_{l_j}| \mu_{\phi,l_j},\sigma_{\phi,l_j}^2 ).
\end{equation*}
Similar notation is used for the likelihood term approximations of $\theta$
in the second line and the prior term approximations of $\{ v_k \}_{k=1}^K$
in the sixth line.
The last line of \eqref{eq_marg_likelih} contains the constant normalization
terms related to the fixed Gaussian priors of the output bias $v_0$, the
noise level $\theta=\log \sigma^2$, and the input weight scales $\{ \phi_l
\}_{l=1}^L$.

All terms of equation \eqref{eq_marg_likelih} excluding $\Psi
(\mq_\vv,\Sq_\vv)$ and $\Psi (\mq_{\w_k},\Sq_{\w_k})$ can be computed without
significant additional cost simultaneously during the EP update of the
corresponding site approximation.
Term $\Psi (\mq_\vv,\Sq_\vv)$ can be computed using one Cholesky
decomposition at each parallel update step of $q(\w_k)$ in line 7 of
Algorithm \ref{alg_mlp_ep}.
Similarly, if parallel updates are used for the input weight approximations,
$\Psi (\mq_{\w_k},\Sq_{\w_k})$ can be computed using the same Cholesky
decompositions that are used to recompute $q(\w_k)$ in line 6 of Algorithm
\ref{alg_mlp_ep}.
In case sequential EP is used for $q(\w_k)$ in line 5 of Algorithm
\ref{alg_mlp_ep}, vectors $\bm{\nu}_{\w_k}= \Sq_{\w_k}^{-1} \mq_{\w_k}$ and
determinant term $\log|\Sq_\vv|$ can be updated simultaneously with the
rank-1 updates of $\mq_{\w_k}$ and $\Sq_{\w_k}$.

The EP approximation $\log Z_{EP}$ has the appealing property that its
partial derivatives with respect to the site parameters in their canonical
forms (for example, $\tauwl_{i,k}$, $\nuwl_{i,k}$, $\Tt_{i,\vv\vv} = \Tauvl_i
\Tauvl_i^\Tr$, $\Nuvl_i$, $\tilde{\tau}_{\theta,i} =
\sigmat_{\theta,i}^{-2}$, and $\tilde{\nu}_{\theta,i} =
\sigmat_{\theta,i}^{-2} \mut_{\theta,i}$) are zero when the algorithm has
been iterated until convergence \citep{Opper:2005}. This follows form the
fact that the fixed points of the EP algorithm correspond to the stationary
points of \eqref{eq_marg_likelih} with respect to the site parameters (or
equivalently the cavity parameters defined using constraints of the form
$\tauwl_{i,k} = V_{i,k}^{-1} - V_{-i,k}^{-1}$).
Thereby the marginal likelihood approximation can be used in gradient-based
estimation of hyperparameters such as $\theta$, $\sigma_{v,0}^2$,
$\sigma_{v_0,0}^2$, or $\{ \phi_l \}_{l=1}^L$ in case they are not inferred
within the EP framework for determining $\{ q(\w_k) \}_{k=1}^K$ and $q(\vv)$.
Because the convergence of the likelihood approximation can take many
iterations it is advisable to initialize the hyperparameters to sensible
values and run the EP algorithm once until sufficient convergence starting
from zero initialization for the site parameters. After that gradient-based
local update steps can be taken for the hyperparameter values by continuing
the EP iterations from the previous site parameter values at each new
hyperparameter configuration.

%\appendix
%
%\section*{Appendix A.}
%\label{app:theorem}
%
%% Note: in this sample, the section number is hard-coded in. Following
%% proper LaTeX conventions, it should properly be coded as a reference:
%
%%In this appendix we prove the following theorem from
%%Section~\ref{sec:textree-generalization}:
%
%In this appendix we prove the following theorem from Section~6.2:
%
%\noindent {\bf Theorem} {\it Let $u,v,w$ be discrete variables such that $v,
%w$ do not co-occur with $u$ (i.e., $u\neq0\;\Rightarrow \;v=w=0$ in a given
%dataset $D$). Let $N_{v0},N_{w0}$ be the number of data points for which
%$v=0, w=0$ respectively, and let $I_{uv},I_{uw}$ be the respective empirical
%mutual information values based on the sample $D$. Then
%\[
%	N_{v0} \;>\; N_{w0}\;\;\Rightarrow\;\;I_{uv} \;\leq\;I_{uw}
%\]
%with equality only if $u$ is identically 0.} \hfill\BlackBox
%
%\noindent {\bf Proof}. We use the notation:
%\[
%P_v(i) \;=\;\frac{N_v^i}{N},\;\;\;i \neq 0;\;\;\;
%P_{v0}\;\equiv\;P_v(0)\; = \;1 - \sum_{i\neq 0}P_v(i).
%\]
%These values represent the (empirical) probabilities of $v$ taking value
%$i\neq 0$ and 0 respectively.  Entropies will be denoted by $H$. We aim to
%show that $\frac
%{I_{uv}}{P_{v0}} < 0$....\\

%\vskip 0.2in

\bibliography{ref}

\end{document}